\documentclass[10pt,twocolumn,letterpaper]{article}
\usepackage[numbers,sort&compress]{natbib}

\usepackage[pagenumbers]{cvpr} 

\usepackage{graphicx}
\usepackage{amsmath}
\usepackage{amssymb}
\usepackage{booktabs}
\usepackage{breqn}
\usepackage{mathtools}
\usepackage[hypcap=false]{caption}
\usepackage{multicol}
\usepackage{multirow}
\usepackage{makecell}
\usepackage{float}

\usepackage[pagebackref,breaklinks,colorlinks]{hyperref}

\usepackage[capitalize]{cleveref}
\crefname{section}{Sec.}{Secs.}
\Crefname{section}{Section}{Sections}
\Crefname{table}{Table}{Tables}
\crefname{table}{Tab.}{Tabs.}

\newcommand{\myparagraph}[1]{\medskip\noindent\textbf{#1}}

\def\cvprPaperID{****} 
\def\confName{CVPR}
\def\confYear{2023}

\begin{document}

\title{Teaching Matters: Investigating the Role of Supervision in Vision Transformers \vspace{-0.5em}}
\newcommand\blfootnote[1]{%
  \begingroup
  \renewcommand\thefootnote{}\footnote{#1}%
  \addtocounter{footnote}{-1}%
  \addtocounter{Hfootnote}{-1}%
  \endgroup
}

\author{%
Matthew Walmer\thanks{Equal contributors.} 
\newcommand\CoAuthorMark{\footnotemark[\arabic{footnote}]}
\quad
Saksham Suri\protect\CoAuthorMark~
\quad Kamal Gupta
\quad Abhinav Shrivastava \\
\quad University  of Maryland, College Park
}

\maketitle
\blfootnote{Accepted to CVPR 2023.}

\begin{abstract}
Vision Transformers (ViTs) have gained significant popularity in recent years and have proliferated into many applications.
However, their behavior under different learning paradigms is not well explored.
We compare ViTs trained through different methods of supervision, and show that they learn a diverse range of behaviors in terms of their attention, representations, and downstream performance.
We also discover ViT behaviors that are consistent across supervision, including the emergence of Offset Local Attention Heads. These are self-attention heads that attend to a token adjacent to the current token with a fixed directional offset, a phenomenon that to the best of our knowledge has not been highlighted in any prior work.
Our analysis shows that ViTs are highly flexible and learn to process local and global information in different orders depending on their training method. We find that contrastive self-supervised methods learn features that are competitive with explicitly supervised features, and they can even be superior for part-level tasks. We also find that the representations of reconstruction-based models show non-trivial similarity to contrastive self-supervised models. Project \href{https://www.cs.umd.edu/~sakshams/vit_analysis}{website} and \href{https://www.github.com/mwalmer-umd/vit_analysis}{code} are publicly available.
\end{abstract}

\begin{figure}
    \centering
    \includegraphics[width=\linewidth]{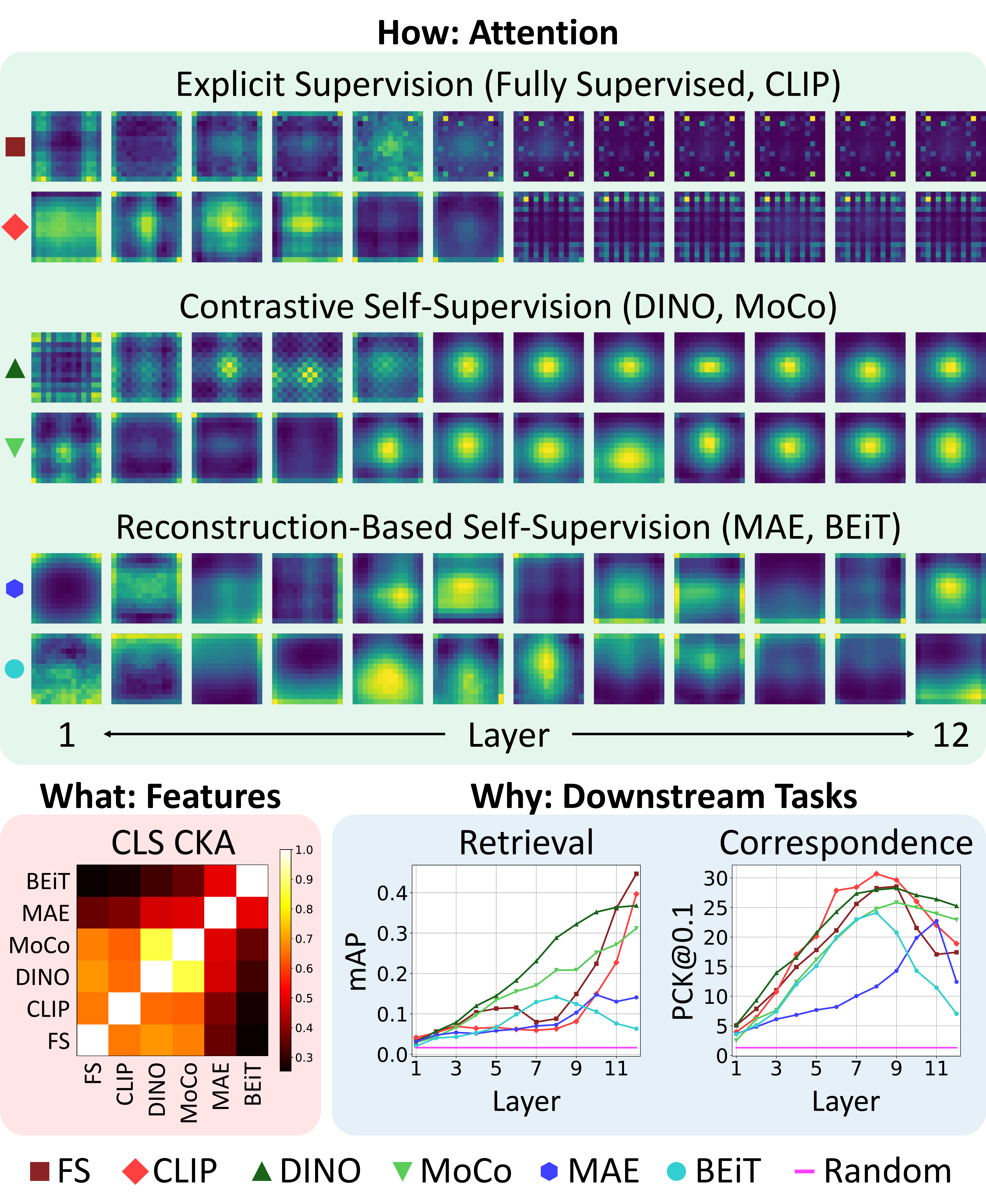}
    \vspace{-0.2in}
    \caption{
    \textbf{ViTs exhibit highly varied behaviors depending on their method of training.}
    In this work, we compare ViTs through three domains of analysis representing the How, What, and Why of ViTs.
    \textbf{How} do ViTs process information through attention? (Top) Attention maps averaged over 5000 images show clear differences in the mid-to-late layers.
    \textbf{What} do ViTs learn to represent? (Left) Contrastive self-supervised ViTs have a greater feature similarity to explicitly supervised ViTs, but also have some similarity with ViTs trained through masked reconstruction.
    \textbf{Why} do we care about using ViTs? (Right) We evaluate ViTs on a variety of global and local tasks and show that the best model and layer vary greatly.
    }
    \label{fig:teaser}
    \vspace{-0.2in}
\end{figure}
\section{Introduction}
\label{sec:intro}

The field of Computer Vision has advanced massively in the past decade, largely built on the backbone of Convolutional Neural Networks (CNNs). More recently, Vision Transformers (ViTs)~\citep{dosovitskiy2020image}  have shown the potential to overtake CNNs as the go-to visual processing model.
Prior works have asked the question \textit{do ViTs see like CNNs do?}~\citep{raghu2021vision}, but in this work, we ask: \textit{how do ViTs learn under different supervision?} Past examinations of ViTs have largely focused on models trained through full supervision. Instead, we aim to characterize the differences and similarities of ViTs trained through varying training methods, including self-supervised methods.
Unlike CNNs, the ViT architecture imposes few structural biases to guide the learning of representations.
This gives them the flexibility to learn diverse information processing strategies, and through our analyses, we uncover a wide array of ViT behaviors.

There are countless ways to analyze ViTs, so to guide this analysis we choose three major domains which correspond to the \textit{How}, \textit{What}, and \textit{Why} of ViTs. For the \textit{How}, we focus on \textit{how} ViTs process information through \textbf{Attention}. Multi-Headed Attention (MHA) layers are arguably the key element of ViTs, and they most distinguish them from CNNs. For the \textit{What}, we examine the \textbf{Features} of ViTs, as these are typically \textit{what} practitioners take away from them.
Finally for the \textit{Why}, we focus on
\textbf{Downstream Tasks}, which are \textit{why} we care about using ViTs.

\begin{figure}
    \centering
    \includegraphics[width=\linewidth]{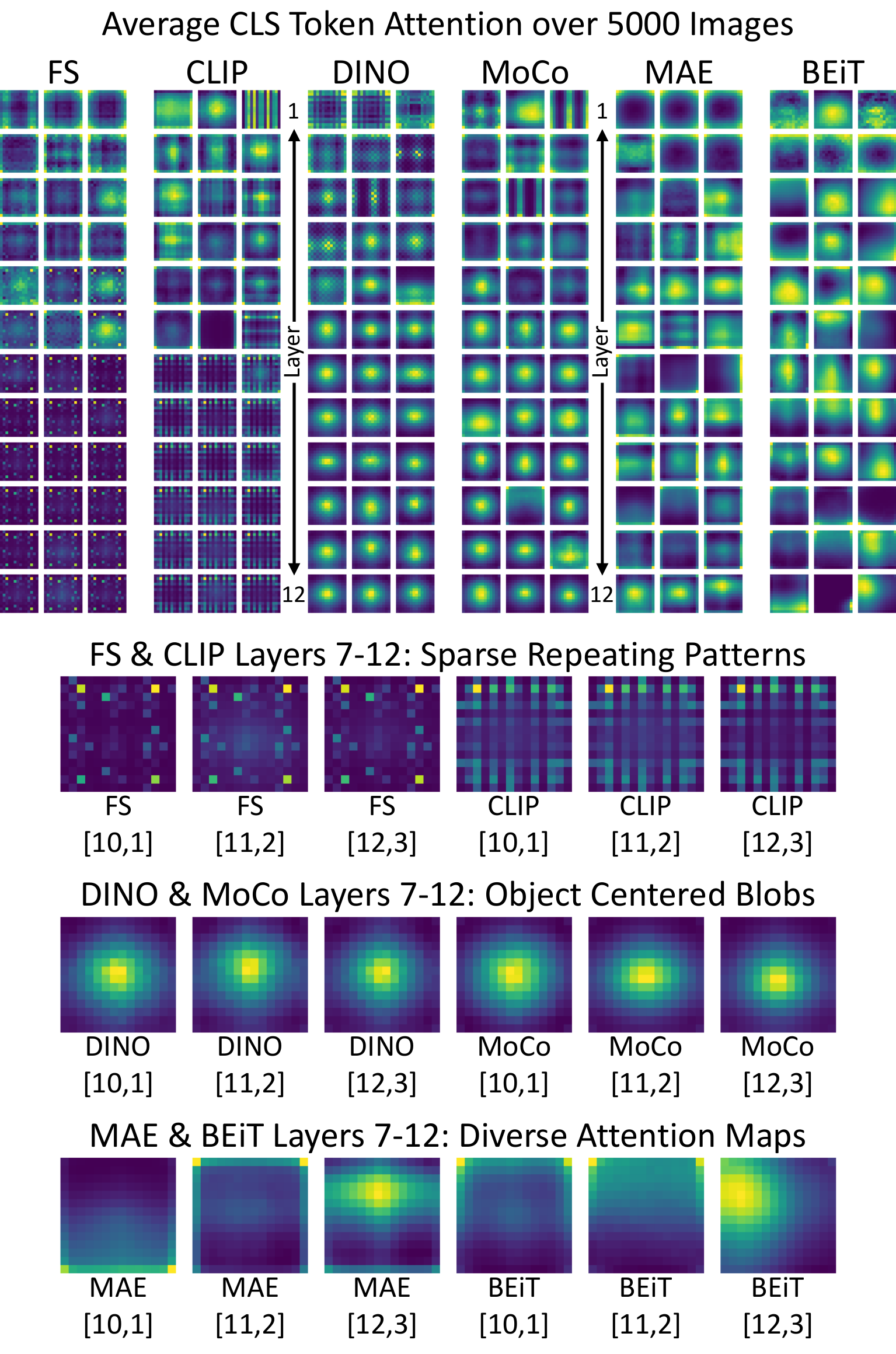}
    \vspace{-0.25in}
    \caption{\textbf{Clear differences in attention emerge in the mid-to-late layers under different supervision methods}. These plots show the attention maps of CLS tokens averaged over 5000 ImageNet images. Rows indicate layers and columns indicate heads. For brevity, we show only three heads per layer. The bracketed numbers in the lower half denote the layer and head.
    }
    \label{fig:att-grid-5k}
    \vspace{-0.2in}
\end{figure}

Our work unveils that a powerful aspect of the ViT architecture is its local-global dual nature, which plays a role in all three aspects of our analyses.
While standard CNNs are restricted to building representations hierarchically from local to global, in a ViT each token can attend to information from any other image region at any time. 
And unlike popular CNN modifications like Spatial Pyramids~\citep{grauman2005pyramid, lazebnik2006beyond, he2015spatial, lin2017feature} and top-down strategies ~\citep{shrivastava2016beyond, pinheiro2016learning, cao2015look}, ViTs have the freedom to decide when and where global information should be integrated. In this study, we show that the order and the relative ratio of local and global attention in ViTs varies dramatically based on the method of supervision. We also find clearly different trends in the allocation of attention in the mid-to-late layers of these networks, as highlighted in \Cref{fig:att-grid-5k}.
This local-global dual nature is also embedded into the structure and features of the ViT, which encodes both local spatial tokens and a non-local classifier (CLS) token throughout its entire depth.
We analyze the features of ViTs for both the CLS and spatial tokens, and assess how they align with semantics at the image, object, part, and pixel-level. We perform this analysis at every layer of the ViT to show the emergence of different levels of semantic information.
Finally, we assess ViTs on a number of local and global downstream tasks.

Overall, our contributions are:
\textbf{[1]} A detailed comparison of ViTs trained with six different methods, including both fully supervised and self-supervised training.
\textbf{[2]} A cross-cutting analysis spanning three major domains: Attention, Features, and Downstream Tasks.
\textbf{[3]} Multiple insights into the inner workings of ViTs to guide future development of ViT variants, training strategies, and applications.
    
In addition, we summarize some of our key observations about ViT behavior:
\textbf{[1]} The attention maps of explicitly supervised ViTs devolve into \textbf{Sparse Repeating Patterns} in the mid-to-late layers, but the quality of features continues to improve in these layers (\Cref{sec:att_vis}).
\textbf{[2]} All ViTs studied learn to use \textbf{Offset Local Attention Heads}, suggesting they are fundamentally necessary in ViTs  (\Cref{sec:att_olah}). To the best of our knowledge, no prior work has brought attention to this phenomenon.
\textbf{[3]} ViTs learn to process local and global information in different orders depending on their method of supervision (\Cref{sec:att_dist}).
\textbf{[4]} All ViTs studied differentiate salient foreground objects by the early-to-mid layers (\Cref{sec:att_iou}).
\textbf{[5]} Reconstruction-based self-supervised methods can learn semantically meaningful CLS representations, even when the CLS token is only a placeholder (\Cref{sec:cka_lb}, \ref{sec:cluster-cls}).
\textbf{[6]} Supervised method’s features are the most semantically rich, but contrastive self-supervised methods are comparable or even superior in some cases (\Cref{sec:cluster-cls}, \ref{sec:cluster-spatial}).
\textbf{[7]} For localized tasks, the best performance often comes from a mid-to-late layer (\Cref{sec:down-seg}).
\textbf{[8]} There is no single ``best" training method or layer for all downstream tasks (\Cref{sec:down-st}).
\section{Related Work}
\label{sec:related}

Previous works have attempted to understand the representation quality for both supervised and self-supervised training for Convolutional Neural Networks (CNNs).~\citep{bau2017network} focuses on understanding the concepts learned by individual neurons while~\citep{mu2020compositional} looks at explaining their compositionality in the case of supervised networks. Simultaneously, due to the popularity of self-supervised learning methods~\citep{caron2021emerging,he2020momentum,he2022masked,bao2021beit,chen2020simple,zbontar2021barlow,bachman2019learning,chen2020improved,hjelm2018learning,misra2020self,oord2018representation,tian2020contrastive} multiple works have analyzed these representations learned from no labels.
Under this umbrella,~\citep{van2021revisiting,cole2022does} tried to understand the effect of training data in terms of both the number and type of samples. Some works~\citep{wang2020understanding,wang2021understanding} analyze the alignment, separability, and uniformity of features while~\citep{purushwalkam2020demystifying} looks at invariance to augmentations like occlusion, illumination, and viewpoint change in the learned representation.~\citep{van2021benchmarking} looks at the downstream performance of self-supervised networks on fine-grained tasks. Finally,~\citep{kotar2021contrasting,grigg2021self,gwilliam2022beyond} analyze multiple self-supervised methods and compare their performance based on representation similarity and downstream task performance over multiple datasets along with comparisons to supervised methods.

Since the proliferation of ViTs, a number of works have tried to understand and explore the different properties of the representations learned by these networks. A few works \citep{paul2022vision,shao2021adversarial,zhou2022understanding} have analyzed the robustness of ViT features against corruptions, perturbations, distribution shifts, and adversarial examples while also analyzing the role of self-attention for robustness.~\citep{li2021benchmarking} benchmarks different pretrained ViTs as backbones for object detection.~\citep{cao2022understand} provides a theoretical understanding of how MAEs work while~\citep{sahiner2022unraveling} analyzes attention using convex duality.~\citep{touvron2022three} gives insights to train and use ViTs more efficiently.~\citep{park2022vision} gives a deeper understanding of how Multi-Headed Attention layers work while comparing and contrasting to how convolution layers behave in terms of the loss landscapes and low-pass/high-pass filtering.~\citep{raghu2021vision} compares fully supervised ViTs and ResNets in terms of the local and global information encoded at different depths, the role of skip connections, and the uniformity of representations.

All these prior works either examine the impact of supervision on CNNs or compare CNNs and ViTs trained with full supervision. Some recent and concurrent works have compared the properties of differently supervised ViTs, though typically focused on a particular task and only two methods of supervision at a time. \citep{amir2021deep} compares the properties of fully supervised and DINO ViT features in the context of dense feature descriptors, and \citep{aygun2022demystifying} further compares these two across several semantic correspondence tasks. \citep{ghiasi2022vision} compares fully supervised and CLIP ViTs through feature visualizations. To the best of our knowledge, we present what is to date the broadest and the most in-depth comparison of ViTs with varying supervision, including six different methods covering three supervision subcategories. Additionally, we propose new attention-based analysis methods along with evaluations on multiple downstream tasks focused on both local and global information.
\section{Experimental Design}
\label{sec:method}

\subsection{A Primer on Vision Transformers}

Vision Transformers (ViTs)~\citep{dosovitskiy2020image} are adapted from Transformers~\citep{vaswani2017attention} for the Natural Language Processing domain. A ViT consists of an array of tokens, each representing an image patch. 
In addition, most ViTs include an extra ``classifier" or ``CLS" token, which is connected to the task-specific output layers during training.
ViTs use Multi-Headed Attention (MHA) layers~\citep{vaswani2017attention}, which use a Query-Key-Value system that allows each token to attend to all other tokens with a variable intensity attention map. This is in stark contrast to the limited receptive fields of convolutions. These layers are ``multi-headed" because they repeat this process multiple times in parallel, allowing tokens to apply multiple attention strategies concurrently.
A ViT architecture includes multiple blocks, each with one MHA layer followed by a position-wise fully connected layer. Unlike CNNs, which usually get narrower in deeper layers, ViTs maintain the same ``width" (number of tokens) throughout. There are some transformer variants, like SWiN transformers~\citep{liu2021swin}, that introduce a narrowing width, but for our analyses, we focus on only traditional ViTs. Specifically, our primary analysis focuses on ViT-Base models with patch size $16\times16$ (ViT-B/16) and input size $224\times224$, which results in a $14\times14$ spatial token array. ViT-Base has 12 blocks and 12 attention heads per MHA layer.
In the Appendix, we provide additional results on a wider range of ViTs, including variations in architecture size and patch size.

\subsection{Methods of Supervision}
\label{sec:methods-of-supervision}

Although a large number of ViT training methods have been proposed in a short span of three years, many of the most popular methods can be loosely categorized into the following three groups. From each group, we select two representative models for in-depth analysis. We further discuss these models' details in \cref{sec:apdx-vits}.

\textbf{Explicit Supervision.}
These models are trained with an explicit objective that is defined either by human annotations or by labels derived from another source, like paired image captions. For this category, we use a Fully Supervised (FS) ViT pretrained on ImageNet21k and fine-tuned on ImageNet1k~\citep{steiner2021train, rw2019timm}, as well as a CLIP ViT~\citep{radford2021learning}.

\textbf{Self-Supervision (Contrastive).} 
Self-supervised learning methods broadly attempt to train a model through a pretext task that can be directly derived from the input data.
Among the more popular pretext tasks are contrastive learning methods~\citep{wu2018unsupervised, he2020momentum} which generally present a model with multiple augmented views of the same image alongside distractor views of other images. The model must learn to identify which of the views came from the same image. For this category, we select DINO~\citep{caron2021emerging} and MoCo-v3~\citep{chen2021empirical} which we denote simply as MoCo for the rest of this paper.

\textbf{Self-Supervision (Reconstruction).}
Another popular category of self-supervision is reconstruction methods, which train models to predict the missing content from masked or otherwise corrupted images.
We select MAE~\citep{he2022masked} and BEiT~\citep{bao2021beit} for this category. Note that MAE has a separate decoder which is discarded after pretraining, while BEiT's decoder is learned in the same ViT. This has a strong impact on the behavior of the later layers of BEiT.

\subsection{Datasets}

We study the ViTs on multiple datasets and downstream tasks. Unless otherwise specified, we use ImageNet-50~\citep{van2020scan}, a subset of ImageNet~\citep{deng2009imagenet} which narrows the dataset down to 50 representative categories. We sample 100 images per class to create a diverse collection of 5000 images.
We additionally use PartImageNet~\citep{he2021partimagenet} to measure Attention Saliency and part-level feature purity, as well as COCO~\citep{lin2014microsoft} to measure object-level feature purity. We use revisited~\citep{radenovic2018revisiting} Oxford~\citep{philbin2008lost} (ROxford5k) for evaluating image retrieval, DAVIS~\citep{pont20172017} for video segmentation, and SPair-71k~\citep{min2019spair} for keypoint correspondence.

\subsection{Proposed Analyses}

Our analysis is broadly divided into three domains covering the \textit{How}, \textit{What}, and \textit{Why} of ViTs:

\noindent{\textbf{\textit{How} ViTs process local/global information (Attention).}}
Do self-attention heads learn to operate in different ways depending on their method of training?
Are there distinctive modes of attention behavior?
How does supervision impact the processing order of local and global information?

\noindent{\textbf{\textit{What} we take away from ViTs (Features).}}
How do the final and intermediate representations of a ViT change depending on the method of supervision? Are these trends similar or different for CLS \vs spatial tokens?

\noindent{\textbf{\textit{Why} we use ViTs (Downstream Tasks).}}
Which forms of supervision are best suited for different downstream tasks? Which layers of a ViT produce features that are best for different local and global tasks?

\section{Attention Analysis}
\label{sec:att}

\begin{figure}
    \centering
    \includegraphics[width=\linewidth]{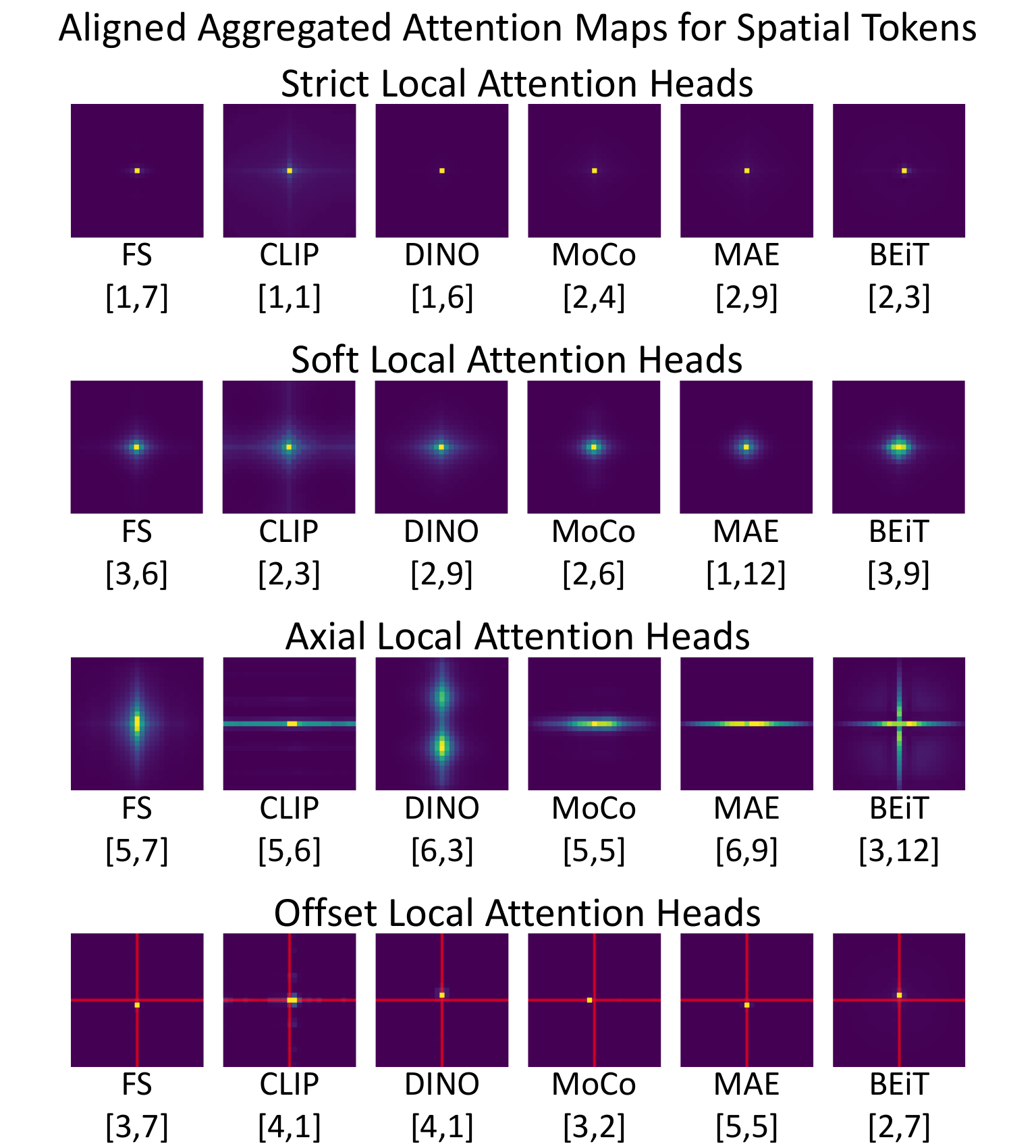}
    \vspace{-0.2in}
    \caption{\textbf{Multiple distinct forms of local attention exist.} We visualize spatial token attention using Aligned Aggregated Attention Maps, and highlight different types of local attention heads, including Strict, Soft, Axial, and Offset Local Attention Heads. In row 4 we draw the mid-lines in red as a visual aid.}
    \label{fig:att-grid-aligned-agg}
    \vspace{-1.5em}
\end{figure}

Multi-Headed Attention layers are one of the defining components of the Transformer architecture, and the attention maps they generate can give key insights into what is similar or different about ViTs trained through different methods.
We perform an in-depth examination of the self-attention maps of ViT-B/16 models at every layer. Through this study, we uncover a diverse range of attention head behavioral modalities.
Additional visualizations are provided in \Cref{sec:apdx-att} for a wide range of ViT variants.

\subsection{Attention Visualizations}
\label{sec:att_vis}

We start by examining the attention maps of the CLS tokens of each head and layer. To gain a comprehensive understanding of each head's behavior, we compute the average attention maps over 5000 ImageNet images, as shown in \Cref{fig:att-grid-5k}. For brevity, we display only three heads per layer, but complete plots can be found in \Cref{sec:apdx-att-vis}, along with additional visualizations for spatial token attention and individual input images.

One of the clearest differences can be seen by comparing the mid-to-final layers. For the contrastive self-supervised methods, DINO and MoCo, the attention maps tend to be centered blobs. These heads tend to focus on salient foreground objects, so these blobs simply reflect object-centered photography bias. For the reconstruction-based methods, MAE and BEiT, we see a more diverse group of attention maps. This is likely because these methods must reconstruct all image regions, and thus their attention in the final layers must be more diverse and cover more of the image. Finally, for the explicitly supervised methods, FS and CLIP, the mid-to-final layers do not focus on salient object regions and instead focus on \textbf{Sparse Repeating Patterns} with seemingly no spatial meaning. This occurs for both the CLS tokens and spatial tokens, and the patterns are repeated across both heads and layers.
We hypothesize that these patterns occur because the mid-to-late layers are no longer focused on parsing the scene structure, and instead are using their processing power to generate their final decisions for their respective tasks.
This phenomenon helps to explain why the attention maps of the later layers of fully-supervised ViTs are poorly suited for segmentation tasks, as was observed by~\citep{caron2021emerging}.

\subsection{Emergence of Offset Local Attention Heads}
\label{sec:att_olah}

It has been shown that ViTs use a mixture of short-range ``local" and long-range ``global" attention heads in any given layer \citep{dosovitskiy2020image,caron2021emerging}. To gain a better understanding of local attention, we propose a visualization strategy of \textbf{Aligned Aggregated Attention Maps (AAAMs)}. We extract all spatial token attention maps for 5000 ImageNet images, but before averaging them, we first realigned them so the current spatial token is always in the center of the array. Additional samples of AAAMs are provided in \Cref{sec:apdx-att-vis}.
Studying these aligned views reveals multiple forms of local attention, shown in \Cref{fig:att-grid-aligned-agg}. We find Strict Local Attention Heads, which attend almost completely to their own position, as well as Soft Local Attention Heads, which attend to a wider neighborhood around them. We also find Axial Local Attention Heads, which are elongated to attend to the local neighborhood along one or both spatial axes.

But perhaps the most noteworthy type of head we observe is the \textbf{Offset Local Attention Head}. These are heads that attend locally, but to a point or region offset from the current token in a vertical or horizontal direction. We find instances of Offset Local Attention Heads in all the models examined, suggesting they are fundamentally necessary for ViTs. To the best of our knowledge, ours is the first work to draw attention to this phenomenon. We believe that such heads are absolutely necessary because ViTs, unlike CNNs, do not have an easy built-in way to test if two features occur next to each other \textit{with a particular spatial arrangement}.
In a CNN, this type of check is naturally embedded into the convolution operator. But in a ViT, there is no inductive bias to induce such a check. For comparison, Soft Local Attention Heads are able to identify if a certain feature is near another feature, but they cannot identify their specific directional arrangement due to their symmetrical attention pattern.
The existence of Offset Local Attention Heads implies one possible path for improvement for the ViT architecture, possibly by adding a self-attention variant that introduces some implicit directional structure.

\subsection{Average Attention Distance}
\label{sec:att_dist}

\begin{figure}
    \centering
    
    \begin{subfigure}[b]{0.235\textwidth}
        \centering
        \includegraphics[width=\textwidth]{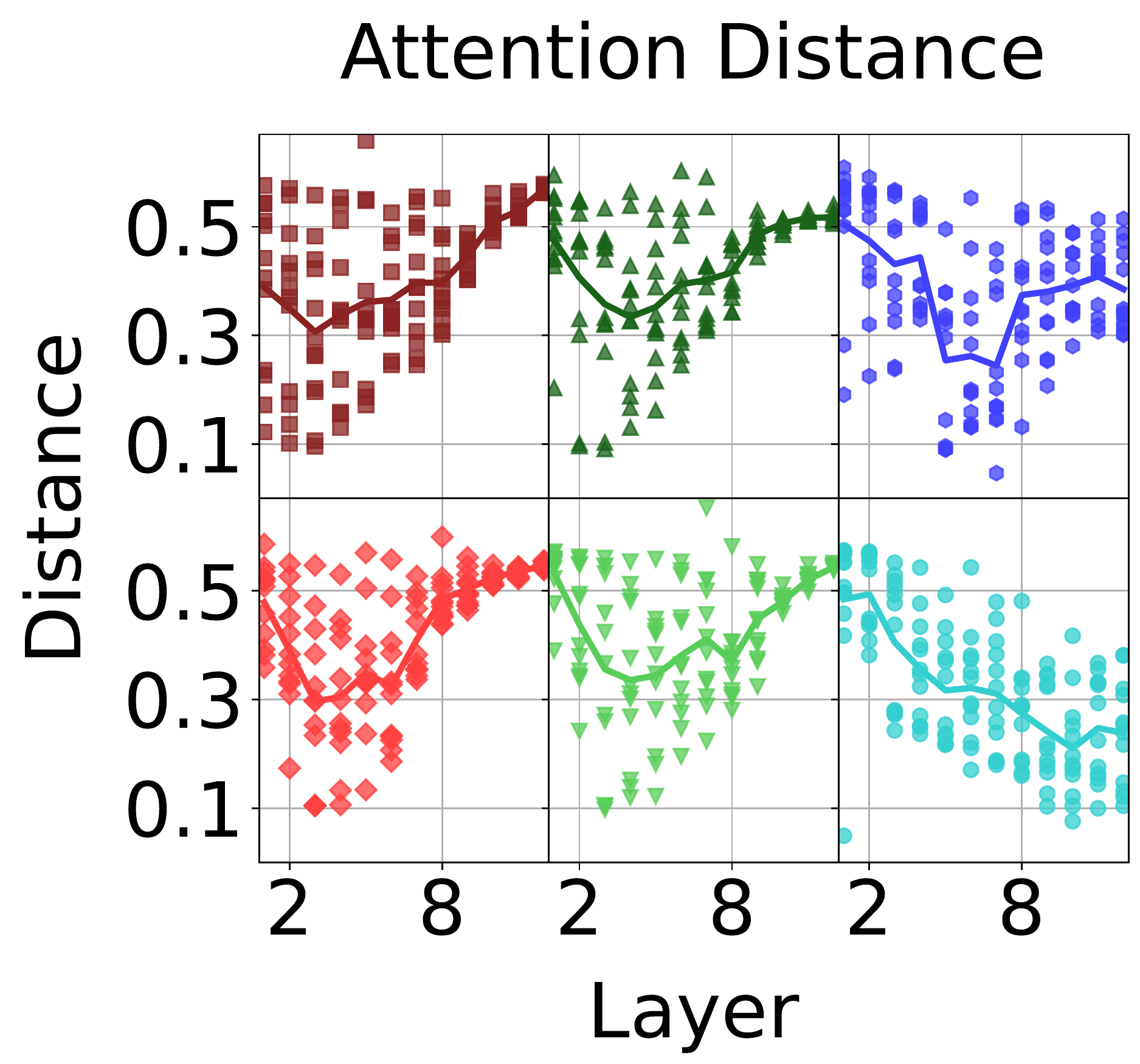}
        \label{fig:att-line-iou-cls-pin}
    \end{subfigure}
    \hfill
    \begin{subfigure}[b]{0.235\textwidth}
        \centering
        \includegraphics[width=\textwidth]{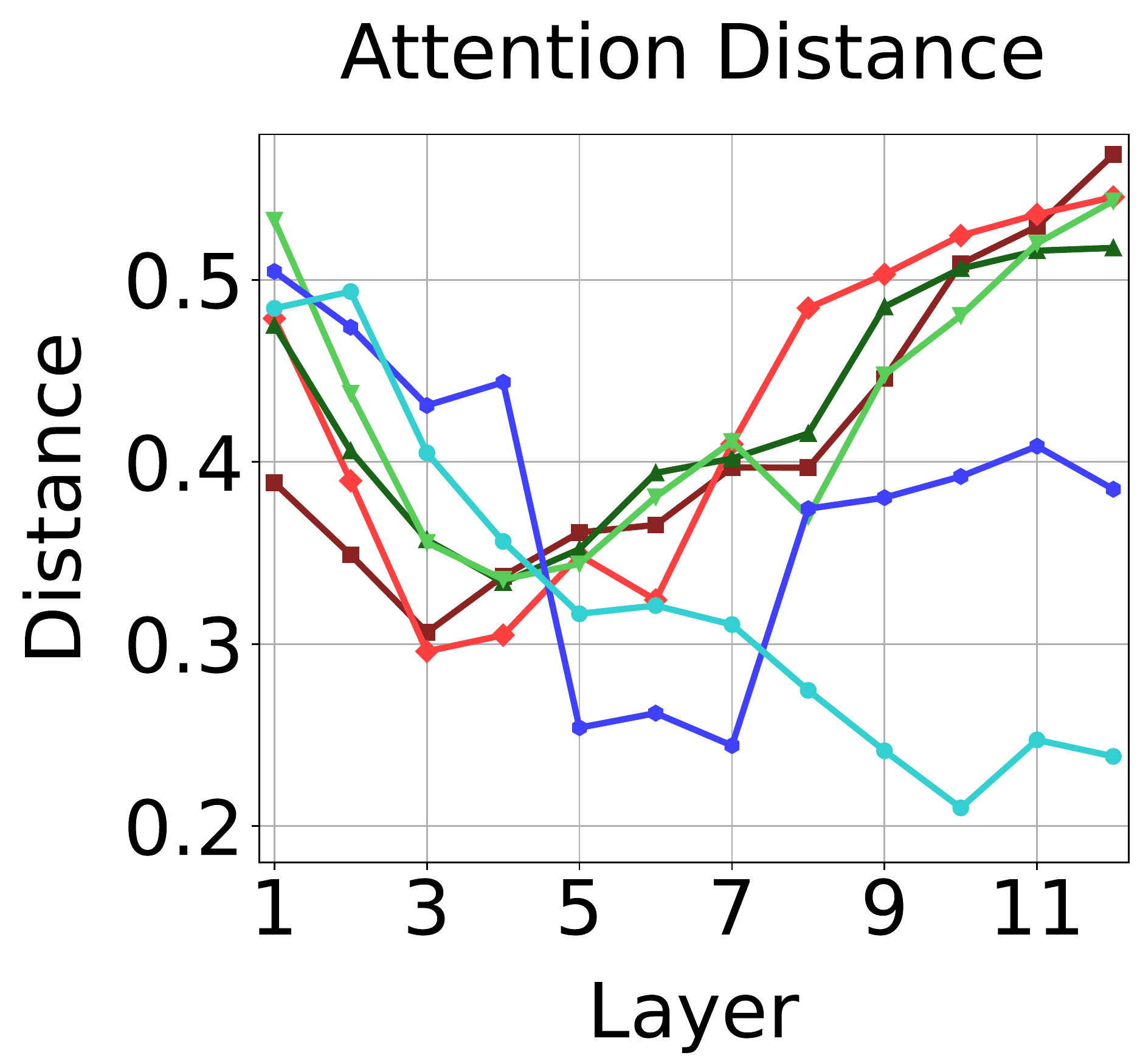}
        \label{fig:att-line-dist}
    \end{subfigure}

    \vspace{-10pt}
    \begin{subfigure}[b]{0.3806\textwidth}
        \centering
        \includegraphics[width=\textwidth]{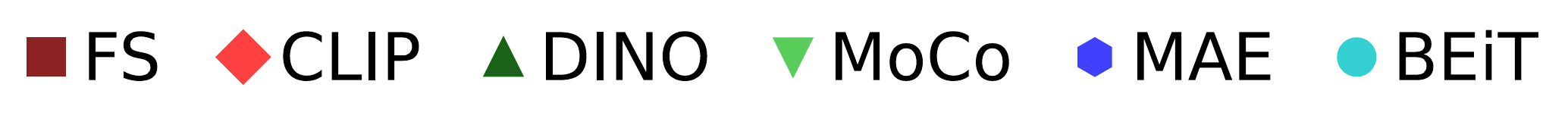}
    \end{subfigure}
    
    \vspace{-5pt}
    \caption{\textbf{Different methods of supervision lead to different orderings and ratios of local and global processing.}
    We show the Average Attention Distance of all ViT attention heads organized by layer (left), and the per-layer averages (right).}
    \label{fig:att-dist-plots}
\end{figure}
\begin{figure}
    \centering
    \begin{subfigure}[b]{0.235\textwidth}
        \centering
        \includegraphics[width=\textwidth]{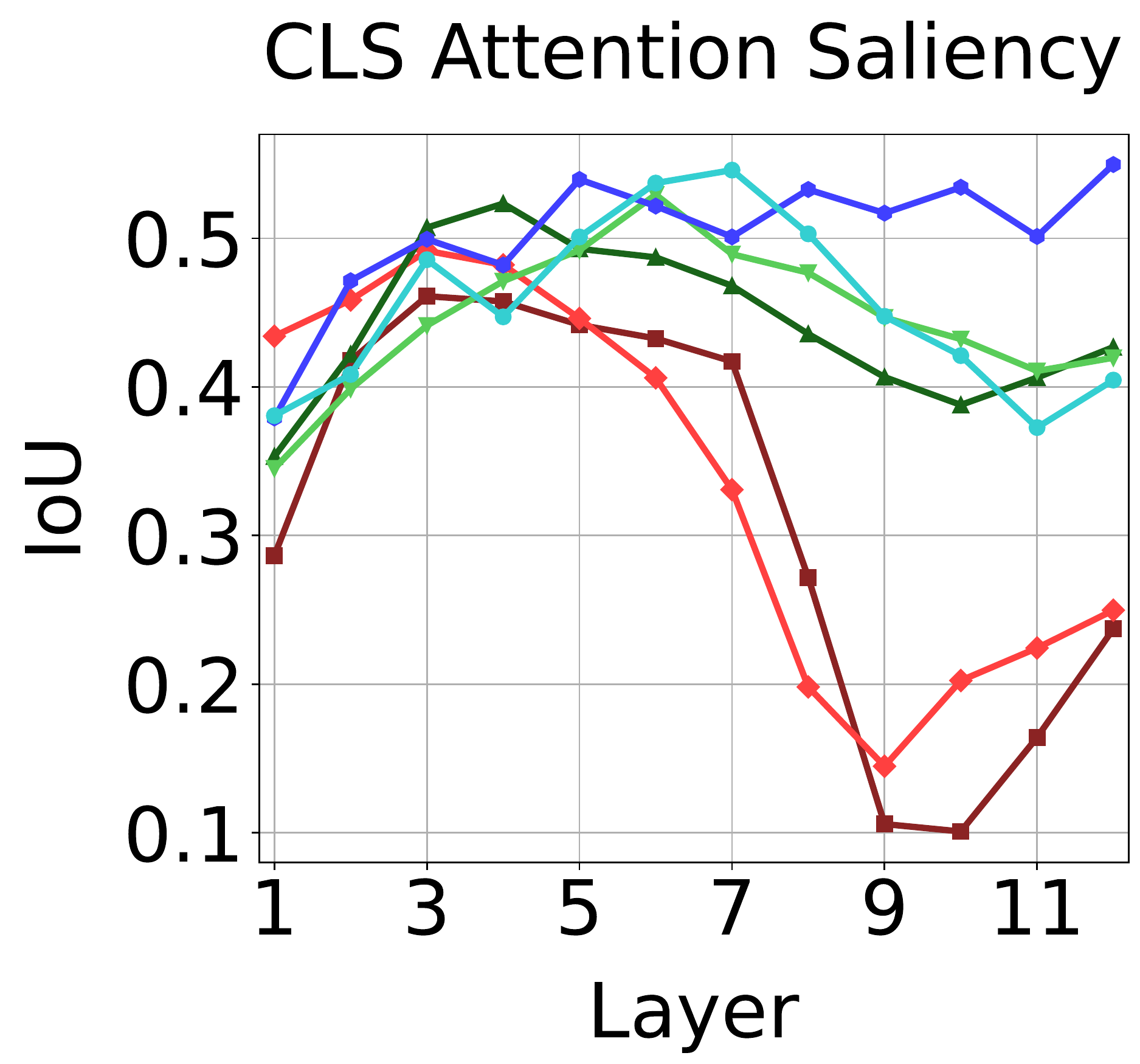}
    \end{subfigure}
    \hfill
    \begin{subfigure}[b]{0.235\textwidth}
        \centering
        \includegraphics[width=\textwidth]{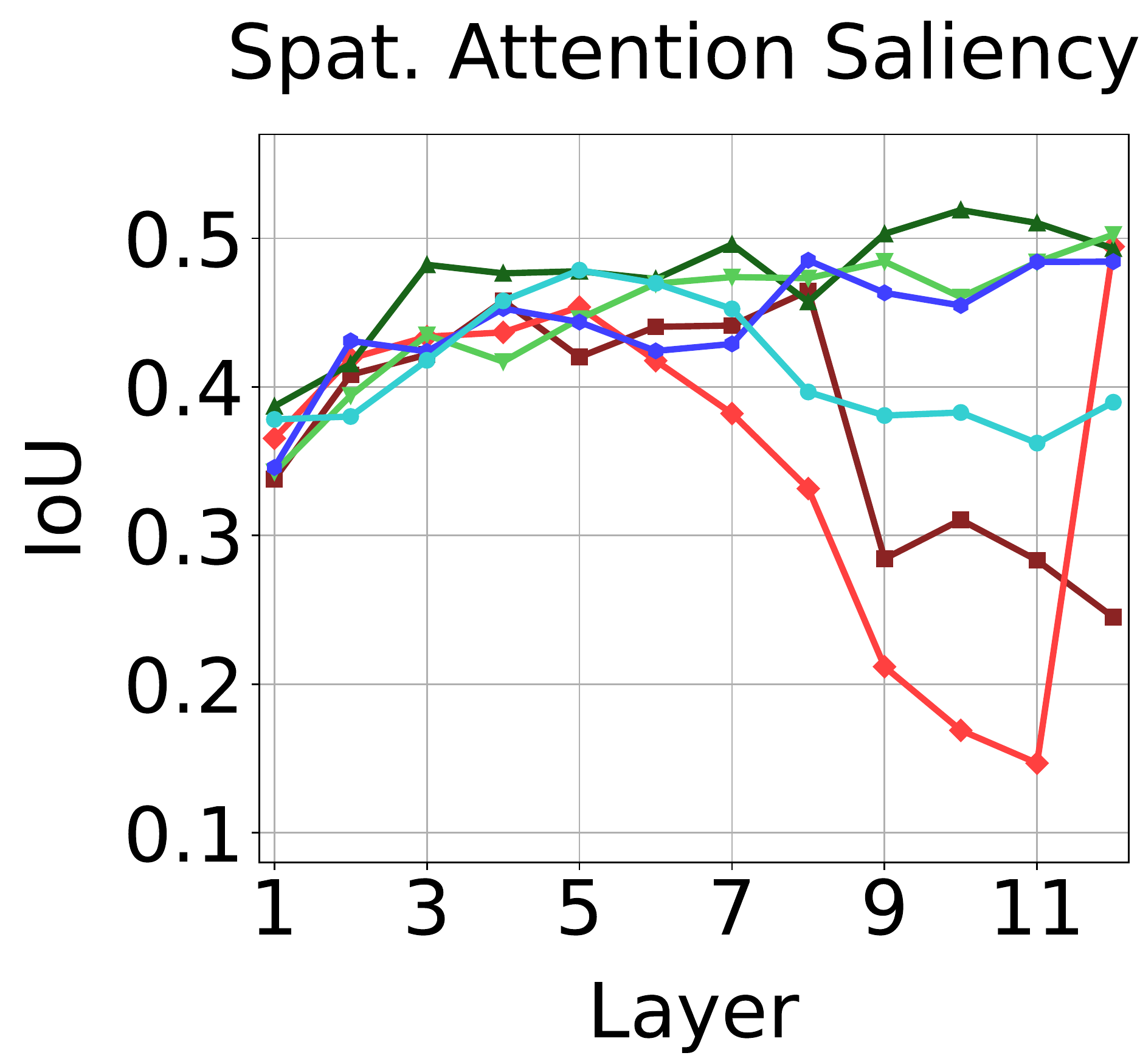}
    \end{subfigure}
    
    \begin{subfigure}{0.3806\textwidth}
        \centering
        \includegraphics[width=\textwidth]{plots/legends/legend_B-16.pdf}
    \end{subfigure}

    \vspace{-5pt}
    \caption{\textbf{Attention IoU with salient content plateaus early for all ViTs evaluated.} We calculate the alignment of ground-truth segmentation masks with CLS token attention maps (left) and the average of spatial token attention maps (right).}
    \label{fig:att-iou-plots}
    \vspace{-0.2in}
\end{figure}

We measure the Average Attention Distance \citep{dosovitskiy2020image, raghu2021vision} of each head to assess if particular heads have a short-range ``local" focus or a long-range ``global" focus. This metric is computed by measuring the distance from each spatial token to all other tokens and taking a weighted average using the attention map. We normalize the distances so the token grid is embedded on a $1\times1$ square.
\citep{raghu2021vision} observed that for a well-trained fully-supervised ViT, the early layers have a mixture of local and global attention heads, while the later layers have only global attention heads.

\Cref{fig:att-dist-plots} (left) shows the Average Attention Distances of all heads organized by layer and model. Like \citep{raghu2021vision}, we see that most layers use a mixture of local and global attention heads, however, we also find that the ordering of local and global processing varies greatly with the supervision type.
FS, CLIP, DINO, and MoCo all use exclusively global attention heads in the last layers, but the reconstruction-based methods MAE and BEiT use a diverse range of heads in their later layers.
\Cref{fig:att-dist-plots} (right) compares the combined Average Attention Distances at the per-layer level.
In all models, we observe a greater number of global attention heads in the initial layers, followed by decreased distances around layers 3-6. This result is again in contrast to \citep{raghu2021vision}.
The behaviors diverge in the mid-to-late layers. For the models trained with explicit or contrastive supervision, the Attention Distance trends upward in the later layers. For the reconstruction-based methods, the Average Attention Distances stay lower.
These results show that, unlike CNNs, ViTs can learn a variable local/global processing order depending on the training method used.

\subsection{Attention Alignment with Salient Content}
\label{sec:att_iou}

One of the most desirable (and exploitable) features of DINO is that the CLS token attention maps of the last layer tend to be well-aligned with salient foreground objects~\citep{caron2021emerging}. Several methods propose to use DINO attention maps, feature maps, or a combination of the two to generate segmentations in a self-supervised manner~\citep{hamilton2022unsupervised, wang2022self, simeoni2021localizing}.
We conduct a quantitative analysis of this property at all layers of the ViTs, both to measure the usefulness of masks and to assess how early the ViTs differentiate salient object regions.
Like \citep{caron2021emerging}, we threshold the CLS token attention masks keeping 60\% of the total attention mass. We then compute the Intersection over Union (IoU) of said masks with ground-truth segmentations from PartImageNet~\citep{he2021partimagenet}. As an alternative to CLS token attention, we also extract masks using the average of spatial token attention maps.
We present results for the single ``best" head per layer in \Cref{fig:att-iou-plots}.

We see a clear drop in FS and CLIP mask IoU around the middle of the network, which directly corresponds to the emergence of the Sparse Repeating Patterns observed in \Cref{sec:att_vis}. We also find that the IoUs plateau around layers 3-6 for all networks. This demonstrates that ViT models already have a solid understanding of foreground/background separation by the middle layers. While the later attention maps of FS and CLIP are much worse than their self-supervised counterparts, their early-to-mid layers are more comparable.
We find that MoCo, MAE, and BEiT can all produce attention maps with IoUs that are comparable with DINO.
In addition, we see that the average of spatial tokens produces maps that are comparable with the CLS token, and for CLIP the IoU increases greatly in the final layer.

\section{Feature Analysis}
\label{sec:feat}

\begin{figure}
    \centering
    \begin{subfigure}{0.23\textwidth}
        \centering
        \includegraphics[width=\textwidth]{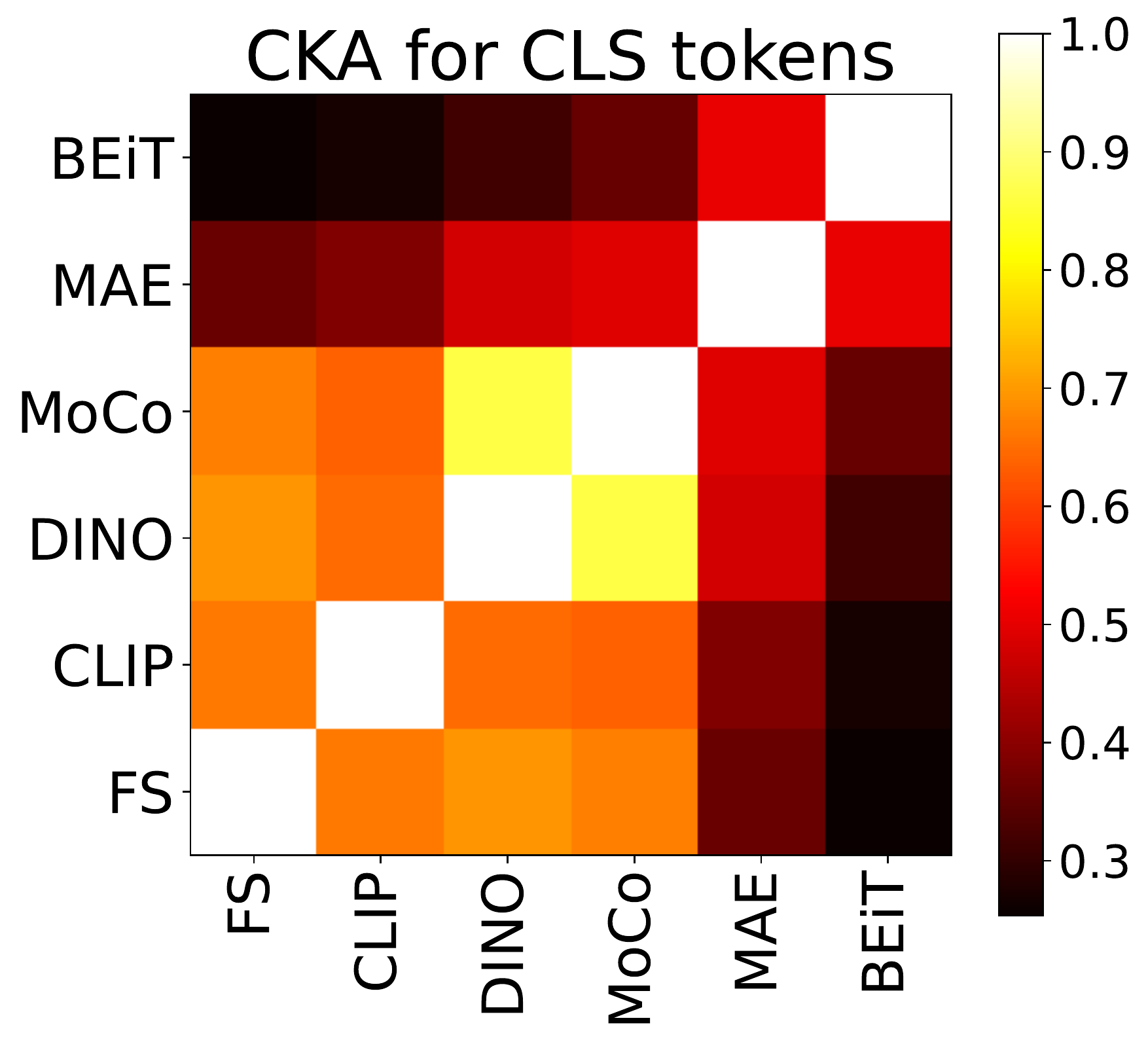}
    \end{subfigure}
    \begin{subfigure}{0.23\textwidth}
        \centering
        \includegraphics[width=\textwidth]{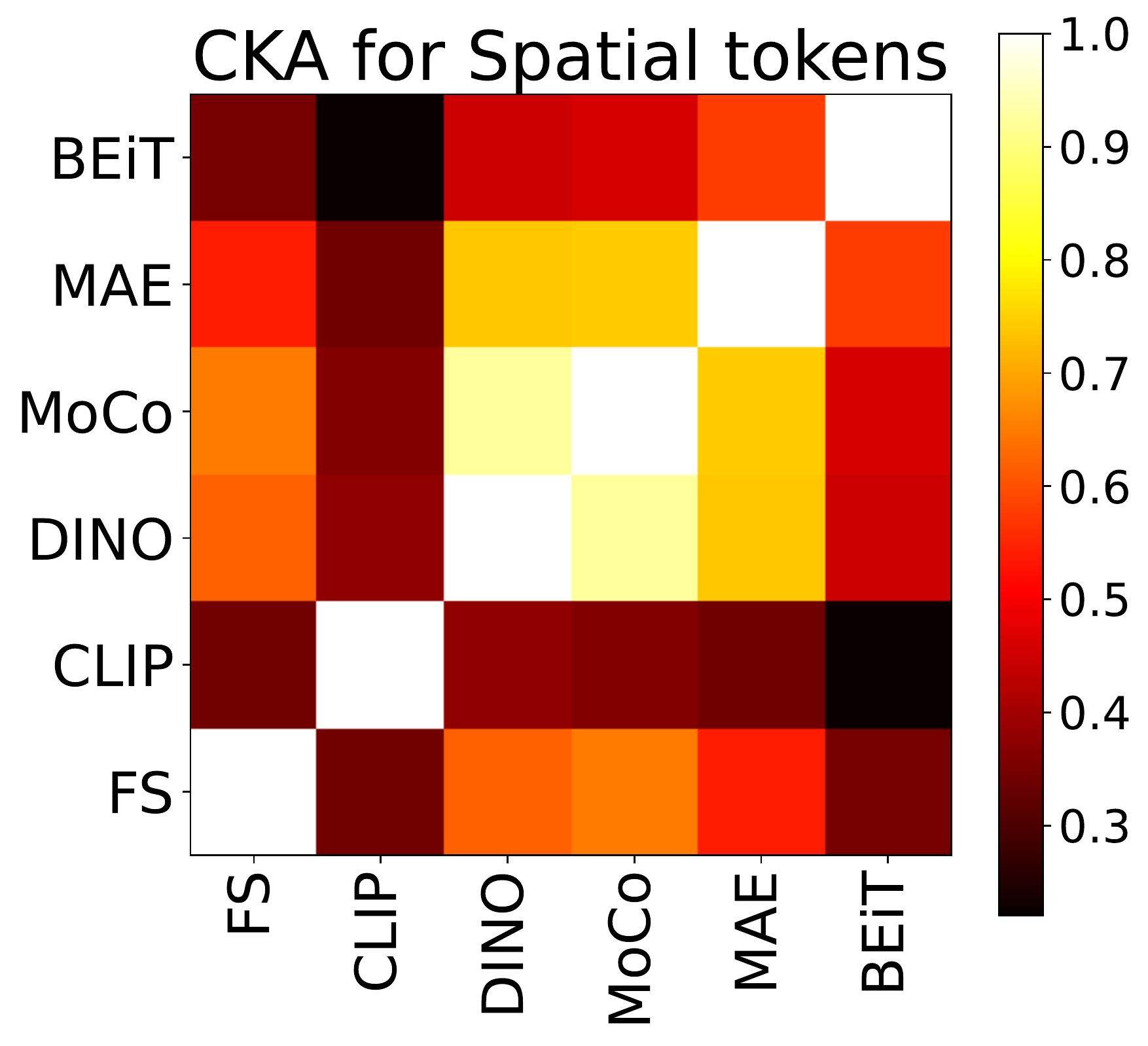}
    \end{subfigure}
    \vspace{-0.05in}
    \caption{CKA similarity between final layer features of different ViTs for their CLS tokens (left) and spatial tokens (right).}
    \label{fig:cka-lastlayer}
    \vspace{-0.15in}
\end{figure}

In this section, we directly compare ViT features across models and layers using Centered Kernel Alignment (CKA)~\citep{cortes2012algorithms,kornblith2019similarity}.
We also study unsupervised clustering performance to compare global and local semantic information in the learned representations. We present addition analysis in \Cref{sec:residual} focused on ViT residual connections.

\subsection{Last Block Feature Comparisons}
\label{sec:cka_lb}

Comparing representations is non-trivial due to varying feature sizes, large feature representations, and lack of alignment between them. To overcome this, we use batched Centered Kernel Alignment (CKA)~\citep{cortes2012algorithms, kornblith2019similarity, nguyen2020wide} which can align features and compute a similarity score. We compare the last layer outputs for each model.

\Cref{fig:cka-lastlayer} (left) shows that the CLS token representations are usually similar for similar supervision strategies (explicit, contrastive, reconstruction).
The contrastive methods, MoCo and DINO, show very high similarity to each other, indicating that the CLS token encodes the same type of information for both these methods. There is also an increased level of similarity between the explicitly supervised methods, FS and CLIP, and the contrastive methods.
Interestingly, we see that MAE has as high a similarity with DINO and MoCo as it does with BEiT. 
This result is surprising because MAE's CLS token has no explicit training objective or loss, and the way these approaches are trained is very different. This presents evidence that training autoencoders with a high masking percentage indeed forces the model to learn image-level semantics.

In \Cref{fig:cka-lastlayer} (right), we look at the similarity of the last-layer spatial token representations. Unlike the CLS token representations, CLIP and FS have low similarity in their spatial representations.
The self-supervised methods DINO, MAE, and MoCo show a high level of similarity to each other, and a lower level of similarity to BEiT. 
MoCo and DINO show the highest similarity due to their similar kind of self-supervision. Once again, MAE has a high similarity to MoCo and DINO despite their very different supervision.

\subsection{Feature Clustering for Global Semantics}
\label{sec:cluster-cls}

Through this analysis, we aim to test how well the learned CLS and spatial token representations encode global (image-level) semantic information at every layer.
We extract the CLS token features from the end of each block for 5000 ImageNet images, and we generate k-Means cluster assignments with $k=50$. We present results for cluster purity measured with respect to ground truth image labels, but additional clustering metrics are also presented in \Cref{sec:apdx-feat-clust}.
For the spatial tokens, we follow the same process except we average-pool over all positions before clustering. We also compute a random chance score by replacing the ViT features with Gaussian random noise.

\begin{figure}
    \centering
    \begin{subfigure}[b]{0.235\textwidth}
        \centering
        \includegraphics[width=\textwidth]{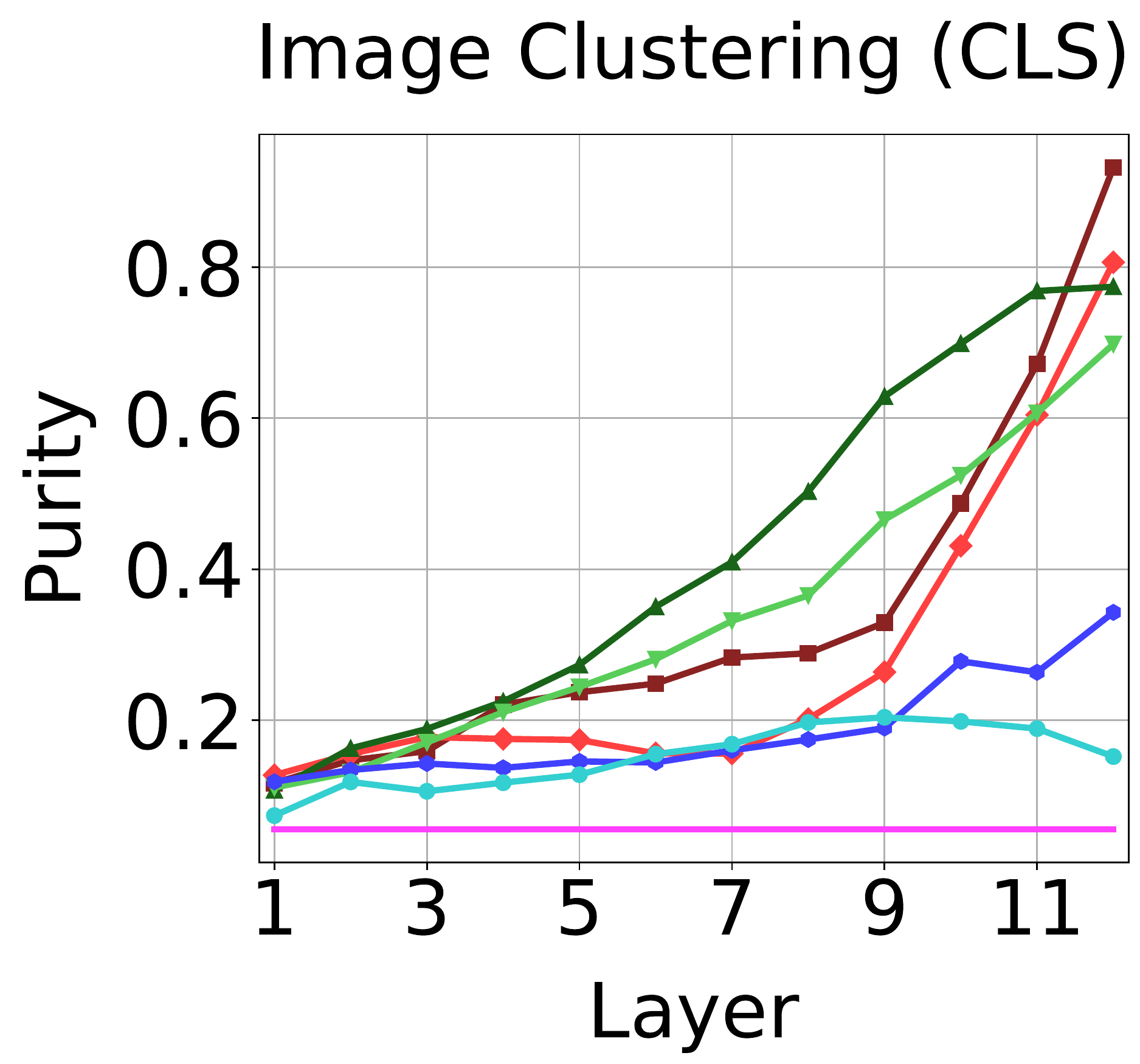}
    \end{subfigure}
    \hfill
    \begin{subfigure}[b]{0.235\textwidth}
        \centering
        \includegraphics[width=\textwidth]{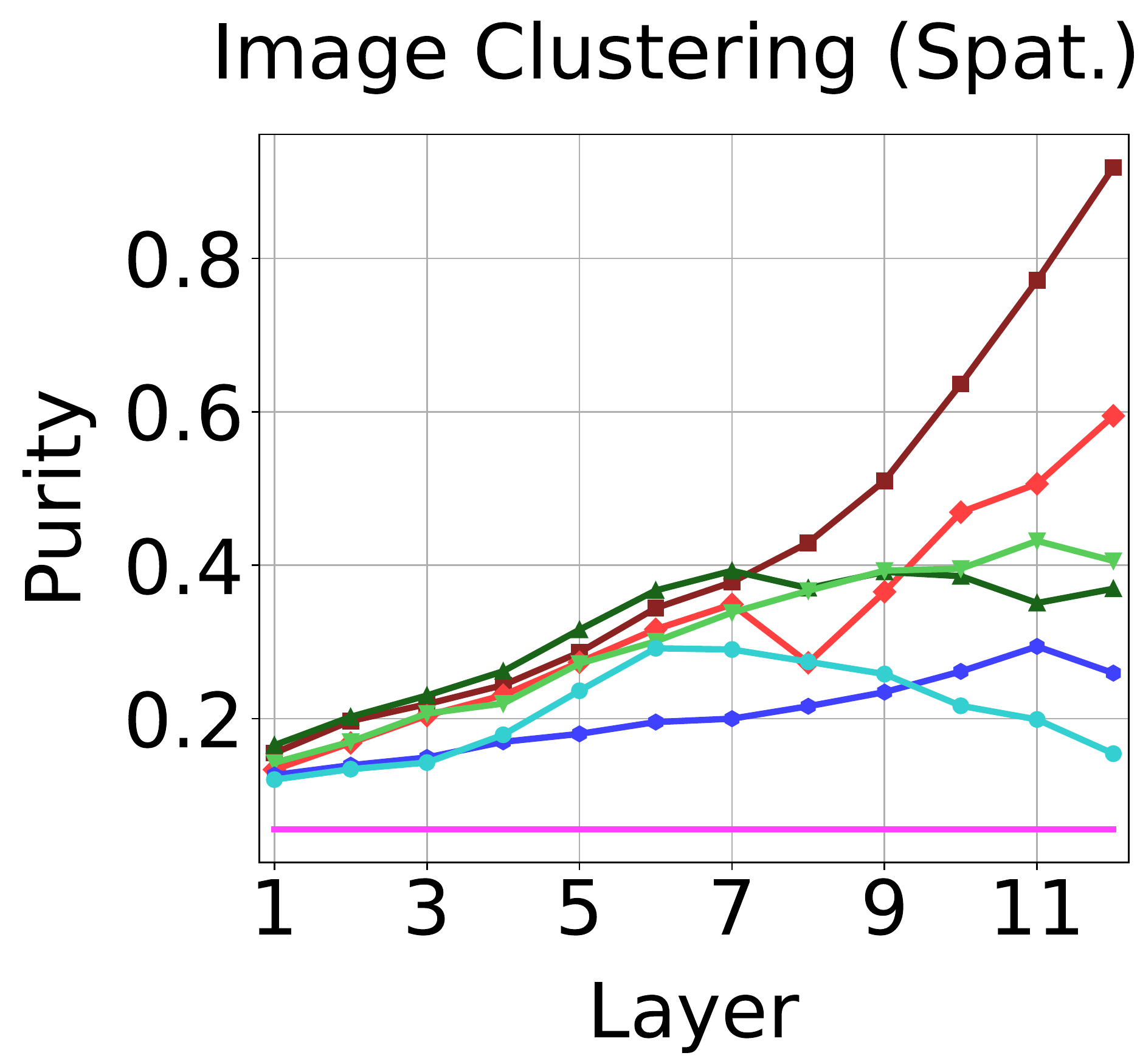}
    \end{subfigure}
    
    \begin{subfigure}[b]{0.477\textwidth}
        \centering
        \includegraphics[width=\textwidth]{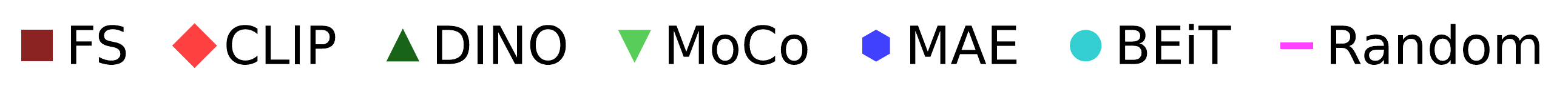}
    \end{subfigure}
    \vspace{-0.2in}
    \caption{Clustering purity analysis with image-level labels in ImageNet-50 for CLS features (left) and average-pooled spatial token features (right).}
    \label{fig:cluster-purity-global}
\end{figure}

For CLS token features, shown in \Cref{fig:cluster-purity-global} (left), cluster purity improves with depth with the exception of the last layers of BEiT.
This is likely because the last layers of BEiT serve as a task-specific decoder, unlike MAE, where the decoder is separate and discarded after pretraining.
Unsurprisingly, FS achieves the best cluster purity, followed by CLIP.
The contrastive methods, DINO and MoCo, achieve scores close to the explicitly supervised methods. The reconstruction-based methods, MAE and BEiT, have the lowest cluster purity, but they are still above random chance, which again indicates that they do learn to encode some image-level semantic information in their CLS tokens.
Also, we find that semantic information emerges earlier for DINO and MoCo.
For the spatial token features, shown in \Cref{fig:cluster-purity-global} (right), the cluster purity of FS rises earlier compared with the FS CLS token.
This suggests that the FS spatial tokens do more work gathering semantic information in the early layers. For all other ViTs, the spatial feature purity is lower in the final layers, but is comparable in layers 1-7.

\begin{figure}
    \centering
    \begin{subfigure}[b]{0.235\textwidth}
        \centering
        \includegraphics[width=\textwidth]{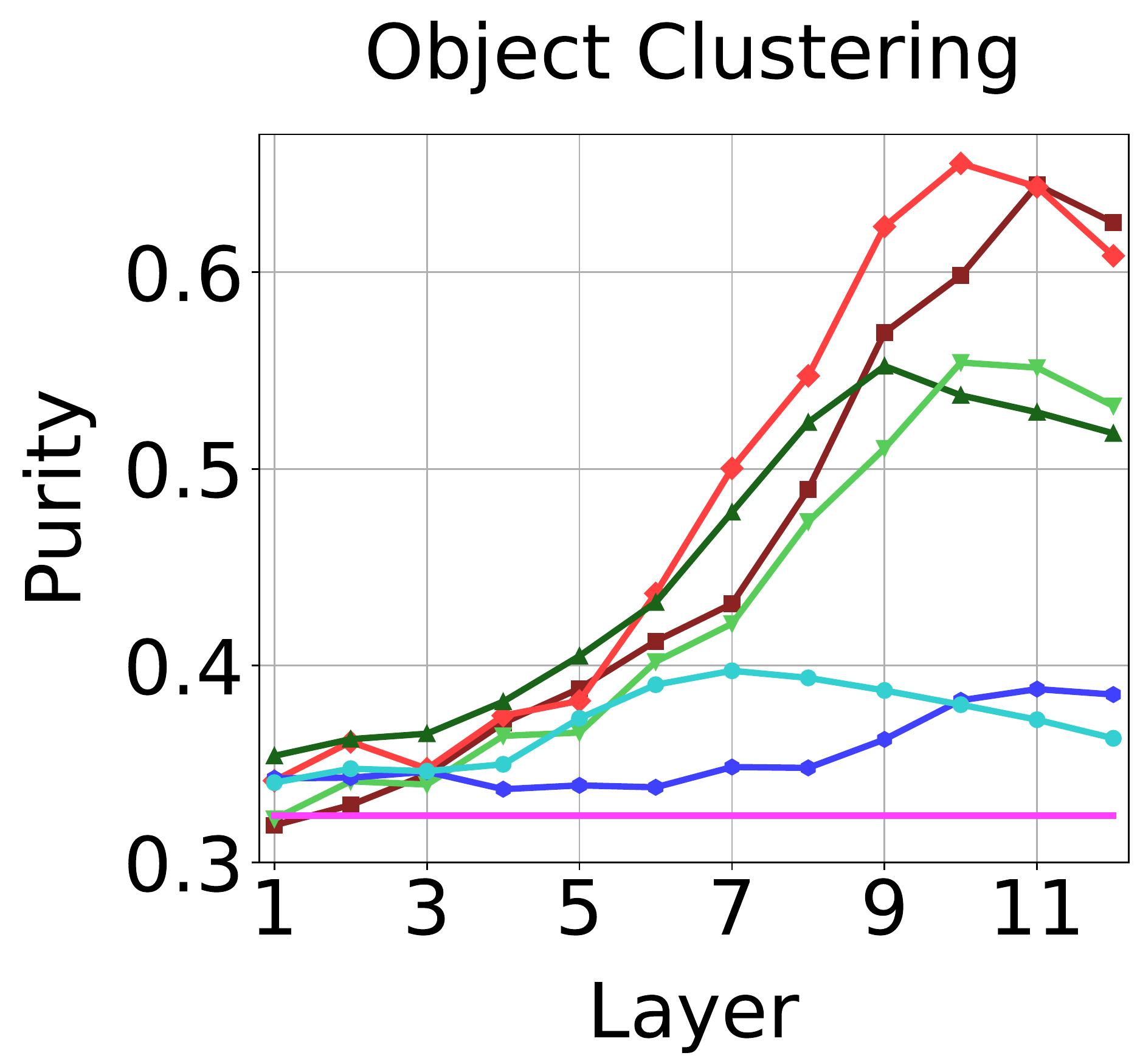}
    \end{subfigure}
    \hfill
    \begin{subfigure}[b]{0.235\textwidth}
        \centering
        \includegraphics[width=\textwidth]{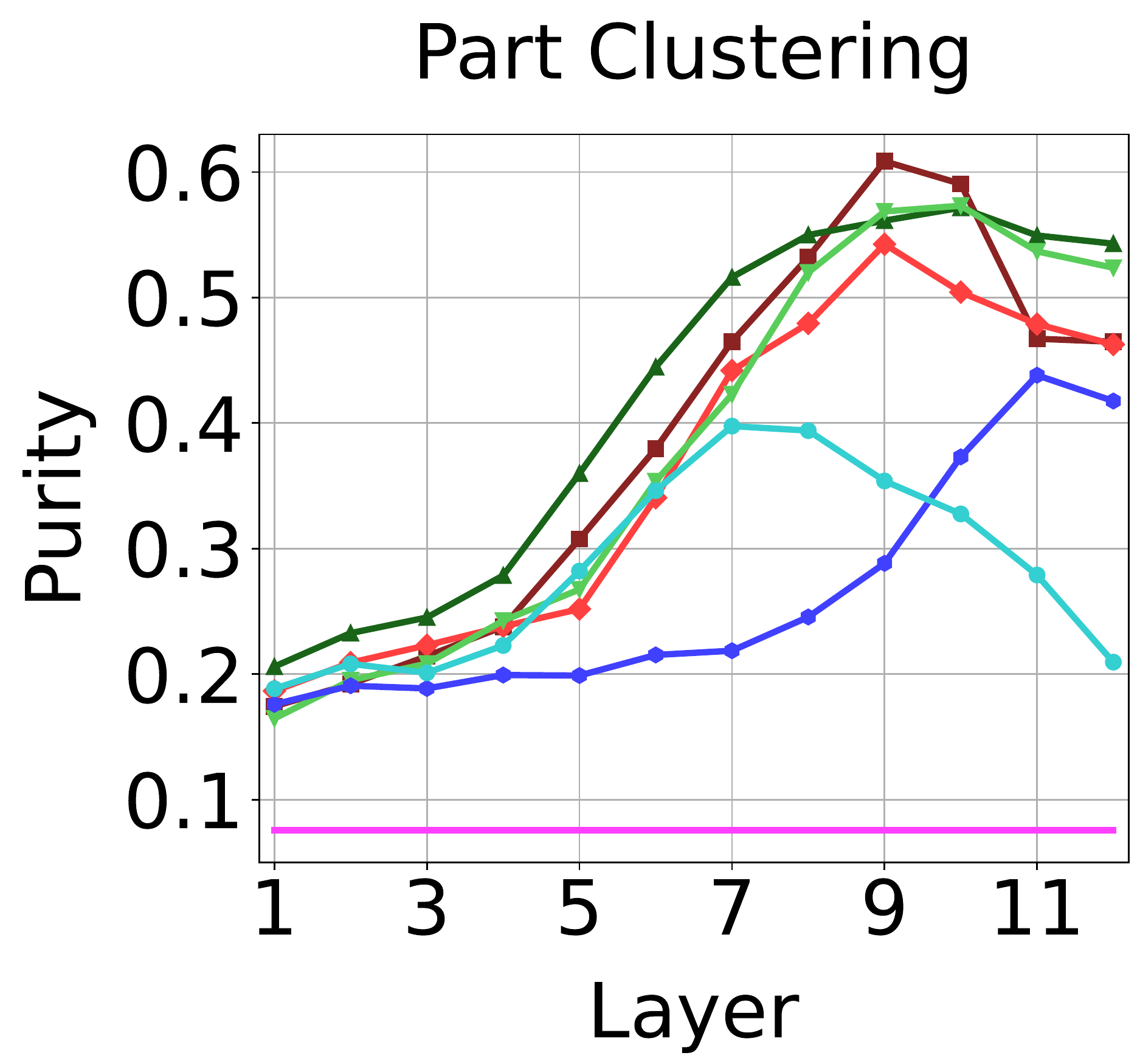}
    \end{subfigure}
    \begin{subfigure}[b]{0.477\textwidth}
        \centering
        \includegraphics[width=\textwidth]{plots/legends/legend_with_random_B-16.pdf}
    \end{subfigure}
    \vspace{-0.2in}
    \caption{Clustering purity of spatial token features gathered at the object-level in COCO (left) and part-level in PartImageNet (right).}
    \label{fig:cluster-purity-local}
    \vspace{-0.2in}
\end{figure}

\subsection{Feature Clustering for Local Semantics}
\label{sec:cluster-spatial}

We next measure how well the spatial token features differentiate salient image content at the object or part-level using COCO~\citep{lin2014microsoft} and PartImageNet~\citep{he2021partimagenet} respectively. We use a tiling strategy to extract a denser array of features, which we detail in \Cref{sec:apdx-dense-feat}.
Using ground truth segmentation masks, we extract and average the features of the tokens overlapping with the masks. This generates a collection of object-level or part-level features which we cluster just like \Cref{sec:cluster-cls}.
The results for object-level features are shown in \Cref{fig:cluster-purity-local} (left). We see that the supervised methods CLIP and FS have the highest feature purity by far, followed by the contrastive methods DINO and MoCo. The purity is much lower for the reconstruction methods MAE and BEiT.
For part-level features, shown in Figure \ref{fig:cluster-purity-local} (right), FS achieves the best purity, but the contrastive methods are very competitive in this case, surpassing CLIP completely. In addition, while still being the lowest scoring, MAE and BEiT are much more competitive at the part-level.
Like the image-level purity, the object and part-level feature purity tend to improve with depth, but the purity peaks early around layers 9 to 11. The peak for BEiT is even earlier, likely due to its integrated decoder.

\section{Downstream Task Analysis}
\label{sec:down}
Finally, we analyze the performance of these models on downstream tasks that can be performed directly without any fine-tuning or training.
We follow the evaluation protocols of~\citep{caron2021emerging, jabri2020space} for $k$-NN classification, image retrieval, and video object segmentation. We also perform keypoint correspondence as a more local-focused task. Again we compute random chance scores by replacing all ViT features with Gaussian noise.

\subsection{Global Tasks}

\begin{figure}
    \centering
    \begin{subfigure}[b]{0.235\textwidth}
        \centering
        \includegraphics[width=\textwidth]{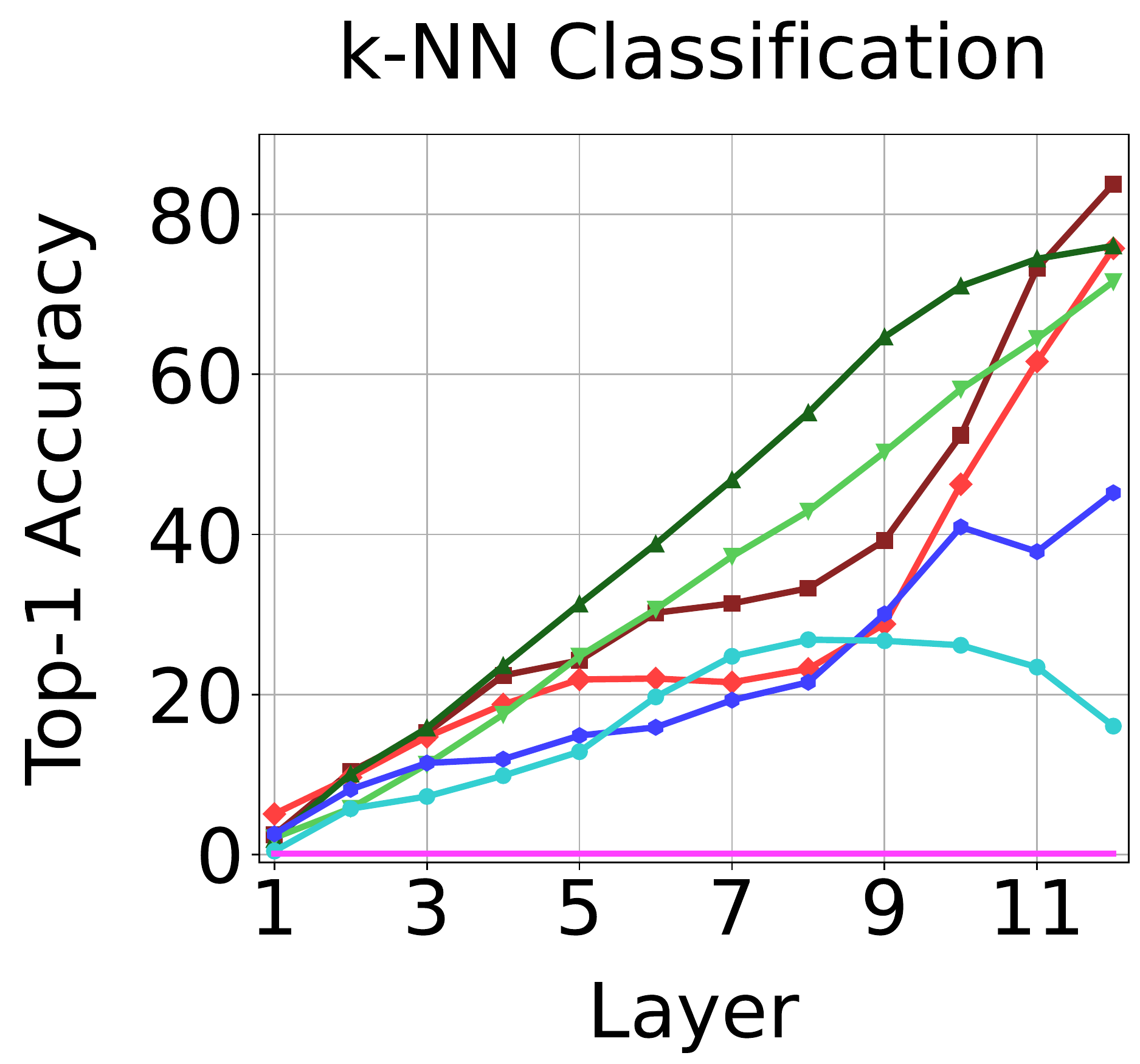}
    \end{subfigure}
    \hfill
    \begin{subfigure}[b]{0.235\textwidth}
        \centering
        \includegraphics[width=\textwidth]{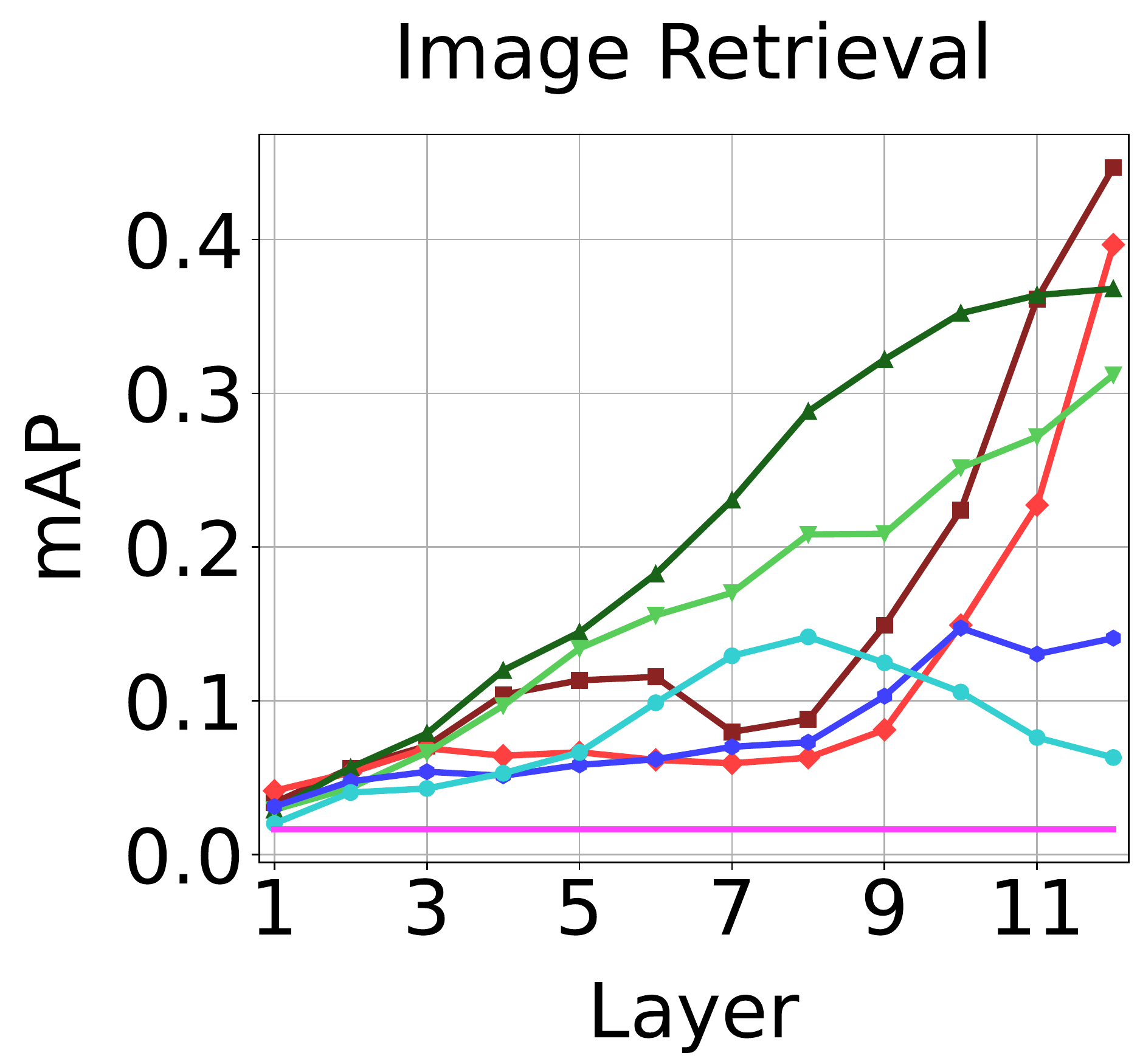}
    \end{subfigure}
    
    \begin{subfigure}[b]{0.477\textwidth}
        \centering
        \includegraphics[width=\textwidth]{plots/legends/legend_with_random_B-16.pdf}
    \end{subfigure}
    \vspace{-0.2in}
    \caption{Global (image-level) downstream task analysis using the CLS token. We present $k$-NN classifier Top-1 Accuracy on ImageNet (left) and image retrieval mAP on ROxford5k (right).}
    \label{fig:global-tasks}
    \vspace{-0.1in}
\end{figure}

\textbf{ImageNet Classification.}
We perform $k$-Nearest Neighbor ($k$-NN) image classification on ImageNet~\citep{deng2009imagenet} with $k=20$. We use the CLS token features from each network and assign the label for a test sample based on the training set features and labels.
As can be seen from \Cref{fig:global-tasks} (left), FS performs the best as it has been trained to classify the same dataset.
DINO and MoCo follow a similar trend as \Cref{sec:cluster-cls} and better encode semantic information in the earlier layers.
FS and CLIP also follow a similar trend where their performance shoots up in the last few layers. It is also interesting to see how MAE and BEiT, for which the CLS tokens have no explicit objective, do better than chance, although MAE is considerably better than BEiT.

\textbf{Image Retrieval.}
Similar to $k$-NN classification, we utilize the CLS token representation for retrieval. We evaluate on ROxford5k~\citep{radenovic2018revisiting} for the Medium (M) split and report the Mean Average Precision (mAP).
In \Cref{sec:apdx-downstream} we also report results for the Hard (H) split and the RParis6k~\citep{radenovic2018revisiting} dataset, which follow similar trends.
The results, shown in \Cref{fig:global-tasks} (right), align closely with those for $k$-NN Classification on ImageNet.
FS performs the best followed by CLIP and then DINO and MoCo, and finally by MAE and BEiT with the lowest performance. We hypothesize that the local/global crops used in DINO training help it perform competitively in these global tasks.

\begin{figure}
    \centering
    \begin{subfigure}{0.235\textwidth}
        \centering
        \includegraphics[width=\textwidth]{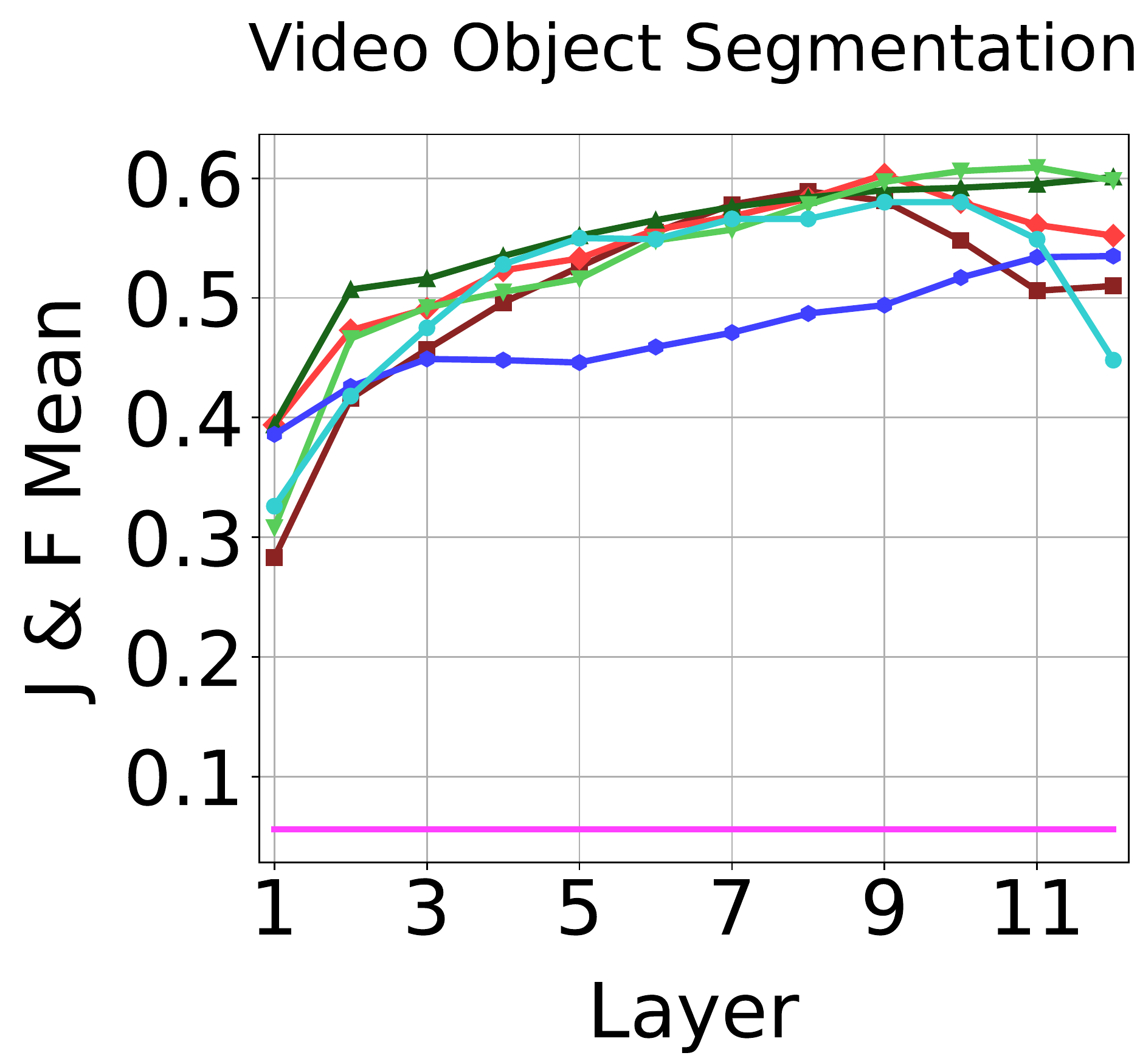}
    \end{subfigure}
    \hfill
    \begin{subfigure}{0.235\textwidth}
        \centering
        \includegraphics[width=\textwidth]{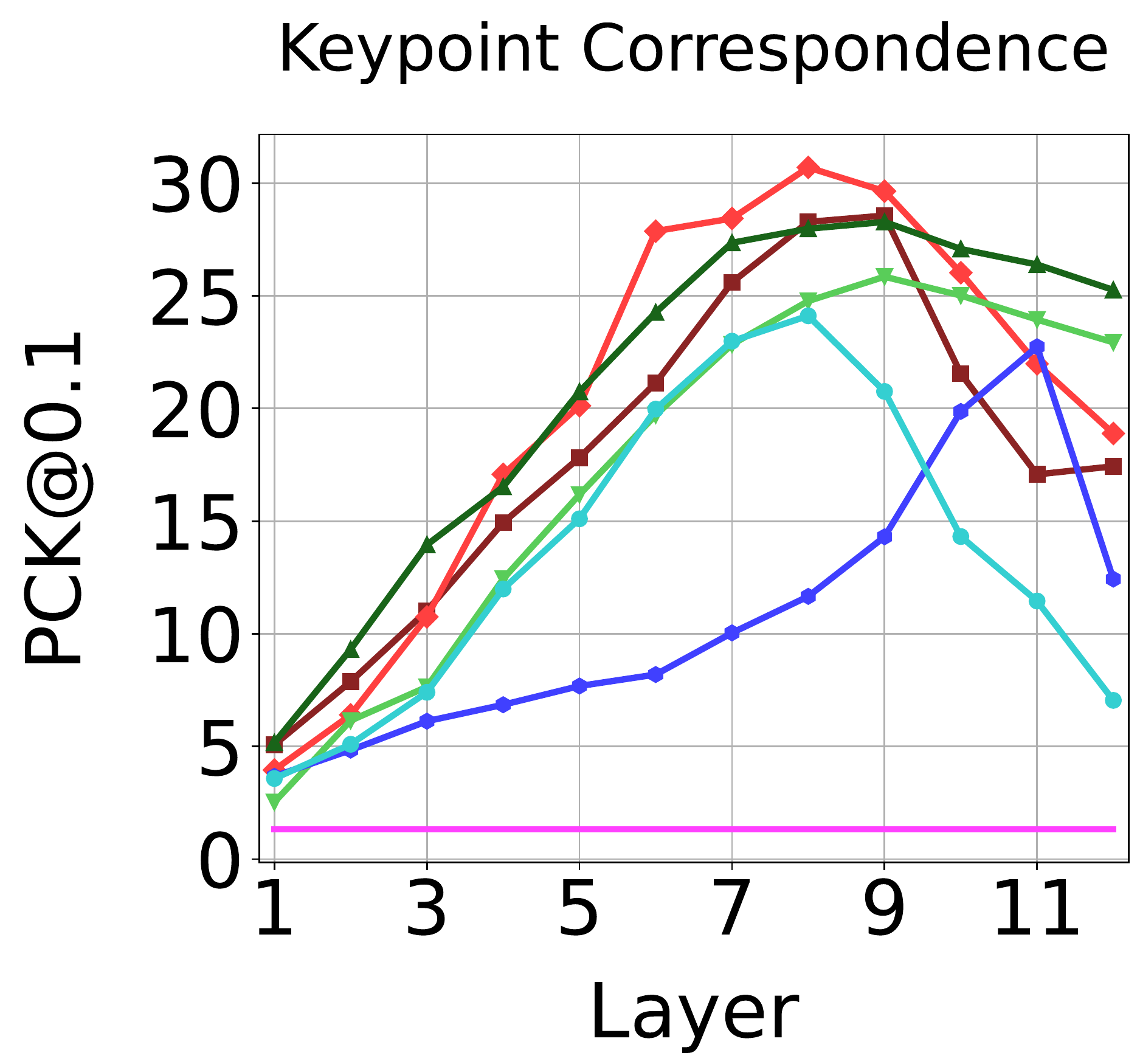}
    \end{subfigure}
    
    \begin{subfigure}{0.477\textwidth}
        \centering
        \includegraphics[width=\textwidth]{plots/legends/legend_with_random_B-16.pdf}
    \end{subfigure}
    \vspace{-0.2in}t
    \caption{Local (pixel-level) downstream task analysis using the dense spatial token features. We perform DAVIS video segmentation (left) and SPair-71k keypoint correspondence (right).}
    \label{fig:local-tasks}
    
    \vspace{-0.2in}
\end{figure}

\subsection{Local Tasks}

\textbf{DAVIS Segmentation Propagation.}
\label{sec:down-seg}
DAVIS Segmentation Propagation is a dense prediction-based video localized task where frame-by-frame features are used to propagate the first frame segmentation mask to subsequent frames.
Like \Cref{sec:cluster-spatial}, we use a tiling-based dense feature extraction strategy. Results are shown in \Cref{fig:local-tasks} (left).
The contrastive techniques of DINO and MoCo perform the best while FS and CLIP, which are more image-level approaches, face a drop in performance towards the later layers.
The local reconstruction-based methods, MAE and BEiT, are also much more competitive in this task.
These results show that for purposes of constructing highly descriptive local features, contrastive methods like DINO and MoCo and reconstructive methods like BEiT can surpass features trained with explicit image-level supervision.

\textbf{Keypoint Correspondence.}
\label{sec:down-kp}
We choose keypoint correspondence as an additional local-focused downstream task.
Given an image with annotated keypoints, the model must predict the position of corresponding keypoints in a paired image with similar content.
We use the SPair-71k~\citep{min2019spair} dataset and follow the evaluation protocol of \citep{amir2021deep} and report the Percentage of Correct Keypoints (PCK)~\citep{yang2012articulated}.
The results are summarized in \Cref{fig:local-tasks} (right). CLIP excels at this task, closely followed by both FS and DINO. Meanwhile, MoCo, BEiT, and MAE are all very competitive also. The position of the best layer varies significantly, from 8 for CLIP and BEiT to 11 for MAE.

\begin{table}[!t]
 \setlength{\cmidrulewidth}{0.01em}
\renewcommand{\tabcolsep}{8pt}
\renewcommand{\arraystretch}{1.1}
\caption{Best performance for each ViT on each downstream task with the corresponding best layer in parenthesis.}
\centering
\footnotesize
\resizebox{\linewidth}{!}{
\begin{tabular}{@{}lcccc@{}}
\toprule
 \textbf{Model} &  \multicolumn{4}{c}{\textbf{Task Performance (Best Performing Layer)}} \\
 \cmidrule[\cmidrulewidth](l){2-5}

 Dataset & ImageNet & ROxford5k (M) & Davis & SPair-71k\\
 Metric & Top-1↑ & mAP↑ & J and F Mean↑ & PCK@0.1↑\\
 \midrule
 
 FS	& \textbf{83.79 (12)} & \textbf{0.45 (12)} &  0.59 (8) &  28.56 (9)\\
CLIP & 75.75 (12) &  0.40 (12) & 0.60 (9) & \textbf{30.70 (8)}\\
DINO & 76.06 (12) & 0.37 (12) & 0.60 (12) & 28.28 (9)\\
MoCo & 71.59 (12) & 0.31 (12) & \textbf{0.61 (11)} &  25.85 (9)\\
MAE	& 45.19 (12) & 0.15 (10) & 0.54 (12) & 22.74 (11)\\
BEiT & 26.84 (8) & 0.14 (8) & 0.58 (9) & 24.11 (8)\\
Random & 0.10 & 0.02 & 0.06 & 1.32 \\
\bottomrule
\end{tabular}
  }
\label{tab:best_performance}
\vspace{-1em}
\end{table}

\subsection{Summary of Downstream Tasks}
\label{sec:down-st}

We summarize the best results for all downstream tasks in \Cref{tab:best_performance}. We denote the best-performing layers in parenthesis.
These results show that ViTs with different supervision methods \textbf{[1]} peak at different layers, and \textbf{[2]} perform best at different tasks.
In image-level tasks like $k$-NN classification and retrieval, usually, the last layer works the best. For localized tasks like keypoint correspondence and video object segmentation, most models' peak performance happens a few layers before the last one. This shows that always picking the last layer output is not optimal.
\section{Conclusion}
\label{sec:conc}
In this work, we have performed an in-depth comparison of ViTs trained through different methodologies by examining their attention patterns, learned representations, and downstream task performance. We review some of the key findings of our analyses. First, different methods of supervision lead to ViTs that process local and global information in different orders. All ViTs have heads that align well with salient image content, but for the explicitly supervised models, the late-layer attention maps change into Sparse Repeating Patterns. In addition, all ViTs examined have learned to use Offset Local Attention Heads in multiple layers.
While explicitly supervised ViTs have the most semantically rich representations at the image level, contrastive methods are competitive, and reconstruction-based methods can also learn meaningful CLS token representations even though said token is a placeholder and has no explicit supervision. Finally, there is no single best model for all the downstream tasks, and the best layer to extract representations from also varies greatly by task and model, so one should not simply take the last layer representation.
ViTs have shown a great deal of potential, and we expect they will become more widely used in the coming years. We hope these insights can help with the future development of losses and architectures for Vision Transformers.

\begin{small}
\myparagraph{Acknowledgements.} This project was partially funded by DARPA SAIL-ON (W911NF2020009), DARPA SemaFor (HR001119S0085), and DARPA GARD (HR00112020007) programs. We would like to thank our collegues Vatsal Agarwal, Matthew Gwilliam, Alex Hanson, Pulkit Kumar, Saketh Rambhatla, and Gaurav Shrivastava for their feedback on this work.
\end{small}

{\small
\bibliographystyle{ieee_fullname}
\bibliography{egbib}

\begin{thebibliography}{10}\itemsep=-1pt

\bibitem[Amir et~al.(2022)]{amir2021deep}
Shir Amir, Yossi Gandelsman, Shai Bagon, and Tali Dekel.
\newblock Deep vit features as dense visual descriptors.
\newblock {\em ECCVW What is Motion For?}, 2022.

\bibitem[Ayg{\"u}n and Mac~Aodha(2022)]{aygun2022demystifying}
Mehmet Ayg{\"u}n and Oisin Mac~Aodha.
\newblock Demystifying unsupervised semantic correspondence estimation.
\newblock In {\em Computer Vision--ECCV 2022: 17th European Conference, Tel
  Aviv, Israel, October 23--27, 2022, Proceedings, Part XXX}, pages 125--142.
  Springer, 2022.

\bibitem[Bachman et~al.(2019)]{bachman2019learning}
Philip Bachman, R~Devon Hjelm, and William Buchwalter.
\newblock Learning representations by maximizing mutual information across
  views.
\newblock {\em Advances in neural information processing systems}, 32, 2019.

\bibitem[Bao et~al.(2021)]{bao2021beit}
Hangbo Bao, Li Dong, and Furu Wei.
\newblock Beit: Bert pre-training of image transformers.
\newblock {\em arXiv preprint arXiv:2106.08254}, 2021.

\bibitem[Bau et~al.(2017)]{bau2017network}
David Bau, Bolei Zhou, Aditya Khosla, Aude Oliva, and Antonio Torralba.
\newblock Network dissection: Quantifying interpretability of deep visual
  representations.
\newblock In {\em Proceedings of the IEEE conference on computer vision and
  pattern recognition}, pages 6541--6549, 2017.

\bibitem[Cao et~al.(2015)]{cao2015look}
Chunshui Cao, Xianming Liu, Yi Yang, Yinan Yu, Jiang Wang, Zilei Wang, Yongzhen
  Huang, Liang Wang, Chang Huang, Wei Xu, et~al.
\newblock Look and think twice: Capturing top-down visual attention with
  feedback convolutional neural networks.
\newblock In {\em Proceedings of the IEEE international conference on computer
  vision}, pages 2956--2964, 2015.

\bibitem[Cao et~al.(2022)]{cao2022understand}
Shuhao Cao, Peng Xu, and David~A Clifton.
\newblock How to understand masked autoencoders.
\newblock {\em arXiv preprint arXiv:2202.03670}, 2022.

\bibitem[Caron et~al.(2020)]{caron2020unsupervised}
Mathilde Caron, Ishan Misra, Julien Mairal, Priya Goyal, Piotr Bojanowski, and
  Armand Joulin.
\newblock Unsupervised learning of visual features by contrasting cluster
  assignments.
\newblock {\em Advances in neural information processing systems},
  33:9912--9924, 2020.

\bibitem[Caron et~al.(2021)]{caron2021emerging}
Mathilde Caron, Hugo Touvron, Ishan Misra, Herv{\'e} J{\'e}gou, Julien Mairal,
  Piotr Bojanowski, and Armand Joulin.
\newblock Emerging properties in self-supervised vision transformers.
\newblock In {\em Proceedings of the IEEE/CVF International Conference on
  Computer Vision}, pages 9650--9660, 2021.

\bibitem[Chen et~al.(2020)]{chen2020simple}
Ting Chen, Simon Kornblith, Mohammad Norouzi, and Geoffrey Hinton.
\newblock A simple framework for contrastive learning of visual
  representations.
\newblock In {\em International conference on machine learning}, pages
  1597--1607. PMLR, 2020.

\bibitem[Chen et~al.(2020)]{chen2020improved}
Xinlei Chen, Haoqi Fan, Ross Girshick, and Kaiming He.
\newblock Improved baselines with momentum contrastive learning.
\newblock {\em arXiv preprint arXiv:2003.04297}, 2020.

\bibitem[Chen et~al.(2021)]{chen2021empirical}
Xinlei Chen, Saining Xie, and Kaiming He.
\newblock An empirical study of training self-supervised vision transformers.
\newblock In {\em Proceedings of the IEEE/CVF International Conference on
  Computer Vision}, pages 9640--9649, 2021.

\bibitem[Cole et~al.(2022)]{cole2022does}
Elijah Cole, Xuan Yang, Kimberly Wilber, Oisin Mac~Aodha, and Serge Belongie.
\newblock When does contrastive visual representation learning work?
\newblock In {\em Proceedings of the IEEE/CVF Conference on Computer Vision and
  Pattern Recognition}, pages 14755--14764, 2022.

\bibitem[Cortes et~al.(2012)]{cortes2012algorithms}
Corinna Cortes, Mehryar Mohri, and Afshin Rostamizadeh.
\newblock Algorithms for learning kernels based on centered alignment.
\newblock {\em The Journal of Machine Learning Research}, 13:795--828, 2012.

\bibitem[Cubuk et~al.(2020)]{cubuk2020randaugment}
Ekin~D Cubuk, Barret Zoph, Jonathon Shlens, and Quoc~V Le.
\newblock Randaugment: Practical automated data augmentation with a reduced
  search space.
\newblock In {\em Proceedings of the IEEE/CVF conference on computer vision and
  pattern recognition workshops}, pages 702--703, 2020.

\bibitem[Deng et~al.(2009)]{deng2009imagenet}
Jia Deng, Wei Dong, Richard Socher, Li-Jia Li, Kai Li, and Li Fei-Fei.
\newblock Imagenet: A large-scale hierarchical image database.
\newblock In {\em 2009 IEEE conference on computer vision and pattern
  recognition}, pages 248--255. Ieee, 2009.

\bibitem[Devlin et~al.(2018)]{devlin2018bert}
Jacob Devlin, Ming-Wei Chang, Kenton Lee, and Kristina Toutanova.
\newblock Bert: Pre-training of deep bidirectional transformers for language
  understanding.
\newblock {\em arXiv preprint arXiv:1810.04805}, 2018.

\bibitem[Dosovitskiy et~al.(2020)]{dosovitskiy2020image}
Alexey Dosovitskiy, Lucas Beyer, Alexander Kolesnikov, Dirk Weissenborn,
  Xiaohua Zhai, Thomas Unterthiner, Mostafa Dehghani, Matthias Minderer, Georg
  Heigold, Sylvain Gelly, et~al.
\newblock An image is worth 16x16 words: Transformers for image recognition at
  scale.
\newblock {\em arXiv preprint arXiv:2010.11929}, 2020.

\bibitem[Ghiasi et~al.(2022)]{ghiasi2022vision}
Amin Ghiasi, Hamid Kazemi, Eitan Borgnia, Steven Reich, Manli Shu, Micah
  Goldblum, Andrew~Gordon Wilson, and Tom Goldstein.
\newblock What do vision transformers learn? a visual exploration.
\newblock {\em arXiv preprint arXiv:2212.06727}, 2022.

\bibitem[Grauman and Darrell(2005)]{grauman2005pyramid}
Kristen Grauman and Trevor Darrell.
\newblock The pyramid match kernel: Discriminative classification with sets of
  image features.
\newblock In {\em Tenth IEEE International Conference on Computer Vision
  (ICCV'05) Volume 1}, volume~2, pages 1458--1465. IEEE, 2005.

\bibitem[Grigg et~al.(2021)]{grigg2021self}
Tom~George Grigg, Dan Busbridge, Jason Ramapuram, and Russ Webb.
\newblock Do self-supervised and supervised methods learn similar visual
  representations?
\newblock {\em arXiv preprint arXiv:2110.00528}, 2021.

\bibitem[Grill et~al.(2020)]{grill2020bootstrap}
Jean-Bastien Grill, Florian Strub, Florent Altch{\'e}, Corentin Tallec, Pierre
  Richemond, Elena Buchatskaya, Carl Doersch, Bernardo Avila~Pires, Zhaohan
  Guo, Mohammad Gheshlaghi~Azar, et~al.
\newblock Bootstrap your own latent-a new approach to self-supervised learning.
\newblock {\em Advances in neural information processing systems},
  33:21271--21284, 2020.

\bibitem[Gwilliam and Shrivastava(2022)]{gwilliam2022beyond}
Matthew Gwilliam and Abhinav Shrivastava.
\newblock Beyond supervised vs. unsupervised: Representative benchmarking and
  analysis of image representation learning.
\newblock In {\em Proceedings of the IEEE/CVF Conference on Computer Vision and
  Pattern Recognition}, pages 9642--9652, 2022.

\bibitem[Hamilton et~al.(2022)]{hamilton2022unsupervised}
Mark Hamilton, Zhoutong Zhang, Bharath Hariharan, Noah Snavely, and William~T
  Freeman.
\newblock Unsupervised semantic segmentation by distilling feature
  correspondences.
\newblock {\em arXiv preprint arXiv:2203.08414}, 2022.

\bibitem[He et~al.(2021)]{he2021partimagenet}
Ju He, Shuo Yang, Shaokang Yang, Adam Kortylewski, Xiaoding Yuan, Jie-Neng
  Chen, Shuai Liu, Cheng Yang, and Alan Yuille.
\newblock Partimagenet: A large, high-quality dataset of parts.
\newblock {\em arXiv preprint arXiv:2112.00933}, 2021.

\bibitem[He et~al.(2022)]{he2022masked}
Kaiming He, Xinlei Chen, Saining Xie, Yanghao Li, Piotr Doll{\'a}r, and Ross
  Girshick.
\newblock Masked autoencoders are scalable vision learners.
\newblock In {\em Proceedings of the IEEE/CVF Conference on Computer Vision and
  Pattern Recognition}, pages 16000--16009, 2022.

\bibitem[He et~al.(2020)]{he2020momentum}
Kaiming He, Haoqi Fan, Yuxin Wu, Saining Xie, and Ross Girshick.
\newblock Momentum contrast for unsupervised visual representation learning.
\newblock In {\em Proceedings of the IEEE/CVF conference on computer vision and
  pattern recognition}, pages 9729--9738, 2020.

\bibitem[He et~al.(2015)]{he2015spatial}
Kaiming He, Xiangyu Zhang, Shaoqing Ren, and Jian Sun.
\newblock Spatial pyramid pooling in deep convolutional networks for visual
  recognition.
\newblock {\em IEEE transactions on pattern analysis and machine intelligence},
  37(9):1904--1916, 2015.

\bibitem[Hjelm et~al.(2018)]{hjelm2018learning}
R~Devon Hjelm, Alex Fedorov, Samuel Lavoie-Marchildon, Karan Grewal, Phil
  Bachman, Adam Trischler, and Yoshua Bengio.
\newblock Learning deep representations by mutual information estimation and
  maximization.
\newblock {\em arXiv preprint arXiv:1808.06670}, 2018.

\bibitem[Jabri et~al.(2020)]{jabri2020space}
Allan Jabri, Andrew Owens, and Alexei Efros.
\newblock Space-time correspondence as a contrastive random walk.
\newblock {\em Advances in neural information processing systems},
  33:19545--19560, 2020.

\bibitem[Kornblith et~al.(2019)]{kornblith2019similarity}
Simon Kornblith, Mohammad Norouzi, Honglak Lee, and Geoffrey Hinton.
\newblock Similarity of neural network representations revisited.
\newblock In {\em International Conference on Machine Learning}, pages
  3519--3529. PMLR, 2019.

\bibitem[Kotar et~al.(2021)]{kotar2021contrasting}
Klemen Kotar, Gabriel Ilharco, Ludwig Schmidt, Kiana Ehsani, and Roozbeh
  Mottaghi.
\newblock Contrasting contrastive self-supervised representation learning
  pipelines.
\newblock In {\em Proceedings of the IEEE/CVF International Conference on
  Computer Vision}, pages 9949--9959, 2021.

\bibitem[Lazebnik et~al.(2006)]{lazebnik2006beyond}
Svetlana Lazebnik, Cordelia Schmid, and Jean Ponce.
\newblock Beyond bags of features: Spatial pyramid matching for recognizing
  natural scene categories.
\newblock In {\em 2006 IEEE computer society conference on computer vision and
  pattern recognition (CVPR'06)}, volume~2, pages 2169--2178. IEEE, 2006.

\bibitem[Li et~al.(2021)]{li2021benchmarking}
Yanghao Li, Saining Xie, Xinlei Chen, Piotr Dollar, Kaiming He, and Ross
  Girshick.
\newblock Benchmarking detection transfer learning with vision transformers.
\newblock {\em arXiv preprint arXiv:2111.11429}, 2021.

\bibitem[Lin et~al.(2017)]{lin2017feature}
Tsung-Yi Lin, Piotr Doll{\'a}r, Ross Girshick, Kaiming He, Bharath Hariharan,
  and Serge Belongie.
\newblock Feature pyramid networks for object detection.
\newblock In {\em Proceedings of the IEEE conference on computer vision and
  pattern recognition}, pages 2117--2125, 2017.

\bibitem[Lin et~al.(2014)]{lin2014microsoft}
Tsung-Yi Lin, Michael Maire, Serge Belongie, James Hays, Pietro Perona, Deva
  Ramanan, Piotr Doll{\'a}r, and C~Lawrence Zitnick.
\newblock Microsoft coco: Common objects in context.
\newblock In {\em European conference on computer vision}, pages 740--755.
  Springer, 2014.

\bibitem[Liu et~al.(2021)]{liu2021swin}
Ze Liu, Yutong Lin, Yue Cao, Han Hu, Yixuan Wei, Zheng Zhang, Stephen Lin, and
  Baining Guo.
\newblock Swin transformer: Hierarchical vision transformer using shifted
  windows.
\newblock In {\em Proceedings of the IEEE/CVF International Conference on
  Computer Vision}, pages 10012--10022, 2021.

\bibitem[Loshchilov and Hutter(2017)]{loshchilov2017decoupled}
Ilya Loshchilov and Frank Hutter.
\newblock Decoupled weight decay regularization.
\newblock {\em arXiv preprint arXiv:1711.05101}, 2017.

\bibitem[Min et~al.(2019)]{min2019spair}
Juhong Min, Jongmin Lee, Jean Ponce, and Minsu Cho.
\newblock Spair-71k: A large-scale benchmark for semantic correspondence.
\newblock {\em arXiv preprint arXiv:1908.10543}, 2019.

\bibitem[Misra and Maaten(2020)]{misra2020self}
Ishan Misra and Laurens van~der Maaten.
\newblock Self-supervised learning of pretext-invariant representations.
\newblock In {\em Proceedings of the IEEE/CVF Conference on Computer Vision and
  Pattern Recognition}, pages 6707--6717, 2020.

\bibitem[Mu and Andreas(2020)]{mu2020compositional}
Jesse Mu and Jacob Andreas.
\newblock Compositional explanations of neurons.
\newblock {\em Advances in Neural Information Processing Systems},
  33:17153--17163, 2020.

\bibitem[Nguyen et~al.(2020)]{nguyen2020wide}
Thao Nguyen, Maithra Raghu, and Simon Kornblith.
\newblock Do wide and deep networks learn the same things? uncovering how
  neural network representations vary with width and depth.
\newblock {\em arXiv preprint arXiv:2010.15327}, 2020.

\bibitem[Oord et~al.(2018)]{oord2018representation}
Aaron van~den Oord, Yazhe Li, and Oriol Vinyals.
\newblock Representation learning with contrastive predictive coding.
\newblock {\em arXiv preprint arXiv:1807.03748}, 2018.

\bibitem[Park and Kim(2022)]{park2022vision}
Namuk Park and Songkuk Kim.
\newblock How do vision transformers work?
\newblock {\em arXiv preprint arXiv:2202.06709}, 2022.

\bibitem[Paul and Chen(2022)]{paul2022vision}
Sayak Paul and Pin-Yu Chen.
\newblock Vision transformers are robust learners.
\newblock In {\em Proceedings of the AAAI Conference on Artificial
  Intelligence}, volume~36, pages 2071--2081, 2022.

\bibitem[Philbin et~al.(2008)]{philbin2008lost}
James Philbin, Ondrej Chum, Michael Isard, Josef Sivic, and Andrew Zisserman.
\newblock Lost in quantization: Improving particular object retrieval in large
  scale image databases.
\newblock In {\em 2008 IEEE conference on computer vision and pattern
  recognition}, pages 1--8. IEEE, 2008.

\bibitem[Pinheiro et~al.(2016)]{pinheiro2016learning}
Pedro~O Pinheiro, Tsung-Yi Lin, Ronan Collobert, and Piotr Doll{\'a}r.
\newblock Learning to refine object segments.
\newblock In {\em European conference on computer vision}, pages 75--91.
  Springer, 2016.

\bibitem[Pont-Tuset et~al.(2017)]{pont20172017}
Jordi Pont-Tuset, Federico Perazzi, Sergi Caelles, Pablo Arbel{\'a}ez, Alex
  Sorkine-Hornung, and Luc Van~Gool.
\newblock The 2017 davis challenge on video object segmentation.
\newblock {\em arXiv preprint arXiv:1704.00675}, 2017.

\bibitem[Purushwalkam and Gupta(2020)]{purushwalkam2020demystifying}
Senthil Purushwalkam and Abhinav Gupta.
\newblock Demystifying contrastive self-supervised learning: Invariances,
  augmentations and dataset biases.
\newblock {\em Advances in Neural Information Processing Systems},
  33:3407--3418, 2020.

\bibitem[Radenovi{\'c} et~al.(2018)]{radenovic2018revisiting}
Filip Radenovi{\'c}, Ahmet Iscen, Giorgos Tolias, Yannis Avrithis, and
  Ond{\v{r}}ej Chum.
\newblock Revisiting oxford and paris: Large-scale image retrieval
  benchmarking.
\newblock In {\em Proceedings of the IEEE conference on computer vision and
  pattern recognition}, pages 5706--5715, 2018.

\bibitem[Radford et~al.(2021)]{radford2021learning}
Alec Radford, Jong~Wook Kim, Chris Hallacy, Aditya Ramesh, Gabriel Goh,
  Sandhini Agarwal, Girish Sastry, Amanda Askell, Pamela Mishkin, Jack Clark,
  et~al.
\newblock Learning transferable visual models from natural language
  supervision.
\newblock In {\em International Conference on Machine Learning}, pages
  8748--8763. PMLR, 2021.

\bibitem[Raghu et~al.(2021)]{raghu2021vision}
Maithra Raghu, Thomas Unterthiner, Simon Kornblith, Chiyuan Zhang, and Alexey
  Dosovitskiy.
\newblock Do vision transformers see like convolutional neural networks?
\newblock {\em Advances in Neural Information Processing Systems},
  34:12116--12128, 2021.

\bibitem[Sahiner et~al.(2022)]{sahiner2022unraveling}
Arda Sahiner, Tolga Ergen, Batu Ozturkler, John Pauly, Morteza Mardani, and
  Mert Pilanci.
\newblock Unraveling attention via convex duality: Analysis and interpretations
  of vision transformers.
\newblock {\em arXiv preprint arXiv:2205.08078}, 2022.

\bibitem[Schonberger and Frahm(2016)]{schonberger2016structure}
Johannes~L Schonberger and Jan-Michael Frahm.
\newblock Structure-from-motion revisited.
\newblock In {\em Proceedings of the IEEE conference on computer vision and
  pattern recognition}, pages 4104--4113, 2016.

\bibitem[Shao et~al.(2021)]{shao2021adversarial}
Rulin Shao, Zhouxing Shi, Jinfeng Yi, Pin-Yu Chen, and Cho-Jui Hsieh.
\newblock On the adversarial robustness of vision transformers.
\newblock {\em arXiv preprint arXiv:2103.15670}, 2021.

\bibitem[Shrivastava et~al.(2016)]{shrivastava2016beyond}
Abhinav Shrivastava, Rahul Sukthankar, Jitendra Malik, and Abhinav Gupta.
\newblock Beyond skip connections: Top-down modulation for object detection.
\newblock {\em arXiv preprint arXiv:1612.06851}, 2016.

\bibitem[Sim{\'e}oni et~al.(2021)]{simeoni2021localizing}
Oriane Sim{\'e}oni, Gilles Puy, Huy~V Vo, Simon Roburin, Spyros Gidaris, Andrei
  Bursuc, Patrick P{\'e}rez, Renaud Marlet, and Jean Ponce.
\newblock Localizing objects with self-supervised transformers and no labels.
\newblock {\em arXiv preprint arXiv:2109.14279}, 2021.

\bibitem[Steiner et~al.(2021)]{steiner2021train}
Andreas Steiner, Alexander Kolesnikov, Xiaohua Zhai, Ross Wightman, Jakob
  Uszkoreit, and Lucas Beyer.
\newblock How to train your vit? data, augmentation, and regularization in
  vision transformers.
\newblock {\em arXiv preprint arXiv:2106.10270}, 2021.

\bibitem[Subramanian(2021)]{subramanian2021torch_cka}
Anand Subramanian.
\newblock Torch cka.
\newblock \url{https://github.com/AntixK/PyTorch-Model-Compare}, 2021.

\bibitem[Tian et~al.(2020)]{tian2020contrastive}
Yonglong Tian, Dilip Krishnan, and Phillip Isola.
\newblock Contrastive multiview coding.
\newblock In {\em European conference on computer vision}, pages 776--794.
  Springer, 2020.

\bibitem[Touvron et~al.(2022)]{touvron2022three}
Hugo Touvron, Matthieu Cord, Alaaeldin El-Nouby, Jakob Verbeek, and Herv{\'e}
  J{\'e}gou.
\newblock Three things everyone should know about vision transformers.
\newblock {\em arXiv preprint arXiv:2203.09795}, 2022.

\bibitem[Van~Gansbeke et~al.(2021)]{van2021revisiting}
Wouter Van~Gansbeke, Simon Vandenhende, Stamatios Georgoulis, and Luc~V Gool.
\newblock Revisiting contrastive methods for unsupervised learning of visual
  representations.
\newblock {\em Advances in Neural Information Processing Systems},
  34:16238--16250, 2021.

\bibitem[Van~Gansbeke et~al.(2020)]{van2020scan}
Wouter Van~Gansbeke, Simon Vandenhende, Stamatios Georgoulis, Marc Proesmans,
  and Luc Van~Gool.
\newblock Scan: Learning to classify images without labels.
\newblock In {\em European conference on computer vision}, pages 268--285.
  Springer, 2020.

\bibitem[Van~Horn et~al.(2021)]{van2021benchmarking}
Grant Van~Horn, Elijah Cole, Sara Beery, Kimberly Wilber, Serge Belongie, and
  Oisin Mac~Aodha.
\newblock Benchmarking representation learning for natural world image
  collections.
\newblock In {\em Proceedings of the IEEE/CVF conference on computer vision and
  pattern recognition}, pages 12884--12893, 2021.

\bibitem[Vaswani et~al.(2017)]{vaswani2017attention}
Ashish Vaswani, Noam Shazeer, Niki Parmar, Jakob Uszkoreit, Llion Jones,
  Aidan~N Gomez, {\L}ukasz Kaiser, and Illia Polosukhin.
\newblock Attention is all you need.
\newblock {\em Advances in neural information processing systems}, 30, 2017.

\bibitem[Wang and Liu(2021)]{wang2021understanding}
Feng Wang and Huaping Liu.
\newblock Understanding the behaviour of contrastive loss.
\newblock In {\em Proceedings of the IEEE/CVF conference on computer vision and
  pattern recognition}, pages 2495--2504, 2021.

\bibitem[Wang and Isola(2020)]{wang2020understanding}
Tongzhou Wang and Phillip Isola.
\newblock Understanding contrastive representation learning through alignment
  and uniformity on the hypersphere.
\newblock In {\em International Conference on Machine Learning}, pages
  9929--9939. PMLR, 2020.

\bibitem[Wang et~al.(2022)]{wang2022self}
Yangtao Wang, Xi Shen, Shell~Xu Hu, Yuan Yuan, James~L Crowley, and Dominique
  Vaufreydaz.
\newblock Self-supervised transformers for unsupervised object discovery using
  normalized cut.
\newblock In {\em Proceedings of the IEEE/CVF Conference on Computer Vision and
  Pattern Recognition}, pages 14543--14553, 2022.

\bibitem[Wightman(2019)]{rw2019timm}
Ross Wightman.
\newblock Pytorch image models.
\newblock \url{https://github.com/rwightman/pytorch-image-models}, 2019.

\bibitem[Wu et~al.(2018)]{wu2018unsupervised}
Zhirong Wu, Yuanjun Xiong, Stella~X Yu, and Dahua Lin.
\newblock Unsupervised feature learning via non-parametric instance
  discrimination.
\newblock In {\em Proceedings of the IEEE conference on computer vision and
  pattern recognition}, pages 3733--3742, 2018.

\bibitem[Xu et~al.(2009)]{xu2009efficient}
Kun Xu, Yong Li, Tao Ju, Shi-Min Hu, and Tian-Qiang Liu.
\newblock Efficient affinity-based edit propagation using kd tree.
\newblock {\em ACM Transactions on Graphics (TOG)}, 28(5):1--6, 2009.

\bibitem[Yang and Ramanan(2012)]{yang2012articulated}
Yi Yang and Deva Ramanan.
\newblock Articulated human detection with flexible mixtures of parts.
\newblock {\em IEEE transactions on pattern analysis and machine intelligence},
  35(12):2878--2890, 2012.

\bibitem[Zbontar et~al.(2021)]{zbontar2021barlow}
Jure Zbontar, Li Jing, Ishan Misra, Yann LeCun, and St{\'e}phane Deny.
\newblock Barlow twins: Self-supervised learning via redundancy reduction.
\newblock In {\em International Conference on Machine Learning}, pages
  12310--12320. PMLR, 2021.

\bibitem[Zhang et~al.(2017)]{zhang2017mixup}
Hongyi Zhang, Moustapha Cisse, Yann~N Dauphin, and David Lopez-Paz.
\newblock mixup: Beyond empirical risk minimization.
\newblock {\em arXiv preprint arXiv:1710.09412}, 2017.

\bibitem[Zhou et~al.(2022)]{zhou2022understanding}
Daquan Zhou, Zhiding Yu, Enze Xie, Chaowei Xiao, Animashree Anandkumar, Jiashi
  Feng, and Jose~M Alvarez.
\newblock Understanding the robustness in vision transformers.
\newblock In {\em International Conference on Machine Learning}, pages
  27378--27394. PMLR, 2022.

\end{thebibliography}
}

\clearpage
\appendix
\section{Appendix Overview}

In our primary results, we focused on comparing ViT-B/16 models trained with different supervision methods. In this Appendix, we present additional results for a wider range of ViT variants including different architecture sizes and patch sizes. This analysis includes $20$ models in total, summarized in \Cref{tab:architectures}. This Appendix is organized following the same three major domains: Attention, Features, and Downstream Tasks. We summarize all the key findings of our work in \Cref{tab:overview}.

\section{Additional Experimental Details}

\subsection{ViT Variants Examined}

To provide a uniform platform for comparison, our primary results focus on ViT-B/16 models trained with six methods: Full Supervision, CLIP~\citep{radford2021learning}, DINO~\citep{caron2021emerging}, MoCo-v3~\citep{chen2021empirical}, MAE~\citep{he2022masked}, and BEiT~\citep{bao2021beit}.
In this Appendix, we present additional results examining different ViT variants for the above methods as provided by their original authors. In total, this expanded ViT collection contains $20$ models including instances of ViT Small, Base, Large, and Huge as well as patch sizes $8$, $14$, $16$, and $32$. We continue to process all images at a $224\times224$ input resolution, meaning the number of spatial tokens for a given model will vary by patch size, from $47$ spatial tokens for models with patch size $32$ up to $784$ spatial tokens for patch size $8$. Overall, the models evaluated are summarized in \Cref{tab:architectures}. Note that the MoCo ViT-Small model is a modified variant with $12$ heads per layer instead of $6$.

\begin{table}[t]
 \setlength{\cmidrulewidth}{0.01em}
\renewcommand{\tabcolsep}{6pt}
\renewcommand{\arraystretch}{1}
\caption{Summary of all ViT Variants used in Appendix analysis. *MoCo S/16 uses 12 heads per layer instead of 6.}
\vspace{-5pt}
\centering
\footnotesize
\resizebox{0.9\linewidth}{!}{
\begin{tabular}{@{}lccc@{}}
\toprule
 \textbf{Model} &  \textbf{Layers} & \textbf{Heads} & \textbf{ \makecell[c]{Spatial Token \\Grid Size}}\\
 \midrule
 
 FS S/32 & 12 & 6 & 7x7     \\
FS S/16 & 12 & 6 & 14x14   \\
FS B/32 & 12 & 12 & 7x7     \\
FS B/16 & 12 & 12 & 14x14   \\
FS B/8  & 12 & 12 & 28x28   \\
FS L/16 & 24 & 16 & 14x14   \\
CLIP B/32 & 12 & 12 & 7x7     \\
CLIP B/16 & 12 & 12 & 14x14   \\
CLIP L/14 & 24 & 16 & 16x16   \\
DINO S/16 & 12 & 6 & 14x14   \\
DINO S/8  & 12 & 6 & 28x28   \\
DINO B/16 & 12 & 12 & 14x14   \\
DINO B/8  & 12 & 12 & 28x28   \\
MoCo S/16* & 12 & 12 & 14x14   \\
MoCo B/16 & 12 & 12 & 14x14   \\
MAE B/16  & 12 & 12 & 14x14   \\
MAE L/16  & 24 & 16 & 14x14   \\
MAE H/14  & 32 & 16 & 16x16   \\
BEiT B/16 & 12 & 12 & 14x14   \\
BEiT L/16 & 24 & 16 & 14x14   \\
\bottomrule
\end{tabular}
  }
\label{tab:architectures}
\vspace{-9pt}
\end{table}

\subsection{ViT Training Details}
\label{sec:apdx-vits}

In this section, we briefly outline the training protocols that were used to train each of the ViTs examined in this work. The CLIP, DINO, MoCo, MAE, and BEiT models are all official pre-trained models released by their original authors.

\textbf{Fully Supervised (FS).} For FS, we work with models from the TIMM repository~\citep{rw2019timm}. The FS models are pretrained on ImageNet21k and fine-tuned on ImageNet1k. The FS models are the only models in this study that are fine-tuned with ImageNet-1k labels. They are trained following the augmentation protocols of~\citep{steiner2021train}. Specifically, the ViT-Base and Large models are trained using a combination of RandAugment~\citep{cubuk2020randaugment} and Mixup~\citep{zhang2017mixup}, while the ViT-Small models use only RandAugment. All FS models also use weight decay~\citep{loshchilov2017decoupled}.

\textbf{CLIP.} The goal of CLIP (\textbf{C}ontrastive \textbf{L}anguage-\textbf{I}mage \textbf{P}re-Training) is to train models with open-ended supervision provided by paired captions.
The learning objective is simply to match images with their corresponding captions.
CLIP models are joint vision and language models, which include separate encoder networks for the image and text inputs. For our analysis, we focus only on the properties of the visual encoding network. The authors pretrain the model on $400$M image-text pairs with a batch size of $32768$ and mixed-precision to accelerate training and reduce memory usage. The only augmentation used is taking a random square crop from the resized image. They also use a cosine learning rate decay schedule.

\textbf{MoCo.} The \textbf{Mo}mentum \textbf{Co}ntrast (MoCo) method trains using contrastive learning with a momentum encoder, which is an exponential moving average of previous versions of the encoder. Under the contrastive objective, the encoder must generate representations for query image views that are similar to corresponding representations of key image views as generated by the momentum encoder. This strategy was proposed for CNNs and extended to ViTs. For MoCo v3~\citep{chen2021empirical}, the authors pretrain the model on ImageNet-1k without labels with a batch size of $4096$. They follow a cosine learning rate decay. They use data augmentations like random resized cropping, horizontal flipping, color jittering, grayscale conversion, blurring, and solarization. They take two $224\times224$ crops for each image for each iteration.

\textbf{DINO.} The authors of DINO~\citep{caron2021emerging} describe their method as a form of self-\textbf{di}stillation with \textbf{no} labels.
Their training strategy is based on MoCo~\citep{he2020momentum} and they also use a momentum encoder, though they instead view their method as a student/teacher knowledge distillation framework. They pretrain the models on ImageNet-1k without labels with batch

\begin{table*}[t]
\setlength{\cmidrulewidth}{0.01em}
\renewcommand{\tabcolsep}{6pt}
\renewcommand{\arraystretch}{1}
\caption{A comprehensive summary of the key observations of this work, including both the main paper and appendix results.}
\centering\renewcommand\cellalign{lc}
\footnotesize
\resizebox{0.89\linewidth}{!}{
\begin{tabular}{@{}lccc@{}}
\toprule
 \textbf{Key Observations} &  \textbf{Analysis Methods} & \textbf{Sections} & \textbf{Figures \& Tables}\\
 \midrule \\[-8pt]

\makecell*[{{p{7.5cm}}}]{The attention maps of explicitly supervised ViTs devolve into Sparse Repeating Attention Patterns in the mid-to-late layers.} & Average CLS Attention Maps & \ref{sec:att_vis} & \Cref{fig:att-grid-5k} \\

\makecell*[{{p{7.5cm}}}]{All ViTs studied learn to use Offset Local Attention Heads, suggesting they are fundamentally necessary in ViTs.} & Aligned Aggregated Attention Maps & \ref{sec:att_olah} & \Cref{fig:att-grid-aligned-agg} \\

\makecell*[{{p{7.5cm}}}]{ViTs learn to process local and global information in different orders depending on their method of supervision.} & Average Attention Distance & \ref{sec:att_dist} & \Cref{fig:att-dist-plots} \\

\makecell*[{{p{7.5cm}}}]{All ViTs studied differentiate salient foreground objects by the early-to-mid layers.} & Attention Saliency IoU & \ref{sec:att_iou} & \Cref{fig:att-iou-plots} \\

\makecell*[{{p{7.5cm}}}]{Reconstruction-based self-supervised methods can learn semantically meaningful CLS representations, even when the CLS token is only a placeholder.} & \makecell[c]{CKA Feature Similarity, \\ Image Clustering by CLS Features} & \ref{sec:cka_lb}, \ref{sec:cluster-cls} & Figures \ref{fig:cka-lastlayer}, \ref{fig:cluster-purity-global} \\

\makecell*[{{p{7.5cm}}}]{Supervised method’s features are the most semantically rich, but contrastive self-supervised methods are comparable or even superior in some cases.} & \makecell[c]{Image-, Object-, and Part-Level \\ Feature Clustering} & \ref{sec:cluster-cls}, \ref{sec:cluster-spatial} & Figures \ref{fig:cluster-purity-global}, \ref{fig:cluster-purity-local} \\

\makecell*[{{p{7.5cm}}}]{For localized tasks, the best performance often comes from a mid-to-late layer.} & Local Downstream Tasks & \ref{sec:down-seg} & \makecell[c]{\Cref{fig:local-tasks} \\ \Cref{tab:best_performance}} \\

 \makecell*[{{p{7.5cm}}}]{There is no single ``best" training method or layer for all downstream tasks.} & Local \& Global Downstream Tasks & \ref{sec:down-st} & \makecell[c]{Figures \ref{fig:global-tasks}, \ref{fig:local-tasks} \\ \Cref{tab:best_performance}} \\[10pt]

\bottomrule \\[-5pt]

\makecell*[{{p{7.5cm}}}]{The positions of maximal activation in the Sparse Repeating Attention Patterns vary by input.} & CLS Attention Maps & \ref{sec:apdx-att-vis} & Figures \ref{fig:apdx-single-sample-extra-2}-\ref{fig:apdx-single-img-samples} \\

\makecell*[{{p{7.5cm}}}]{All models studied learn to use Offset Local Attention Heads, and some larger models even learn ones with diagonal offsets.} & Aligned Aggregated Attention Maps & \ref{sec:apdx-att-vis} & \Cref{fig:apdx-offset-samples} \\

\makecell*[{{p{7.5cm}}}]{The order of local \vs global information processing in a ViT is primarily determined by the method of supervision and is largely unaffected by changes in architecture and patch size.} & Average Attention Distance & \ref{sec:apdx-att-dist} & \Cref{fig:apdx-att-dist} \\

\makecell*[{{p{7.5cm}}}]{For the expanded ViT collection, again all models differentiate salient foreground objects by the early-to-mid layers.} & Attention Saliency IoU & \ref{sec:apdx-att-sal} & \Cref{fig:apdx-att-sal} \\

\makecell*[{{p{7.5cm}}}]{Explicitly supervised ViTs with patch size $32$ are less impacted by Sparse Repeating Attention Patterns, suggesting they may be an indication of overfitting.} & Averaged CLS Attention Maps & \ref{sec:apdx-att-vis}, \ref{sec:apdx-att-sal} & \Cref{fig:big-cls} \\

\makecell*[{{p{7.5cm}}}]{The last layer spatial representations of self-supervised methods are similar across changes in architecture size and patch size, but this is not consistently true for explicitly supervised methods.} & CKA Feature Similarity & \ref{sec:apdx-feat-cka} & \Cref{fig:cka-lastlayer-all} \\

\makecell*[{{p{7.5cm}}}]{Both MAE and BEiT show X patterns in their depth-wise feature CKAs, suggesting an encoder/decoder internal structure.} & CKA Feature Similarity & \ref{sec:apdx-feat-cka-depth} & Figures \ref{fig:mae_self}, \ref{fig:beit_self} \\

\makecell*[{{p{7.5cm}}}]{For larger MAE ViTs, the later layers appear to act more like decoder layers, even thought MAE has a separate decoder.} & CKA Feature Similarity & \ref{sec:apdx-feat-cka-depth} & \Cref{fig:mae_self} \\

\makecell*[{{p{7.5cm}}}]{Residual connection analysis provides further evidence of a fundamental shift in information processing in the mid-to-late layers of explicitly supervised ViTs.} & Residual Connection Analysis & \ref{sec:residual} & \Cref{fig:residual} \\

\makecell*[{{p{7.5cm}}}]{BEiT L/16 learns extremely expressive part-level features compared with BEiT B/16.} & Part-Level Feature Clustering & \ref{sec:apdx-feat-clust} & 
\Cref{fig:apdx-cluster-parts} \\

\makecell*[{{p{7.5cm}}}]{Larger architectures tend to give better feature quality and downstream performance.} & \makecell[c]{Feature Clustering, Local \& \\ Global Downstream Tasks} & \ref{sec:apdx-feat-clust}, \ref{sec:apdx-downstream} & Figures \ref{fig:apdx-cluster-cls}-\ref{fig:apdx-spair-all} \\

\makecell*[{{p{7.5cm}}}]{ViTs with smaller patch sizes unsurprisingly perform better at localized downstream tasks.} & Local Downstream Tasks & \ref{sec:apdx-downstream} & Figures \ref{fig:apdx-davis-all}, \ref{fig:apdx-spair-all} \\

\makecell*[{{p{7.5cm}}}]{Reconstruction-based ViTs show the largest variance in their performance characteristics on downstream tasks.} & Local \& Global Downstream Tasks & \ref{sec:apdx-downstream} & Figures \ref{fig:apdx-knn-all}-\ref{fig:apdx-spair-all} \\

\makecell*[{{p{7.5cm}}}]{For the expanded group of ViTs, once again there is no single ``best" training method or layer for all downstream tasks.} & Local \& Global Downstream Tasks & \ref{sec:apdx-downstream} & \makecell[c]{Figures \ref{fig:apdx-knn-all}-\ref{fig:apdx-spair-all} \\ \Cref{tab:best_performance_all}} \\

\bottomrule
\end{tabular}
  }
\label{tab:overview}
\end{table*}
\clearpage

\noindent size $1024$. They follow a cosine learning rate and weight decay. They use data augmentations like color jittering, gaussian blur, and solarization similar to BYOL~\citep{grill2020bootstrap}. Multi-crop~\citep{caron2020unsupervised} is also used.

\textbf{MAE.} The \textbf{M}asked \textbf{A}uto\textbf{e}ncoder (MAE)~\citep{he2022masked} method is a reconstruction-based training objective where a large portion of input patches/tokens are masked out. The rational of MAE is that, because a large percentage of the image content is missing, the network must learn representations that embed meaningful high-level semantics to reconstruct the missing regions. MAE uses both an encoder and decoder network, though the decoder is discarded after pretraining. The authors pretrain the model on ImageNet-1k without labels with a batch size of $4096$. They do not use color jittering, drop path or gradient clipping and only apply random resized crop augmentation. They use a masking ratio of $0.75$ which also improves the efficiency of training by significantly decreasing the token count in the encoder. They also use a cosine learning rate decay schedule.

\textbf{BEiT.} BEiT~\citep{bao2021beit} stands for \textbf{B}idirectional \textbf{E}ncoder representation from \textbf{I}mage \textbf{T}ransformers, and it is based on BERT~\citep{devlin2018bert}, a well-known masked reconstruction learning method for NLP. In contrast to MAE, BEiT does not perform pixel-level reconstruction, but instead uses a tokenizer to convert image patches into discrete tokens. The BEiT learning objective is to predict the token values for the masked patches. Unlike MAE, BEiT does not include a separate decoder network.
BEiT is trained with a masking ratio of roughly $0.4$, though they also employ a block-masking method which masks out larger adjacent groups of tokens. They pretrain BEiT on ImageNet-1k with a batch size of $2048$, and they include random resized cropping, horizontal flipping, and color jittering augmentations. They also utilize cosine learning rate decay. Note that the authors have provided both BEiT models before and after fine-tuning with ImageNet labels. For our analysis, we work with the non-fine-tuned versions, in order to focus on just the effects of the BEiT pretraining method.

For more details and exact parameters, please refer to the corresponding papers and codebases for each of the models.

\subsection{Random Chance Scores}

During our Feature Clustering and Downstream Task Analysis, we present random chance scores for both the clustering metrics and downstream task scores. To evaluate these scores, we repeat the task analysis replacing all ViT features with uniformly distributed Gaussian noise. To be more specific, we generate arrays of Gaussian noise with the exact same dimensions as the extracted feature arrays of a ViT B/16 model. These random chance scores are heavily influenced by the underlying data. For example, we see that the random chance score is quite high for the object-level clustering purity scores on COCO. This can be attributed to the dataset's highly imbalanced object distribution. Still, this method of random chance evaluation is informative as it effectively acts as a baseline model where all feature vectors contain absolutely no useful information.

\subsection{Dense Feature Extraction}
\label{sec:apdx-dense-feat}

Certain local tasks, like object segmentation, benefit from having a denser array of high quality features. For an input image of size $224\times224$, a ViT with patch size $16$ produces a feature array with size $14\times14$.
This low resolution can be very limiting for localized tasks, like DAVIS Video Segmentation Propagation. While some of the ViTs evaluated do support variable input size, others are hard-coded to operate at a fixed size. To generate a denser feature grid while also providing a level playing field, we propose a simple dense feature extraction strategy using image tiling.

We begin by rescaling the smaller image dimension to size $448$ (twice the input resolution). We then scale the larger image dimension to the nearest integer multiple of the model patch size, in order to preserve the image aspect ratio as best as possible. Finally we slice the image into non-overlapping tiles of size $224\times224$. Then we extract ViT features for each of the tiles and concatenate the features together. Unless the image is exactly square, this leaves some leftover image content along the larger image dimension. For these areas, we take two additional crops which do overlap other image tiles, but for these areas the features are discarded, while the non-overlapping features are concatenated to the rest. The final product is a feature array that is twice as dense as the original while also respecting the original image aspect ratio.

In \Cref{tab:dense}, we present an ablative analysis comparing the DAVIS Video Segmentation performance of all models with and without dense feature extraction enabled. All models see a significant performance boost with dense feature extraction, especially models with patch size $32$.

\begin{table}[t]
 \setlength{\cmidrulewidth}{0.01em}
\renewcommand{\tabcolsep}{6pt}
\renewcommand{\arraystretch}{1}
\caption{Comparison of Dense vs. Normal feature extraction for the DAVIS Video Object Segmentation task. Results show for the best layer per model, with layer number in parenthesis.}
\vspace{-5pt}
\centering
\footnotesize
\resizebox{0.6\linewidth}{!}{
\begin{tabular}{@{}lcc@{}}
\toprule
\multirow{2}{*}{\textbf{Model}} & \multicolumn{2}{c}{\textbf{J and F Mean}} \\
 \cmidrule[\cmidrulewidth](l){2-3}
 &  \textbf{Normal} & \textbf{Dense}\\
 \midrule
 
FS S/32   & 0.18 (12)  & 0.39 (9)  \\
FS S/16   & 0.34 (10) & 0.58 (8)  \\
FS B/32   & 0.17 (10) & 0.38 (9)  \\
FS B/16   & 0.34 (10) & 0.59 (8)  \\
FS B/8    & 0.51 (9)  & 0.68 (9)  \\
FS L/16   & 0.34 (19)  & 0.56 (13) \\
CLIP B/32 & 0.19 (12) & 0.41 (9)  \\
CLIP B/16 & 0.35 (9)  & 0.60 (9)  \\
CLIP L/14 & 0.38 (17) & 0.60 (17) \\
DINO S/16 & 0.32 (11) & 0.61 (11) \\
DINO S/8  & 0.52 (11) & 0.73 (12) \\
DINO B/16 & 0.32 (12) & 0.60 (12) \\
DINO B/8  & 0.51 (11) & 0.73 (10)  \\
MoCo S/16 & 0.34 (11) & 0.6 (10)   \\
MoCo B/16 & 0.33 (12) & 0.61 (11) \\
MAE B/16  & 0.29 (12) & 0.54 (12) \\
MAE L/16  & 0.31 (24) & 0.55 (23) \\
MAE H/14  & 0.36 (30) & 0.59 (30) \\
BEiT B/16 & 0.31 (7)  & 0.58 (9)   \\
BEiT L/16 & 0.36 (17) & 0.61 (15) \\
\bottomrule
\end{tabular}
  }
\label{tab:dense}
\vspace{-15pt}
\end{table}

\subsection{Code Release}

Our analysis codebase includes complete scripts for replicating the experiments and figures in the main work and appendix. Our source code is available at \url{www.github.com/mwalmer-umd/vit_analysis} and our project page can be found at \url{www.cs.umd.edu/~sakshams/vit_analysis}.
\section{Attention Analysis}
\label{sec:apdx-att}

\subsection{Expanded Attention Visualizations}
\label{sec:apdx-att-vis}

\begin{figure*}
    \centering
    \begin{subfigure}[b]{\linewidth}
        \centering
        \includegraphics[width=1.0\textwidth]{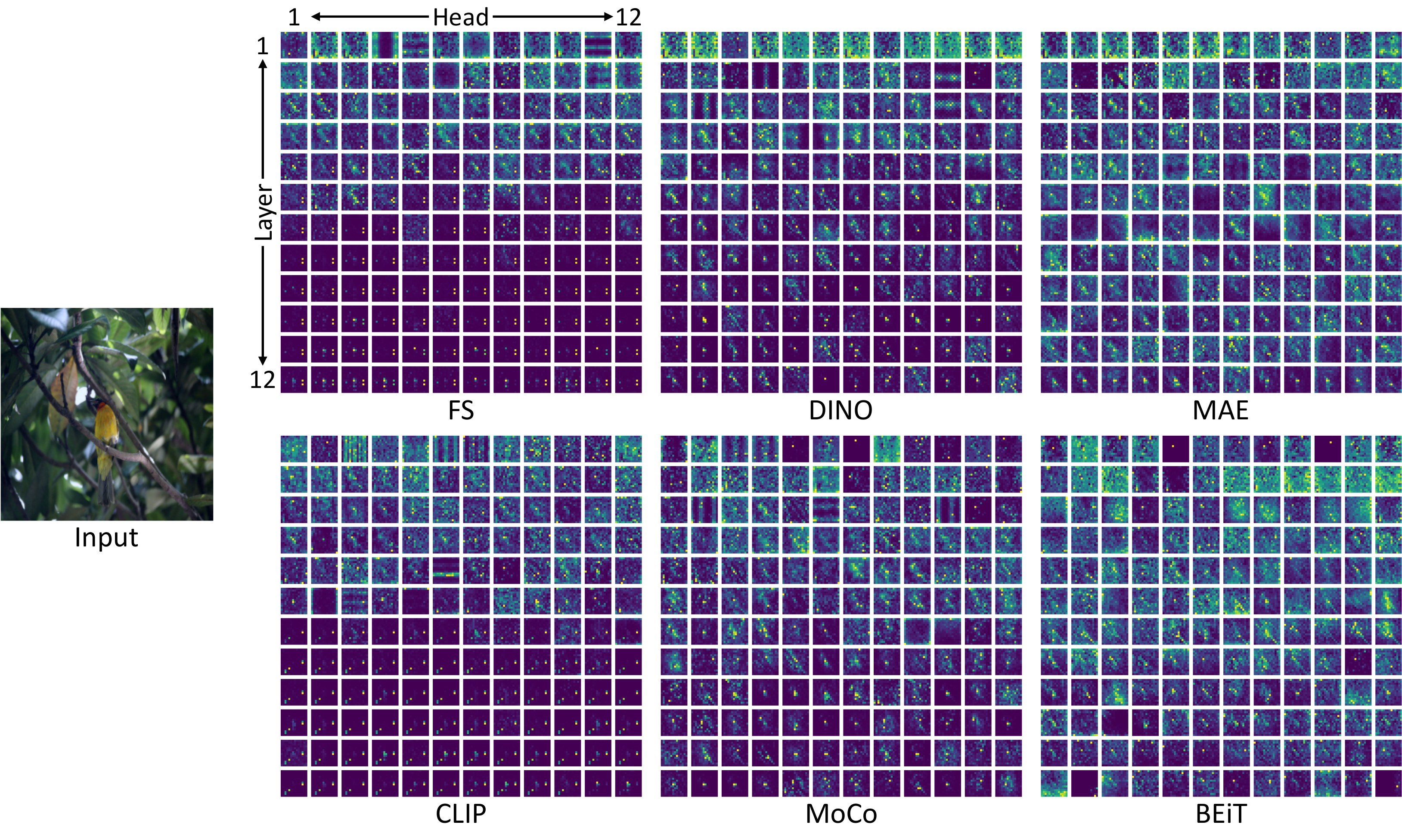}
    \end{subfigure}

    \begin{subfigure}[b]{\linewidth}
        \centering
        \includegraphics[width=1.0\textwidth]{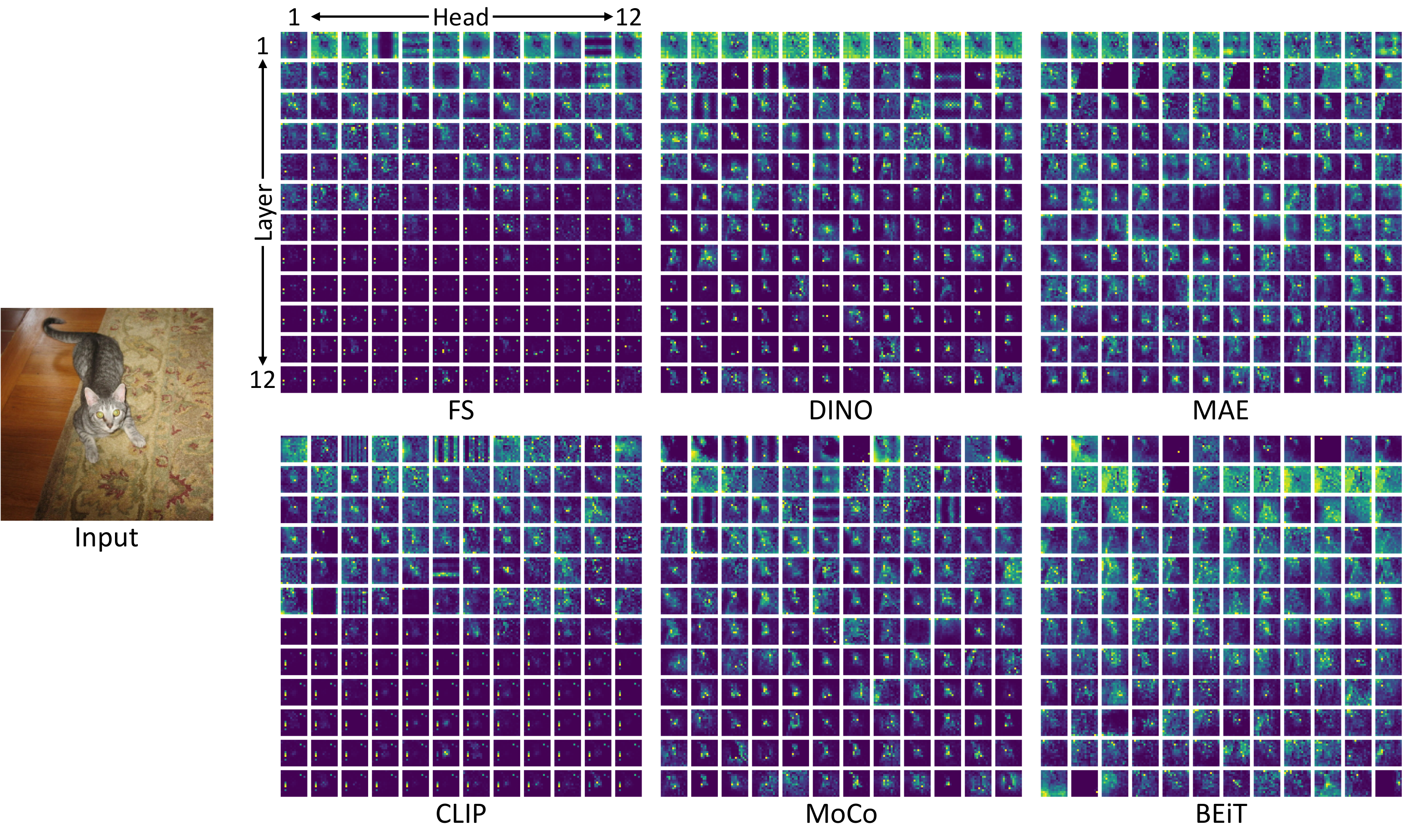}
    \end{subfigure}
    \vspace{-2.1em}
    \caption{Visualizing all CLS token attention maps for all heads and all layers in ViT B/16 models for single input images (left). The FS and CLIP models show Sparse Repeating Attention Patterns in the mid-to-late layers, where a small group of spatial token positions at seemingly arbitrary positions have strong and consistent activations shared across both heads and layers. Note that the positions of these strong repetitive activations are different for the two inputs.}
    \label{fig:apdx-single-sample-extra-2}
\end{figure*}

\begin{figure*}
    \centering
    \begin{subfigure}[b]{\linewidth}
        \centering
        \includegraphics[width=1.0\textwidth]{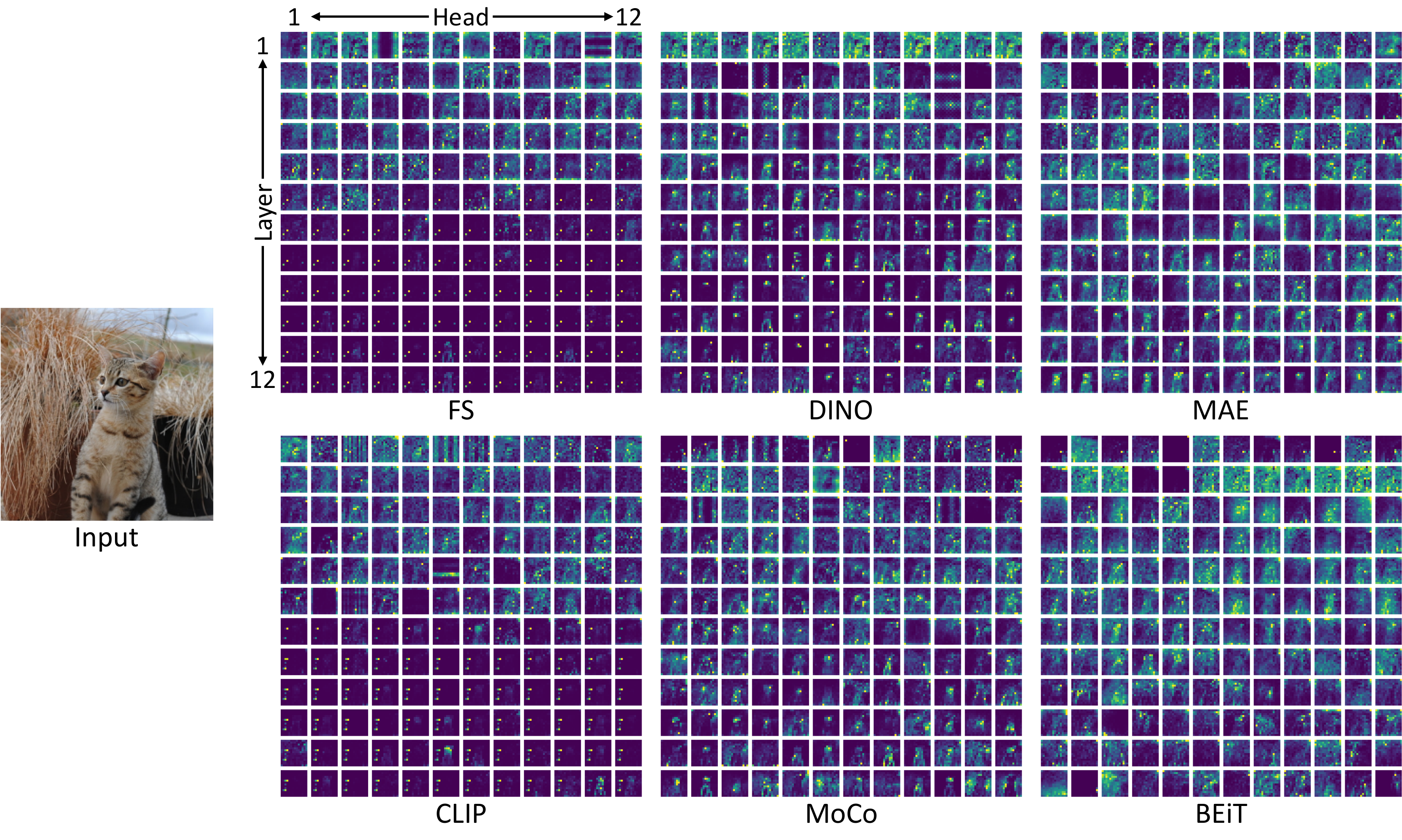}
    \end{subfigure}
    \caption{Visualizing all CLS token attention maps for all heads and all layers in ViT B/16 models for a single input image (left).}
    \label{fig:apdx-single-sample-extra-1}
\end{figure*}

\Cref{fig:apdx-single-sample-extra-2} and \Cref{fig:apdx-single-sample-extra-1} provide additional visualizations of CLS token attention maps in ViT-B/16 models for single input images. These visualizations show all layers (rows) and all heads (columns). From these views, we can see clear signs of the Sparse Repeating Attention Patterns in the mid-to-late layers of the FS and CLIP models. These patterns are strongly repeated across the head and layer axes. Note that the specific token positions that give strong activations are different for the different input images.

\begin{figure*}[h!]
    \centering
    \includegraphics[width=\linewidth]{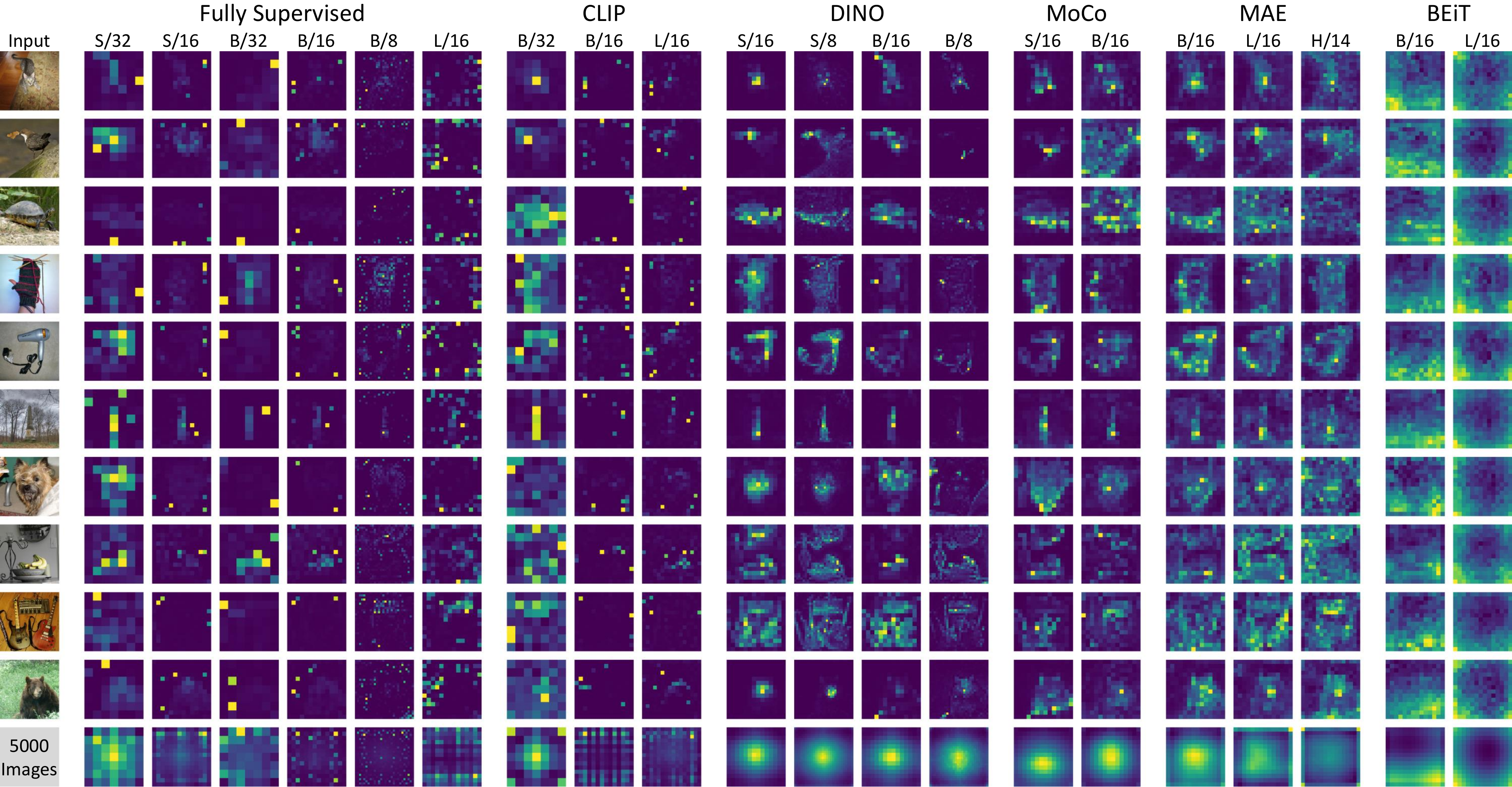}
    \caption{\textbf{Sample CLS token attention maps for a wide range of ViT variants.} For each ViT and input image, we show the the attention map of the CLS token of the first head of the final layer. The bottom row shows the averaged activation over 5000 ImageNet images.}
    \label{fig:apdx-single-img-samples}
\end{figure*}

\Cref{fig:apdx-single-img-samples} shows one ViT attention map per model for $10$ sample images over a wide array of ViT variants. The final row displays the average attention over $5000$ images. The head selected is the first head of the final layer of each ViT. For the explicitly supervised models, FS and CLIP, we again see mainly Sparse Repeating Attention Patterns. However, for the models with patch size 32, we also see some attention on object-centric regions. This holds true for FS S/32, FS B/32, and CLIP B/32. Because these models use larger patches, their token grids are a quarter of the size, making these models four times narrower than the models with patch size 16. The fact that Sparse Repeating Attention Patterns do not emerge as strongly for these smaller models may suggest that they are an indicator of overfitting in ViTs. For the other FS and CLIP models, we sometimes see faint traces of salient objects highlight in the attention maps, but this occurs alongside the Sparse Repeating Attention Patterns. For DINO, MoCo, and MAE, all models produce attention maps that tend to align well with the salient object. For BEiT, the attention maps do not correlate well. As we previously noted, the final layers of BEiT must serve as a built-in decoder, which may explain why its final layers are dissimilar to DINO, MoCo, and MAE.

\begin{figure}[t]
    \centering
    \includegraphics[width=\linewidth]{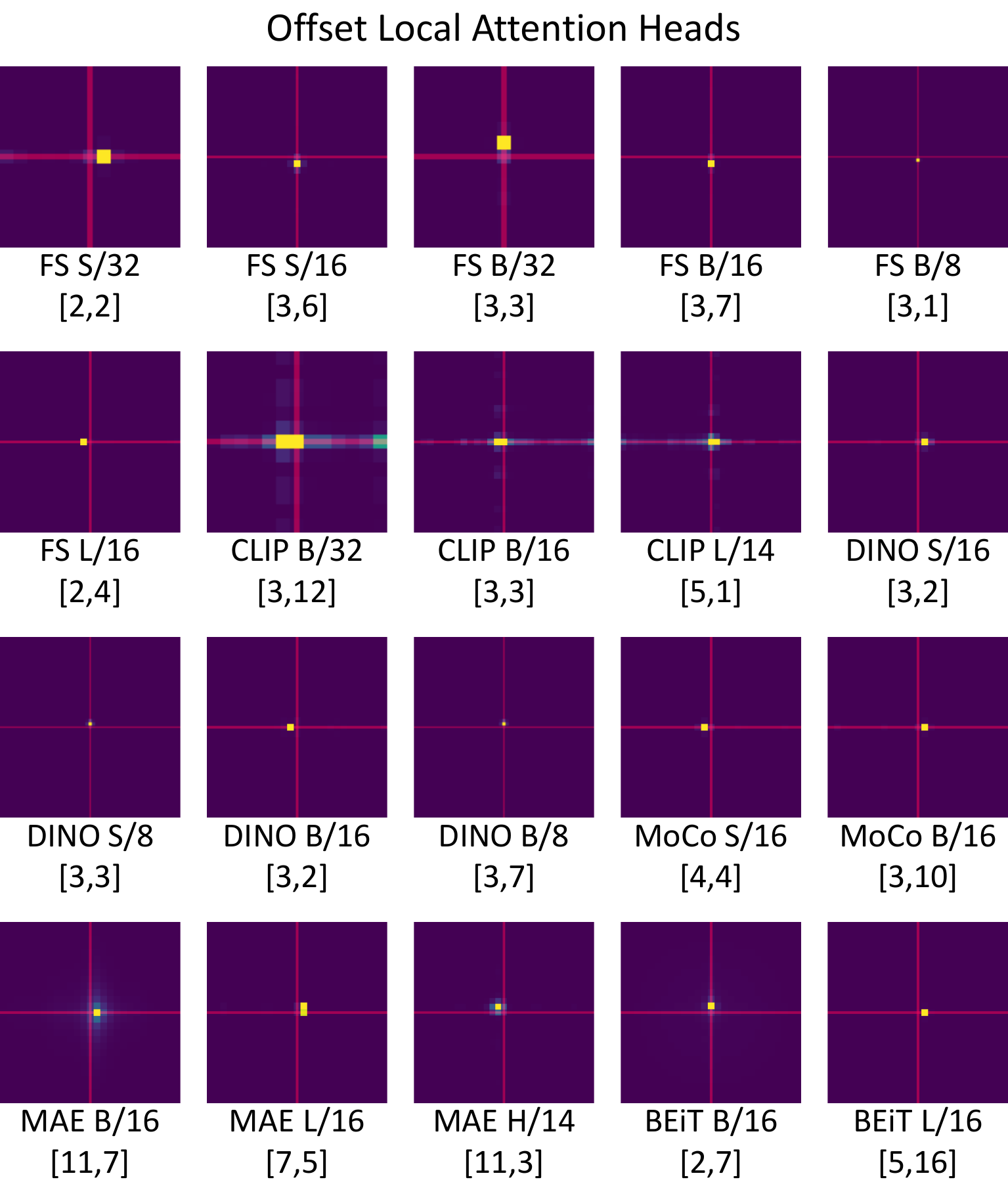}
    \caption{\textbf{Examples of Offset Local Attention Heads in all ViT Variants.} We find that all ViTs examined learn to use Offset Local Attention Heads, and some larger models even use ones with diagonal offsets. Midlines are drawn in red as a visual aid.}
    \label{fig:apdx-offset-samples}
\end{figure}

We find that every model learns instances of Offset Local Attention Heads, and some larger models even have ones with a diagonal offset. We present one example per model in \Cref{fig:apdx-offset-samples}, but be aware that all models have many Offset Local Attention Heads with different offsets. For the explicitly and contrastively supervised models, Offset Local Attention Heads typically only occur in the first 3 to 6 layers, but for the reconstruction-based models, MAE and BEiT, we can find them in deeper layers too.

\begin{figure*}[t]
    \centering
    \begin{subfigure}[b]{\linewidth}
        \centering
        \includegraphics[width=0.92\textwidth]{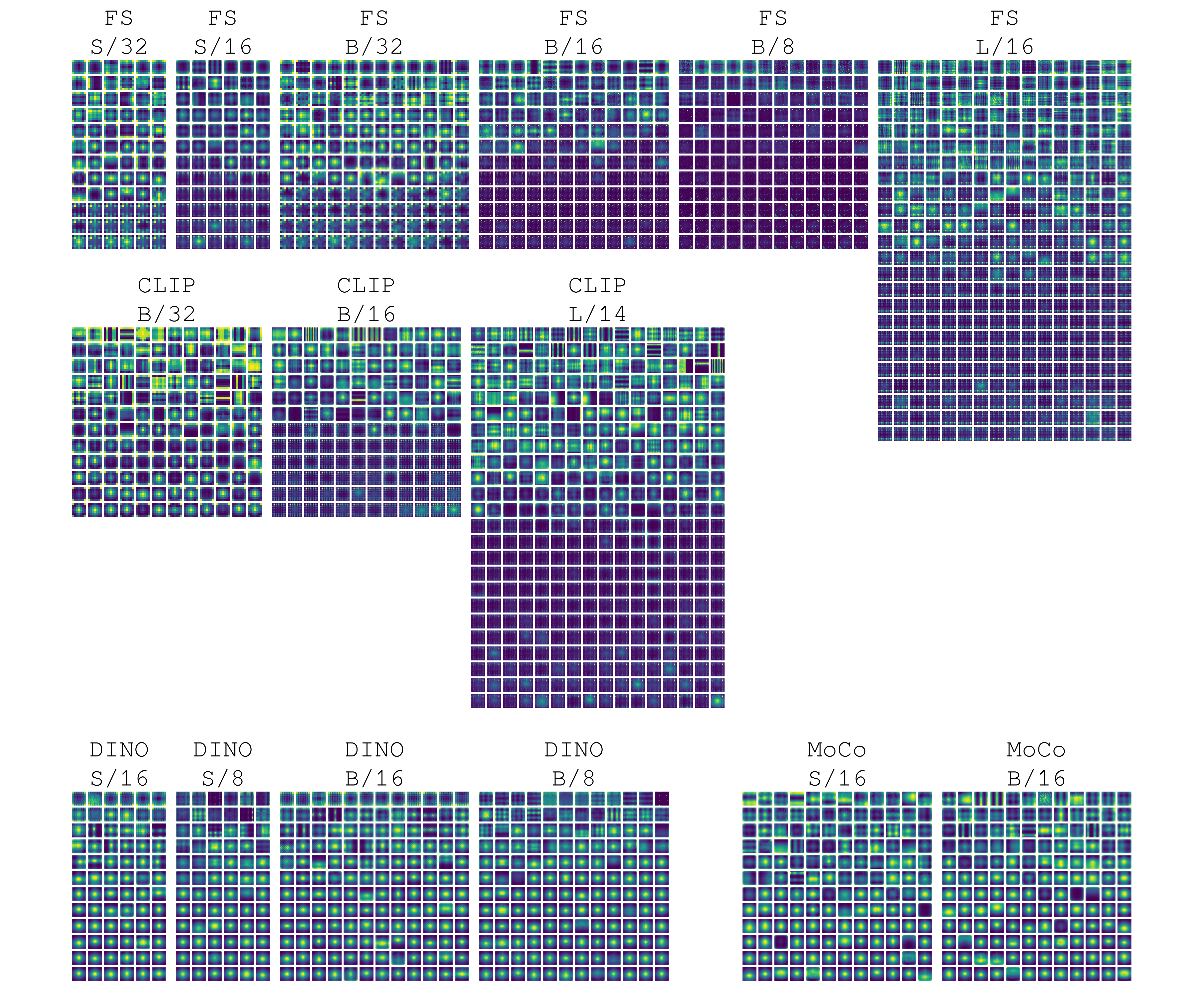}
    \end{subfigure}

    \vspace{1em}
    \begin{subfigure}[b]{\linewidth}
        \centering
        \includegraphics[width=0.92\textwidth]{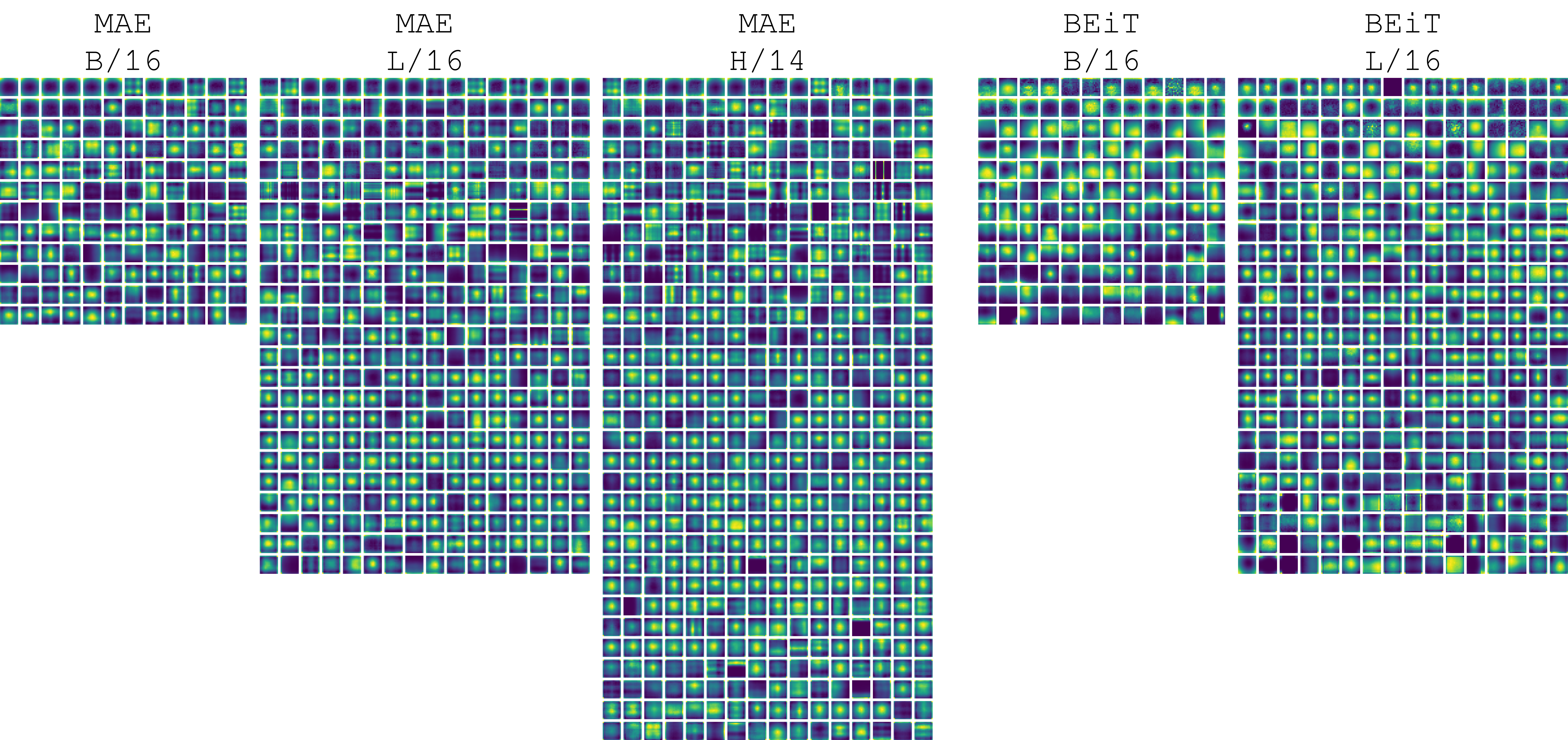}
    \end{subfigure}
    
    \caption{Average CLS token attention over 5000 images for every head of every ViT Variant.}
    \label{fig:big-cls}
\end{figure*}

\begin{figure*}[t]
    \centering
    \begin{subfigure}[b]{\linewidth}
        \centering
        \includegraphics[width=0.92\textwidth]{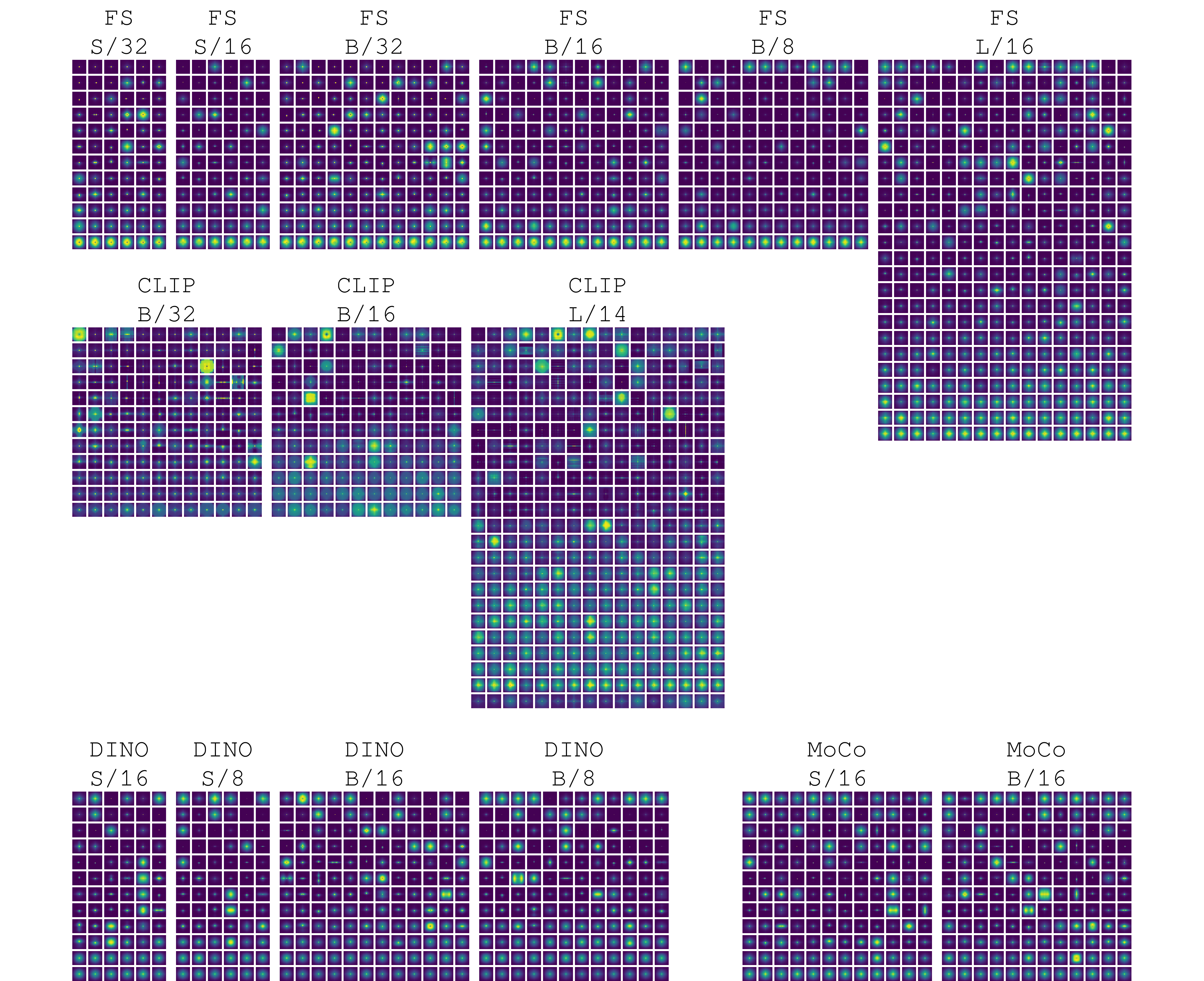}
    \end{subfigure}

    \vspace{1em}
    \begin{subfigure}[b]{\linewidth}
        \centering
        \includegraphics[width=0.92\textwidth]{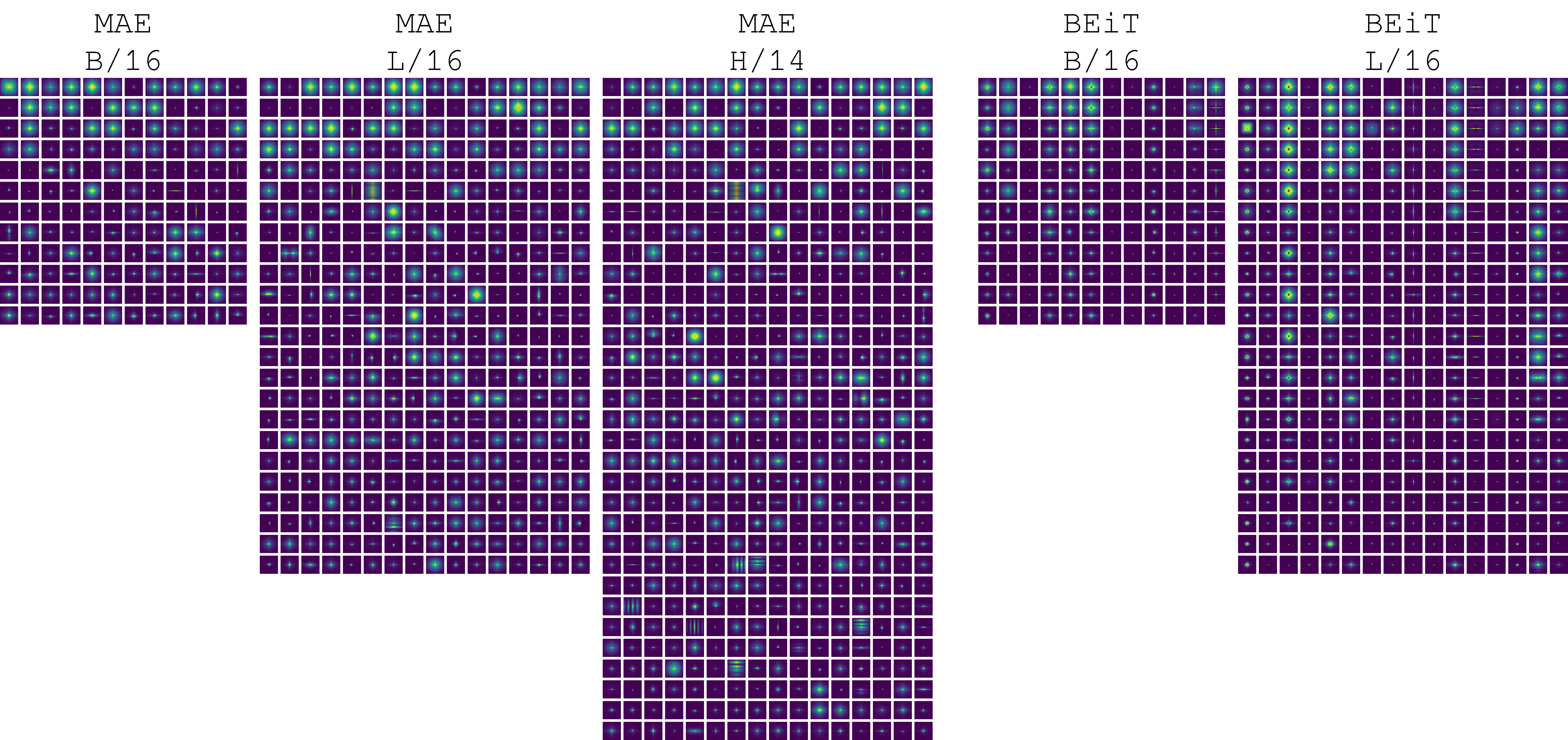}
    \end{subfigure}
    
    \caption{Aligned Aggregated Attention Maps over 5000 images for every head of every ViT Variant.}
    \label{fig:big-aaam}
\end{figure*}

To provide the reader a complete view of the size and number of attention heads in each ViT, we present two plots that visualize all layers and all heads of all 20 ViTs. \Cref{fig:big-cls} presents the average CLS token attention over 5000 sample images. When viewed this way, it is clear how widespread the Sparse Repeating Activation Patterns are over all of the mid-to-later layer heads of the FS and CLIP models. For FS S/32 and B/32 we can see clear signs of Sparse Repeating Activation Patterns, but for CLIP B/32 we instead see far more centered circular blobs, similar to those we observe in the later layers of DINO and MoCo. We also note that some of the early-layer heads (layers 1-3) of the DINO models produce semi-repetitive grid-like patterns that somewhat resemble the Sparse Repeating Attention Patterns seen in CLIP and FS. However, on closer inspection, we believe these heads represent a different phenomenon. The attention patterns in these layers have more variations across heads and layers, and they are not identically repeated as is seen in the FS and CLIP models. Also, these heads come in the early layers, not the mid-to-late layers. For this reason, we hypothesize these heads are learning to extract an initial sparse down-sampling of the image, which would be especially beneficial for the DINO models with patch size 8 due to their much larger token counts.
Finally, \Cref{fig:big-aaam} shows the Aligned Aggregated Attention Maps for all the spatial tokens. This view highlights the great variety of local attention heads used in each ViT.
These figures are best viewed digitally and in color.

\subsection{Attention Distance for ViT Variants}
\label{sec:apdx-att-dist}

\begin{figure*}[t]
    \centering
    \begin{subfigure}[b]{\linewidth}
        \centering
        \includegraphics[width=0.88\textwidth]{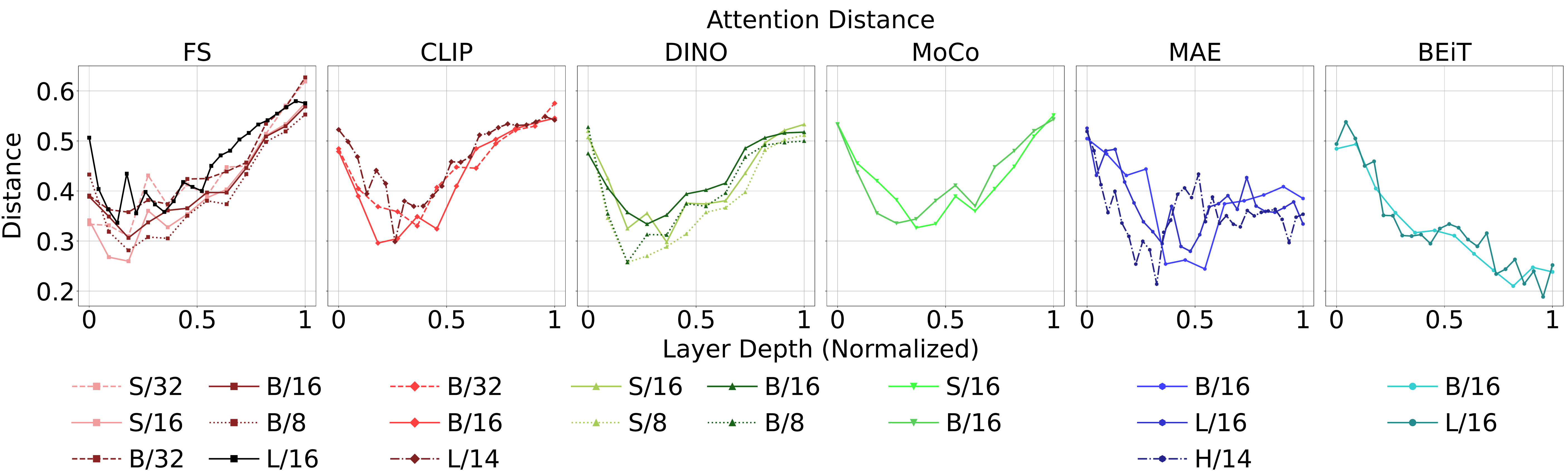}
    \end{subfigure}
    \caption{\textbf{Average Attention Distance for all ViT Variants.} Results are plotted against the normalized layer depth.}
    \label{fig:apdx-att-dist}
\end{figure*}

In \Cref{fig:apdx-att-dist} we present the Average Attention Distances per-layer for the full ViT collection, broken out by supervision type. For our distance computations, we have normalized the distances such that the token grids are within a $1\times1$ square, which allows us to compare models with different patch sizes and hence different token grid sizes. We see that the trends of local-vs-global processing order is consistent within supervision groups. For FS, CLIP, DINO, and MoCo we again see an intial high distance, a dip to lower distances, and an increase again in the later layers. Meanwhile, for MAE we again see lower attention distance in the later layers. This result shows that the order of local-vs-global information processing in a ViT is primarily impacted by the method of supervision and is largely unaffected by changes in architecture size and patch size.

\clearpage
\clearpage

\subsection{Attention Salience for ViT Variants}
\label{sec:apdx-att-sal}

In this section, we present additional experimental details for our Attentional Saliency Analysis, followed by results for the full ViT collection. In addition to PartImageNet~\citep{he2021partimagenet}, we also performed this analysis with COCO~\citep{lin2014microsoft}, however the results are extremely similar for the two datasets. The PartImageNet dataset contains $11$ superclasses, whose members contain similar part structures (biped, quadruped, car). When sampling from PartImageNet, we take $500$ samples per superclass, or all samples for ones with less than $500$. Within superclasses, we evenly sample from each of the subclasses, or if a subclass is fully sampled we continue to sample evenly from the remaining classes. This yields a mostly balanced collection of $5294$ images. PartImageNet divides different subclasses into the train, validation, and test partitions, but for our analysis we work with all three partitions together. For COCO, we simply sample the first $5000$ images of the 2017 validation set.

\begin{figure*}[t]
    \centering
    \begin{subfigure}[b]{\linewidth}
        \centering
        \includegraphics[width=0.88\textwidth]{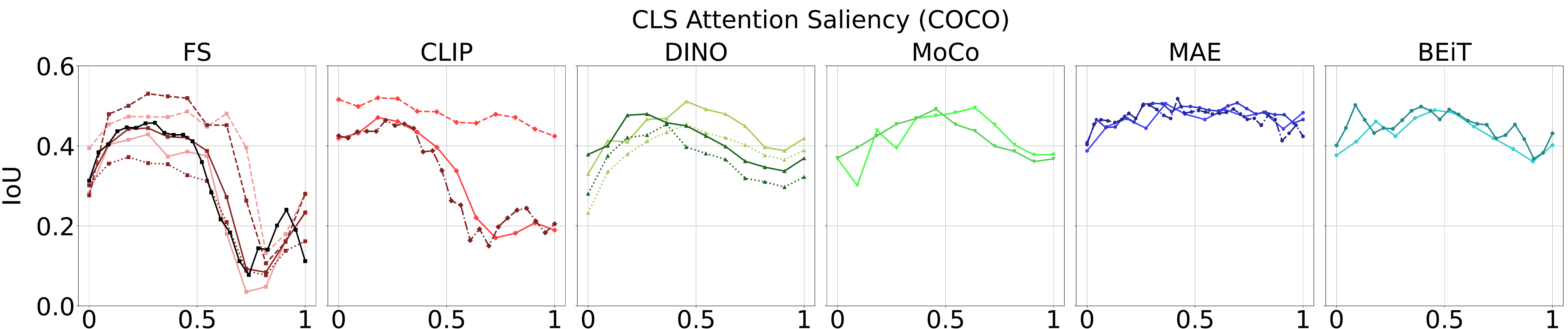}
    \end{subfigure}

    \vspace{0.5em}
    
    \begin{subfigure}[b]{\linewidth}
        \centering
        \includegraphics[width=0.88\textwidth]{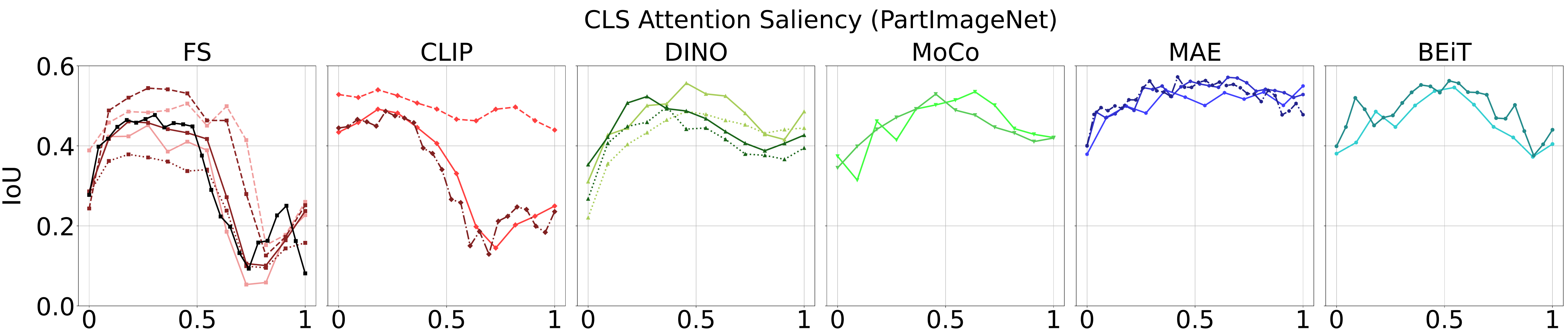}
    \end{subfigure}

    \vspace{0.5em}

    \begin{subfigure}[b]{\linewidth}
        \centering
        \includegraphics[width=0.88\textwidth]{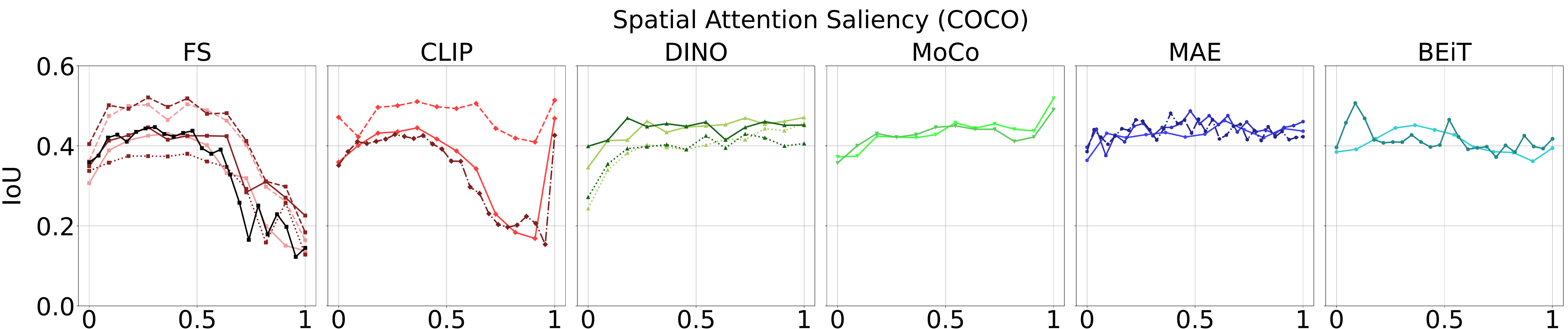}
    \end{subfigure}

    \vspace{0.5em}
    
    \begin{subfigure}[b]{\linewidth}
        \centering
        \includegraphics[width=0.88\textwidth]{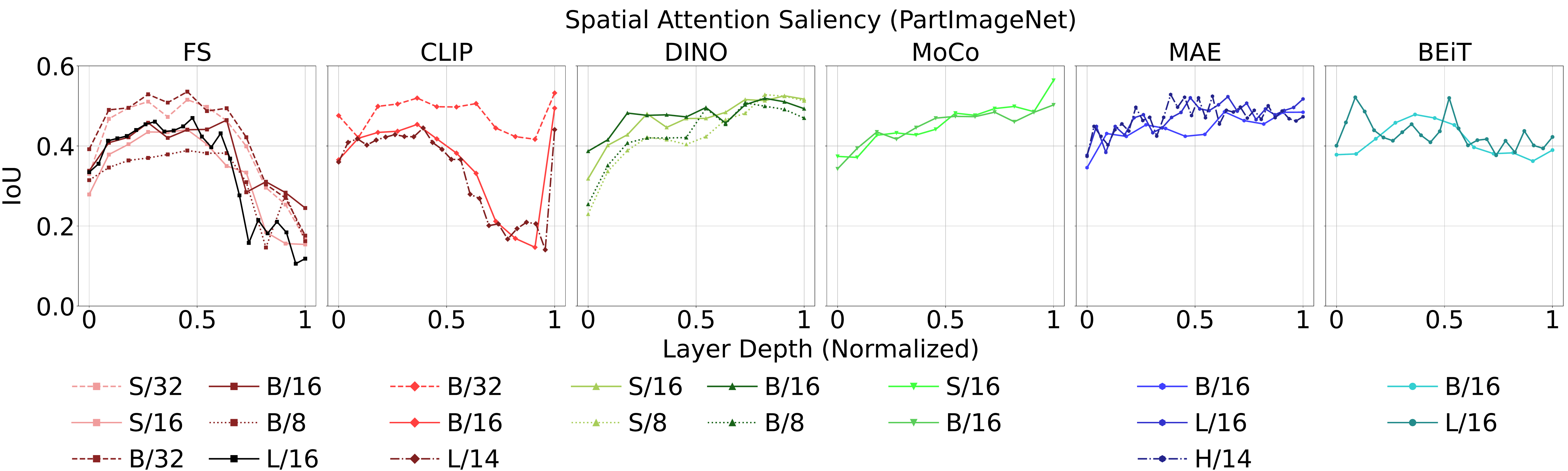}
    \end{subfigure}
    \caption{\textbf{Attention alignment with salient image content for all ViT Variants.} Results are shown for both CLS token attention and average spatial token attention on both COCO and PartImageNet. We see that the results are highly similar for the two datasets.}
    \label{fig:apdx-att-sal}
\end{figure*}

\Cref{fig:apdx-att-sal} summarizes the results for both PartImageNet and COCO with both CLS token attention and average spatial token attention. The patterns of scores are very similar for PartImageNet and COCO. We see that the explicitly supervised methods again face a decrease in IoU in the mid-to-late layers with the emergence of Sparse Repeating Attention Patterns. This is with the exception of CLIP B/32 which has a much better IoU in the later layers. This result matches \Cref{sec:apdx-att-vis}, where we observed that CLIP B/32 is less impacted by the Sparse Repeating Attention Patterns. For the other self-supervised methods, we see that the same general trends hold, usually with slightly higher IoUs from larger models or models with smaller patches. 

\begin{figure*}[h!]
    \centering
    \begin{subfigure}{0.495\textwidth}
        \centering
        \includegraphics[width=\textwidth]{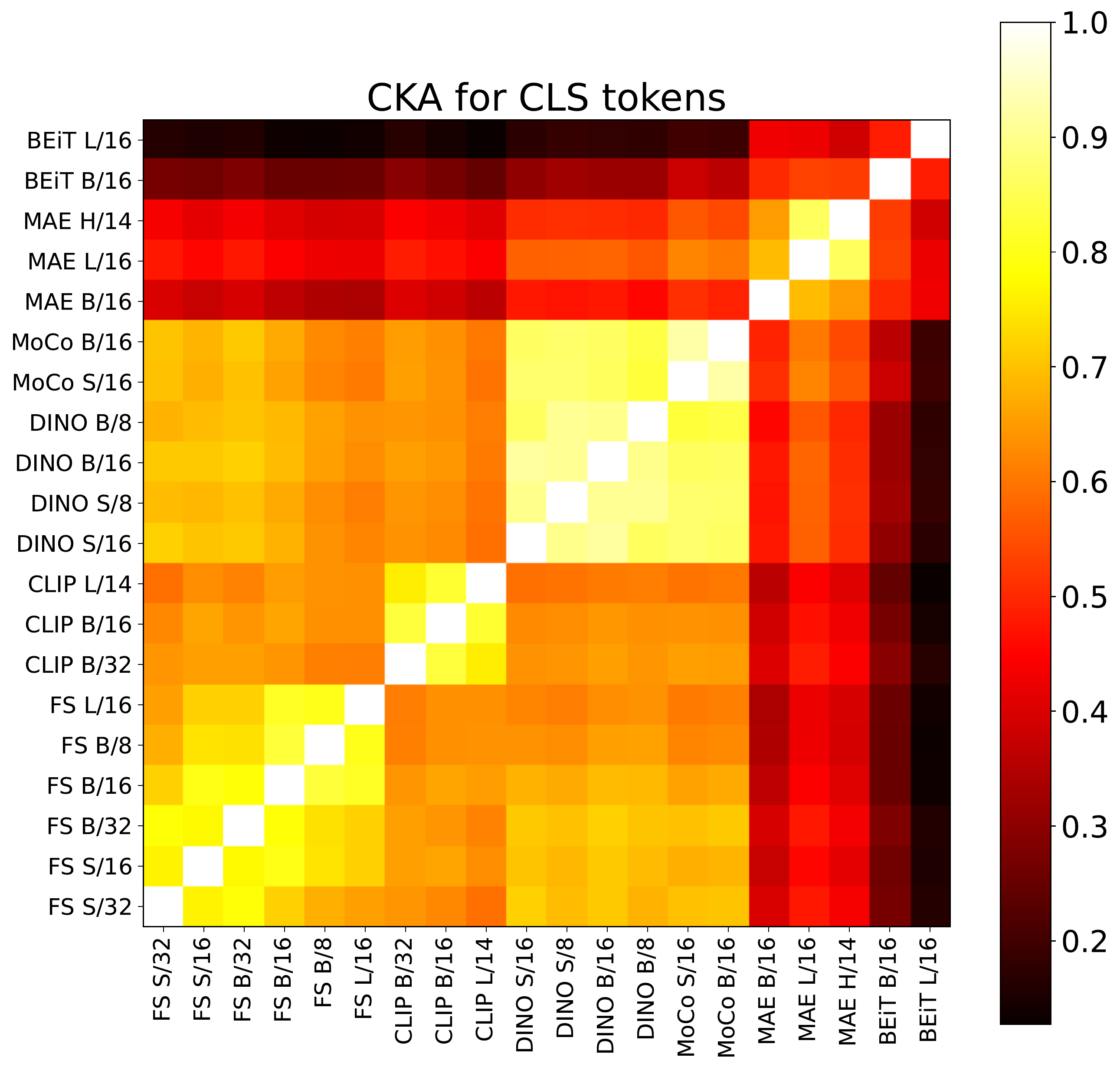}
    \end{subfigure}
    \begin{subfigure}{0.495\textwidth}
        \centering
        \includegraphics[width=\textwidth]{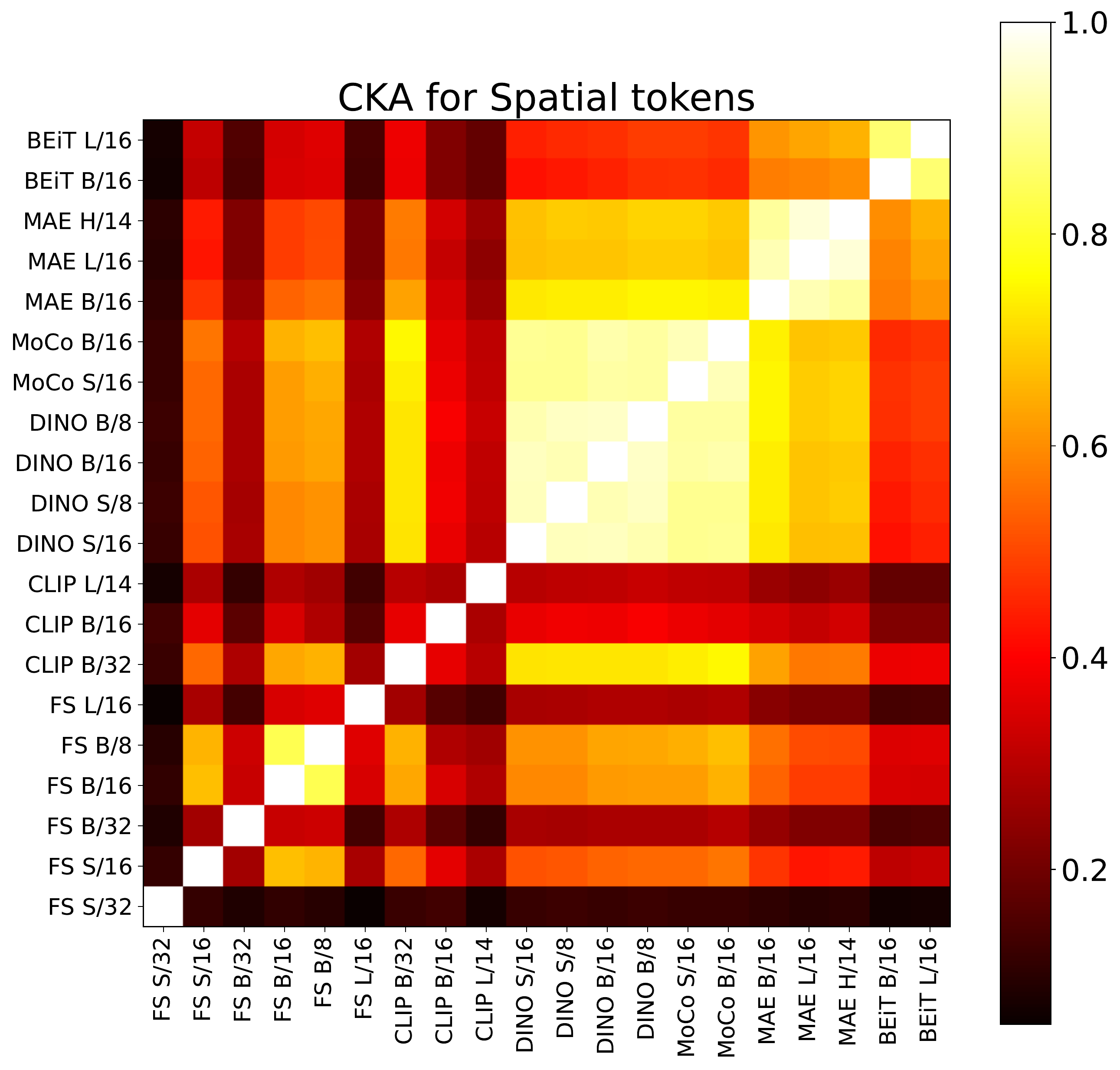}
    \end{subfigure}
    \caption{CKA similarity between final layer features of different ViTs for their CLS tokens (left) and spatial tokens (right).}
    \label{fig:cka-lastlayer-all}
\end{figure*}

\section{Feature Analysis}

\subsection{The CKA metric}

In our Feature Analysis section, we compare learned representations through Centered Kernel Alignment (CKA)~\citep{cortes2012algorithms, kornblith2019similarity}, which is able to align and rescale neural features to enable similarity measurements. Specifically, we used batched CKA~\citep{nguyen2020wide}, which can be represented as follows:

\begin{equation}
    {CKA}_{batched} = \dfrac{\frac{1}{k}\sum_{i=1}^{k}HSIC_1(\mathbf{X}_{i}\mathbf{X}^{T}_{i}, \mathbf{Y}_{i}\mathbf{Y}^{T}_{i})}{(\splitdfrac{\sqrt{\frac{1}{k}\sum_{i=1}^{k}HSIC_1(\mathbf{X}_{i}\mathbf{X}^{T}_{i}, \mathbf{X}_{i}\mathbf{X}^{T}_{i})}}{*\sqrt{\frac{1}{k}\sum_{i=1}^{k}HSIC_1(\mathbf{Y}_{i}\mathbf{Y}^{T}_{i}, \mathbf{Y}_{i}\mathbf{Y}^{T}_{i})}})},
\end{equation}

where $X_i$ and $Y_i$ are the feature representation matrices of the $i^{th}$ batch from the two models, $k$ is the number of batches and $HSIC_1$ is as follows:

\begin{dmath}
    HSIC_1(\mathbf{K},\mathbf{L}) = \frac{1}{n(n-3)}\left(tr(\mathbf{\hat{K}\hat{L}}) + \frac{\mathbf{1^T\hat{K}11^T\hat{L}1}}{(n-1)(n-2)} - \frac{2}{n-2}\mathbf{1^T\hat{K}\hat{L}1} \right),
\end{dmath}
where $\mathbf{\hat{K}}$ and $\mathbf{\hat{L}}$ are $\mathbf{K,L}$ with diagonals set to $0$ and $n$ is the batch size. We use the implementation from~\citet{subramanian2021torch_cka} for our analysis.

\subsection{Last Layer Comparisons for ViT Variants}
\label{sec:apdx-feat-cka}

In the main paper we analyzed the last layer CKA between the CLS and spatial tokens across the B/16 models separately. Here we expand our analysis to the wider collection of ViT variants.
As can be seen from \Cref{fig:cka-lastlayer-all} (left), for the CLS tokens, similar supervision strategies create similar representations. Groups emerge with DINO and MoCo forming one subset while MAE and BEiT form another. The FS and CLIP models form their own sub-groups. Some of the FS models also show comparatively high similarity with MoCo and DINO models. Again we see that the MAE CLS representations have a moderate similarity with explicitly and constrastively supervised methods.

From \Cref{fig:cka-lastlayer-all} (right), we see that the internal similarity within the FS and CLIP groups are more fragmented. Meanwhile, the self-supervised models show more consistency, having higher spatial feature similarity within and between self-supervision methods.
DINO and MoCo show very high similarity amongst themselves due to their similar training methods.
The MAE spatial features also show high similarity to those of DINO and MoCo.
BEiT shows comparatively high similarity with these other self-supervised methods.
FS and CLIP  are not too similar with each other or with the other models with the exception of CLIP B/32 and a few FS models like B/16 and B/8 which show comparatively high similarity with DINO and MoCo models. This separation of CLIP and FS can be attributed to the supervision which is applied to only the CLS token, which may make their final layer spatial representations less consistent.

\begin{figure}[t]
    \centering
\begin{subfigure}{0.155\textwidth}
  \includegraphics[width=\textwidth]{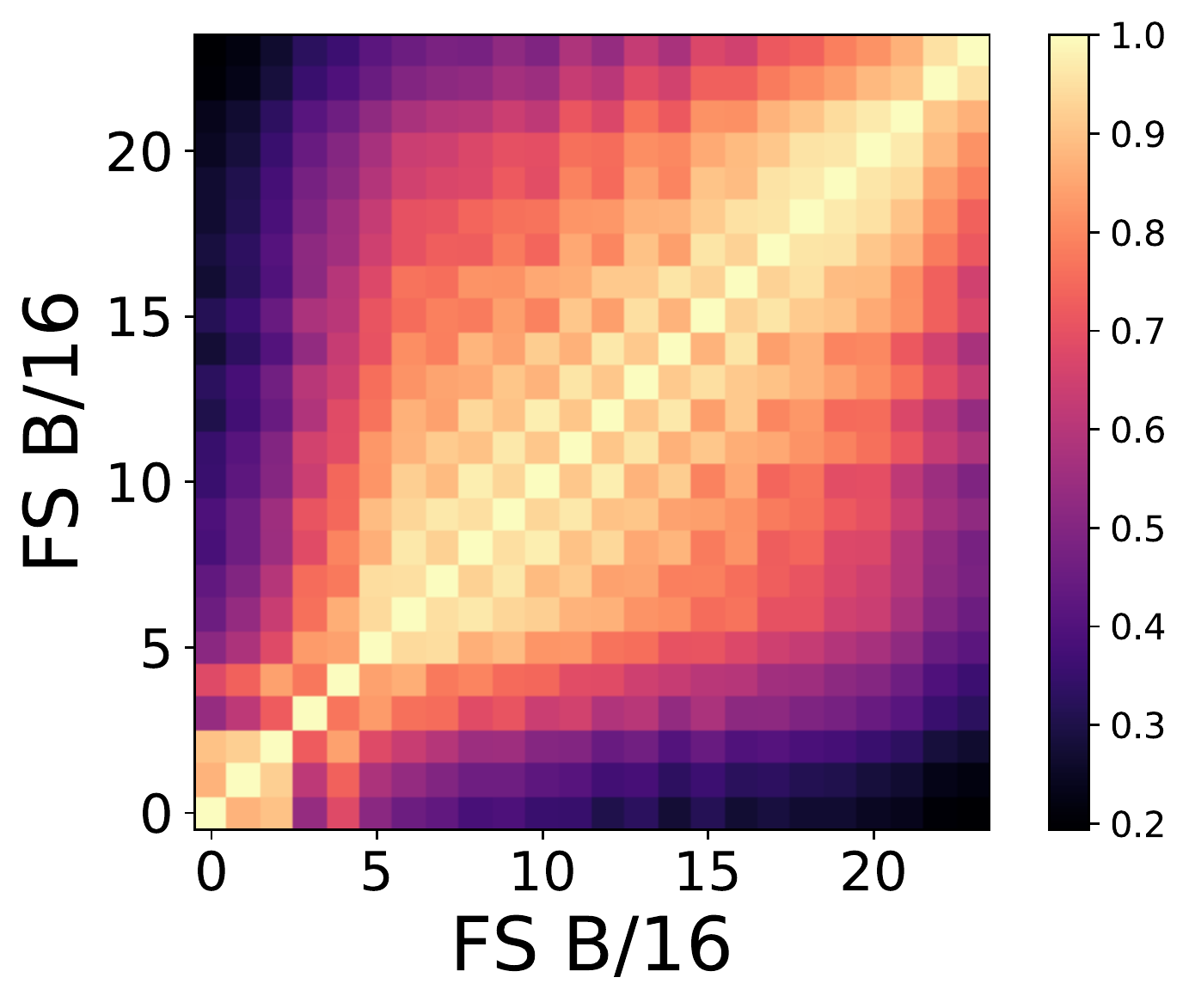}
\end{subfigure}
\begin{subfigure}{0.155\textwidth}
  \includegraphics[width=\textwidth]{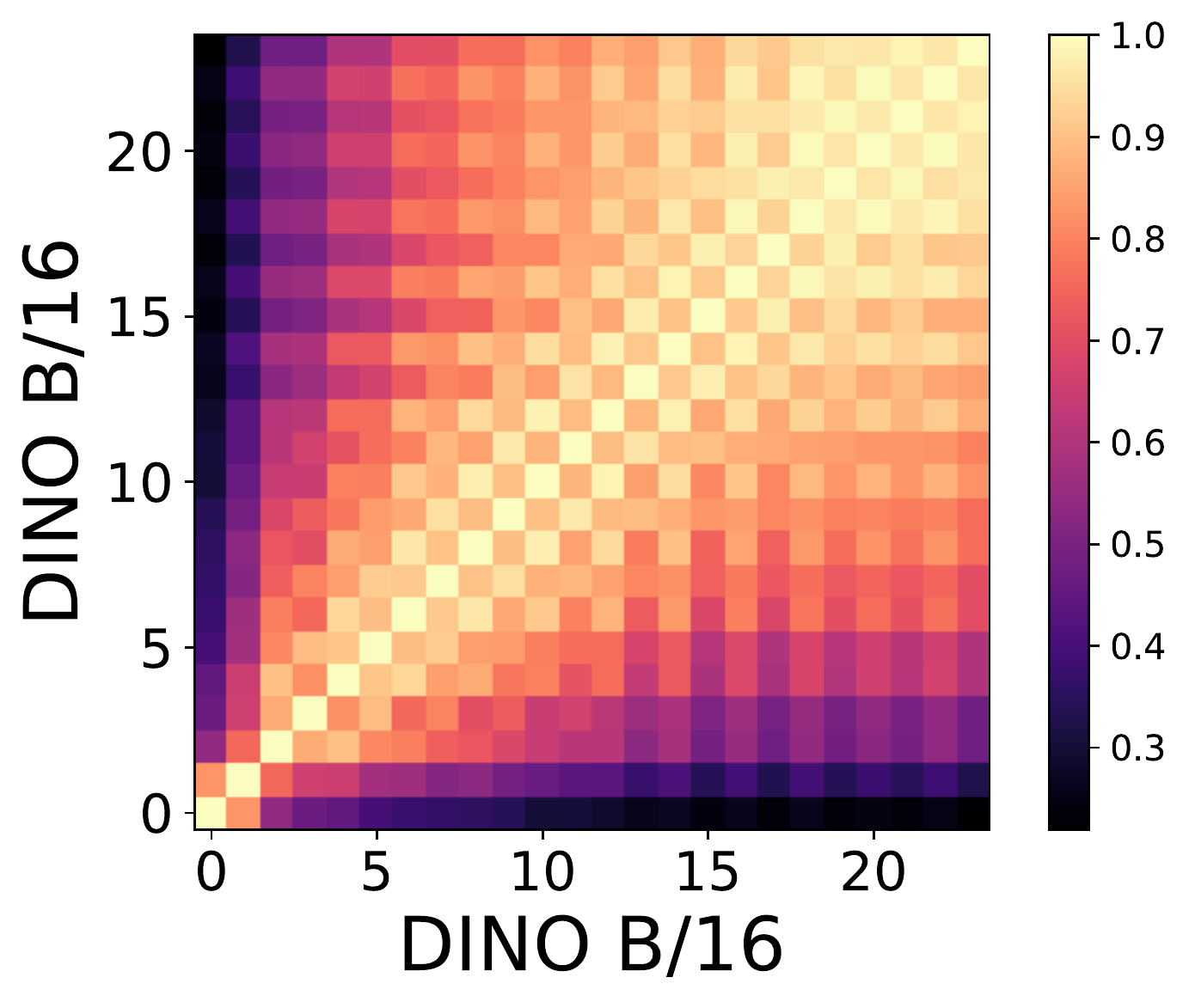}
\end{subfigure}
\begin{subfigure}{0.155\textwidth}
  \includegraphics[width=\textwidth]{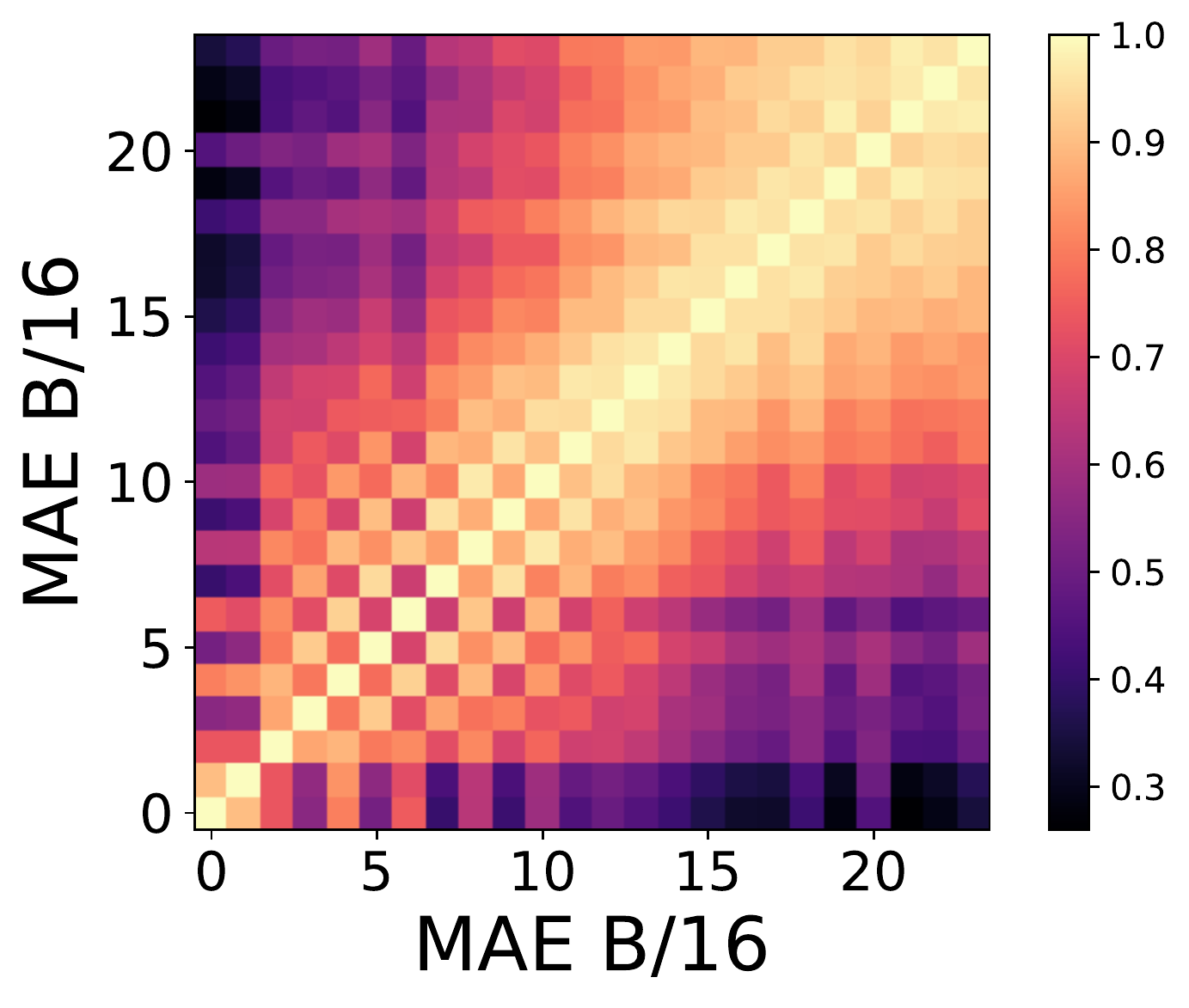}
\end{subfigure}

\medskip
\begin{subfigure}{0.155\textwidth}
  \includegraphics[width=\textwidth]{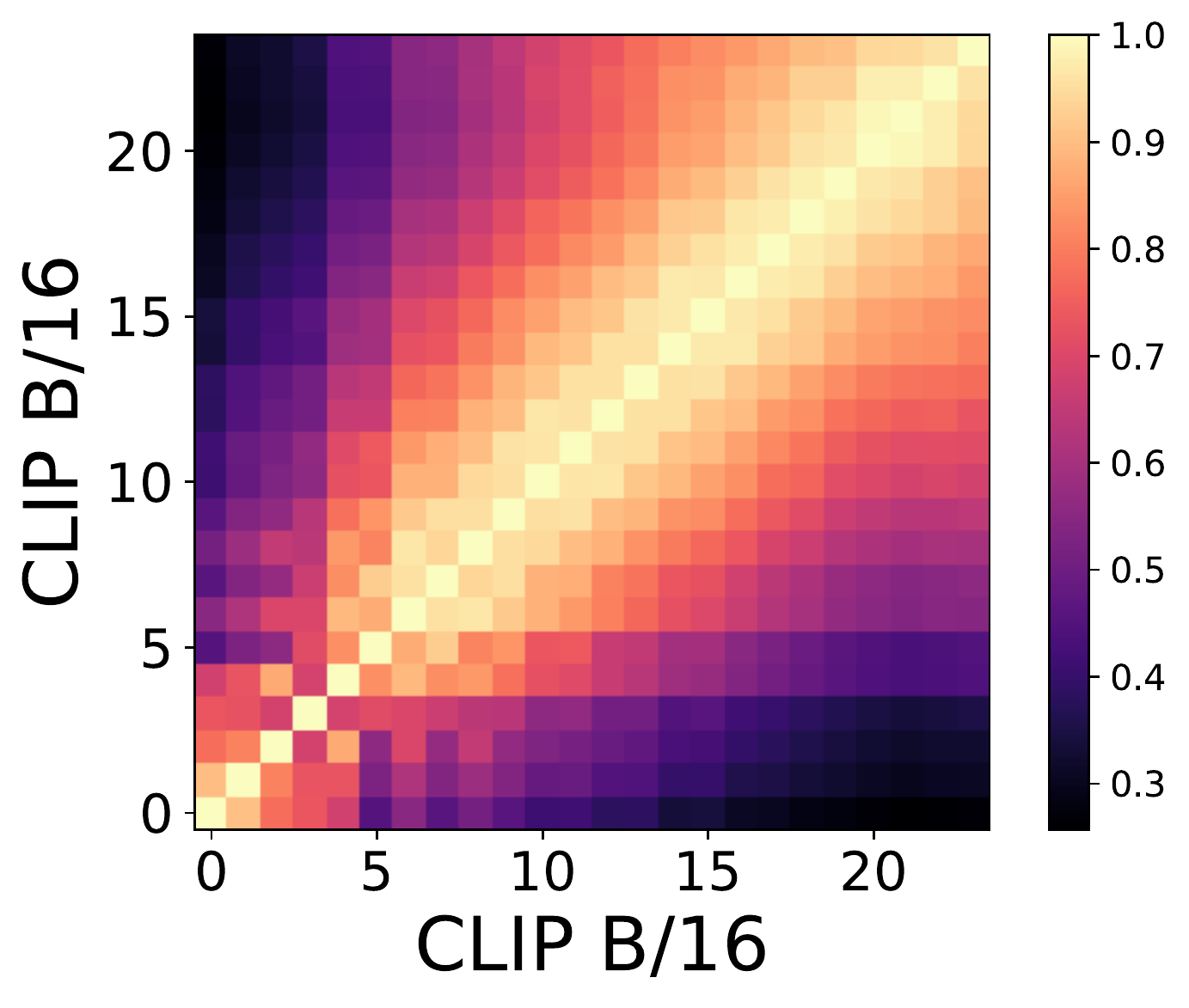}
\end{subfigure}
\begin{subfigure}{0.155\textwidth}
  \includegraphics[width=\textwidth]{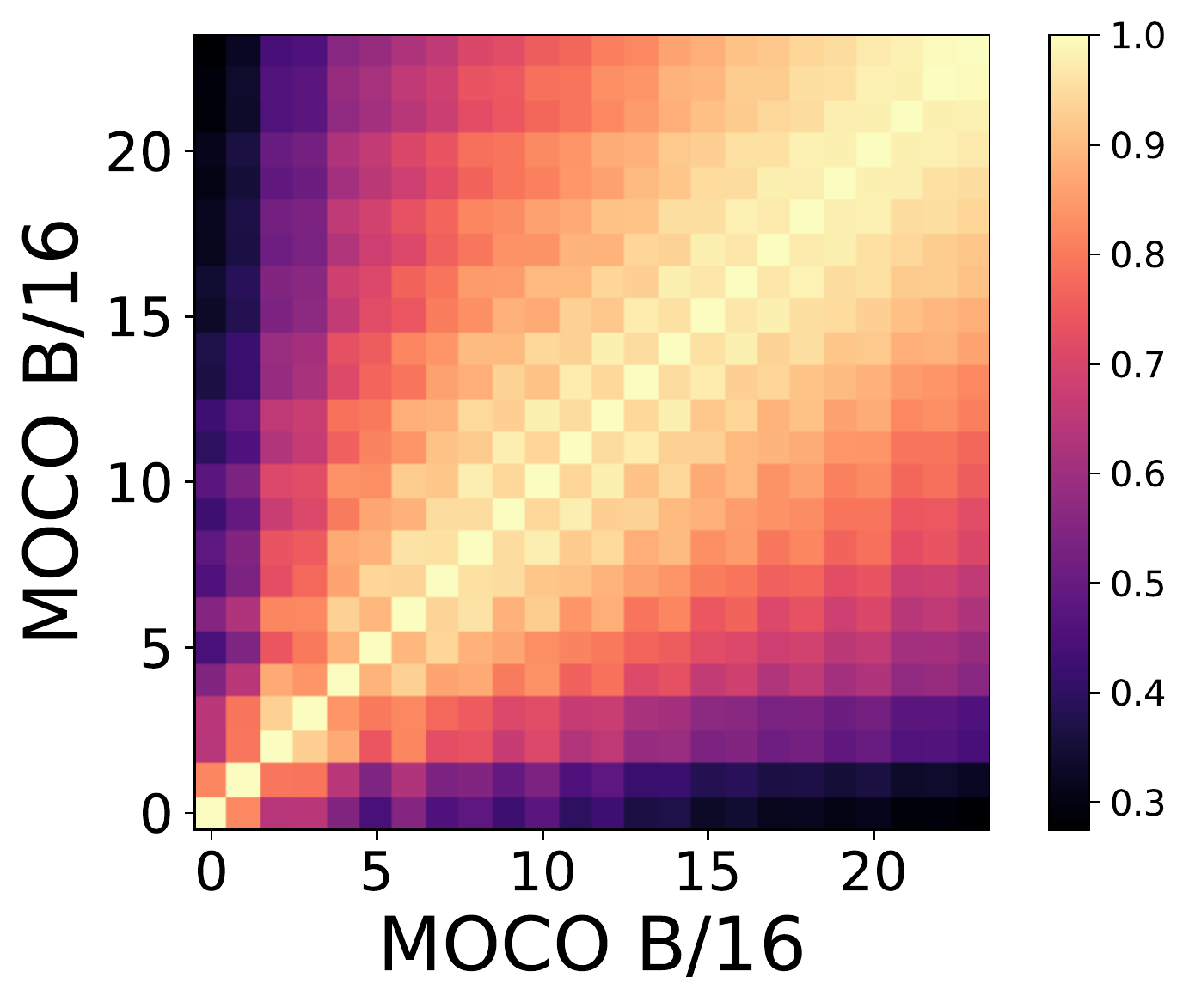}
\end{subfigure}
\begin{subfigure}{0.155\textwidth}
  \includegraphics[width=\textwidth]{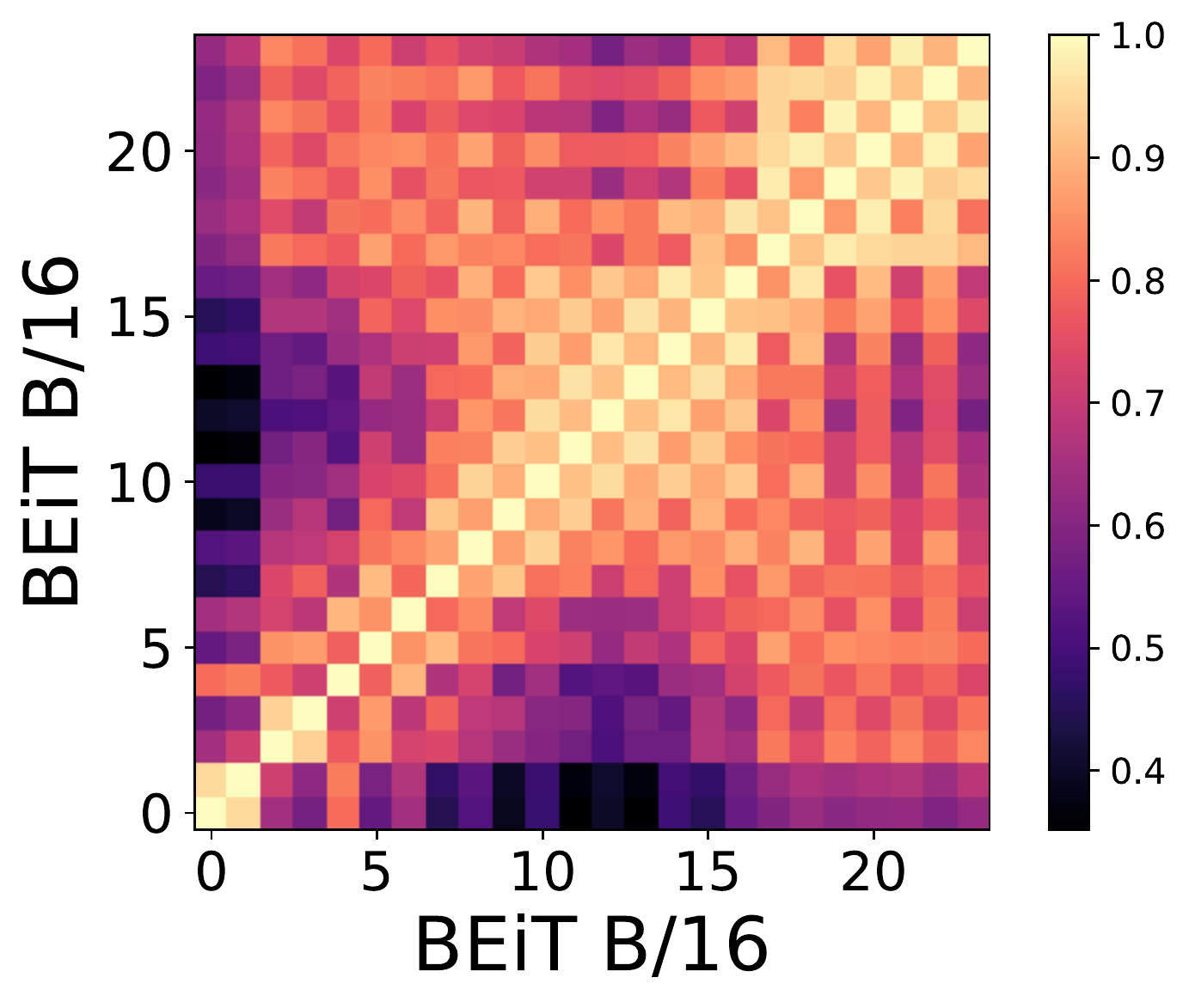}
\end{subfigure}
\caption{CKA for all ViT B/16 models showing their feature similarity across layers. Different supervision techniques result in different patterns.}
\label{fig:all_cka}
\vspace{-10pt}
\end{figure}

\subsection{Depth-Wise CKA Analysis}
\label{sec:apdx-feat-cka-depth}

\paragraph{Self-Comparison of ViT B/16 Models.} 
\Cref{fig:all_cka} shows the CKA plots across multiple layers of the same model for different training methods. For brevity and consistency we focus on the ViT B/16 models. For all the CKA plots we use the features from the batch norm layers due to their well-behaved outputs. We see that the different models show variations in the development of information. For FS, the final layers have a lower similarity to the earlier layers, as compared with CLIP, DINO, and MoCo. For MAE, we see two distinct blocks of similarity divided around the middle layer. For BEiT, we see a clear X pattern, which we analyze more in a subsequent section.

\begin{figure}[t]
    \centering
\vspace{4pt}
\begin{subfigure}{0.15\textwidth}
  \includegraphics[width=\textwidth]{apdx_plots/feat/cka/MAE-ViT-B-16-224xMAE-ViT-B-16-224.pdf}
\end{subfigure}
\begin{subfigure}{0.15\textwidth}
  \includegraphics[width=\textwidth]{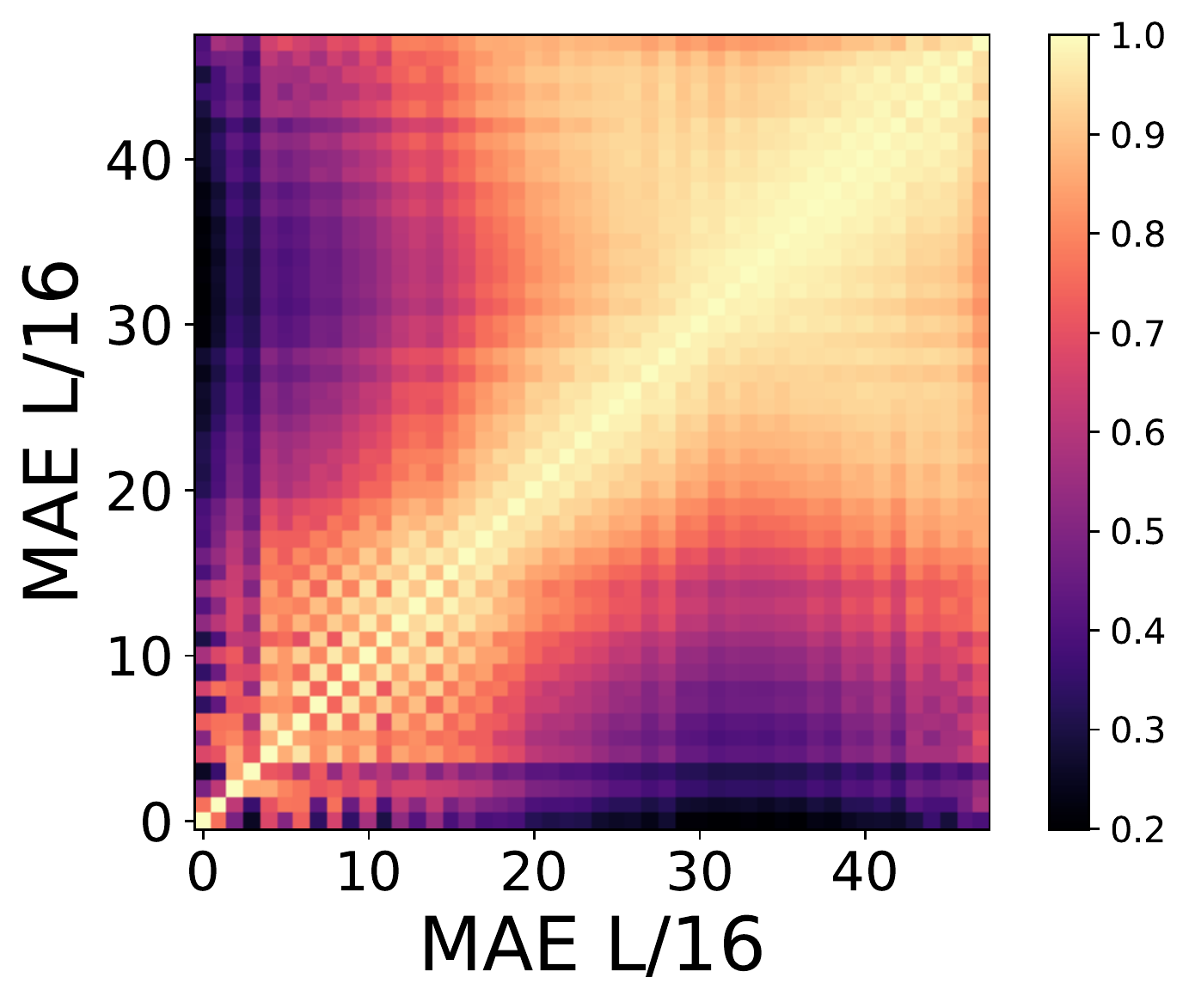}
\end{subfigure}
\begin{subfigure}{0.15\textwidth}
  \includegraphics[width=\textwidth]{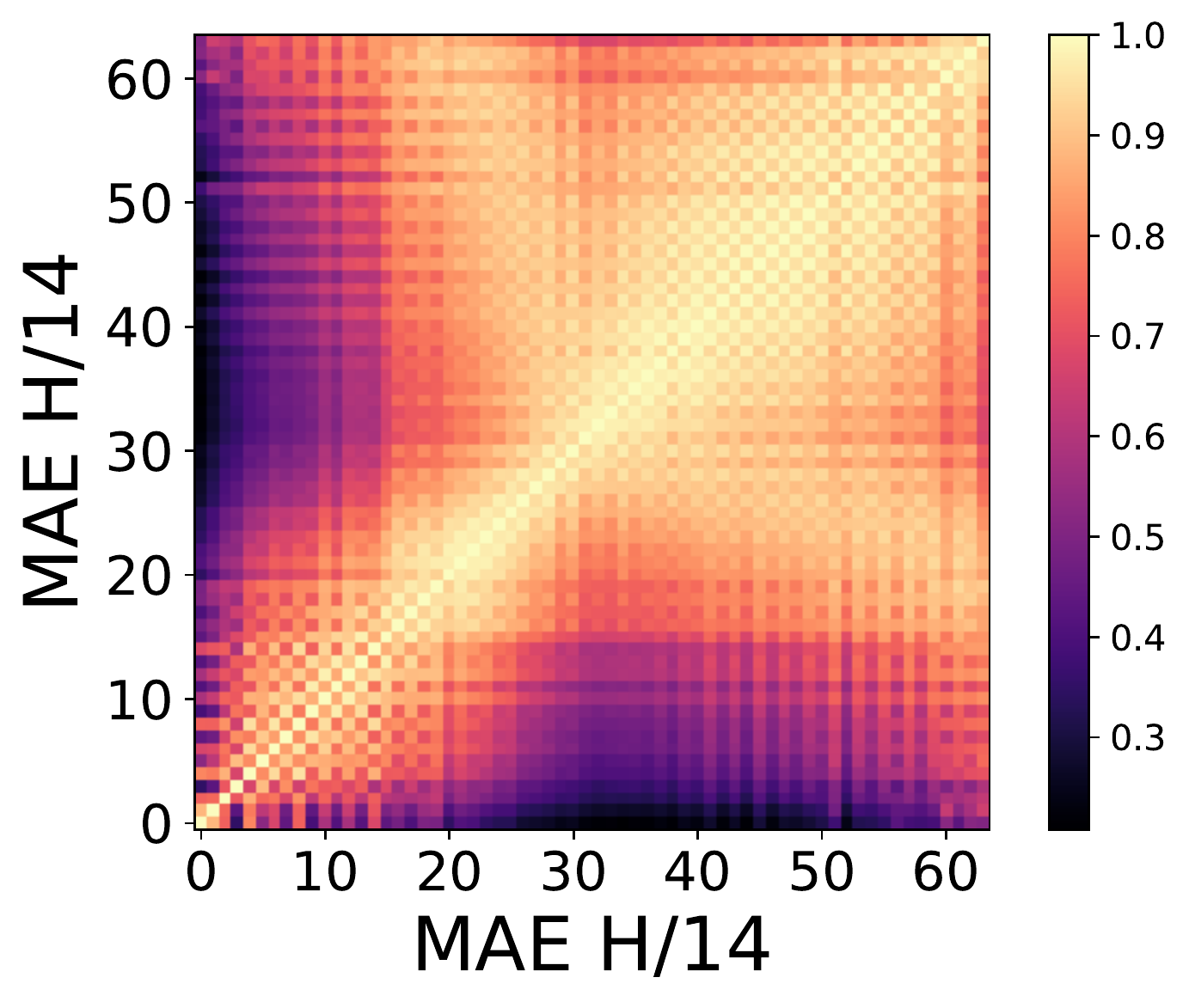}
\end{subfigure}
\caption{CKA across Base, Large and Huge MAE models.}
\label{fig:mae_self}
\vspace{10pt}
\end{figure}

\paragraph{Going deeper into MAEs.} \Cref{fig:mae_self} shows the CKA plot for MAE models Base, Large and Huge from left to right. It can be seen that as we move from a smaller model to a larger model (for example from Base to Large), the bottom left quadrant of larger models CKA matches the full CKA for the smaller model. This indicates that a larger model in this case encodes information in a similar way as the smaller model in its initial layers but ends up having more specialized later layers at the end.
A similar trend can be observed when going from Large to Huge.

\begin{figure}[t]
    \centering
\begin{subfigure}{0.15\textwidth}
  \includegraphics[width=\textwidth]{apdx_plots/feat/cka/BEIT-ViT-B-16-224xBEIT-ViT-B-16-224.pdf}
\end{subfigure}
\begin{subfigure}{0.15\textwidth}
  \includegraphics[width=\textwidth]{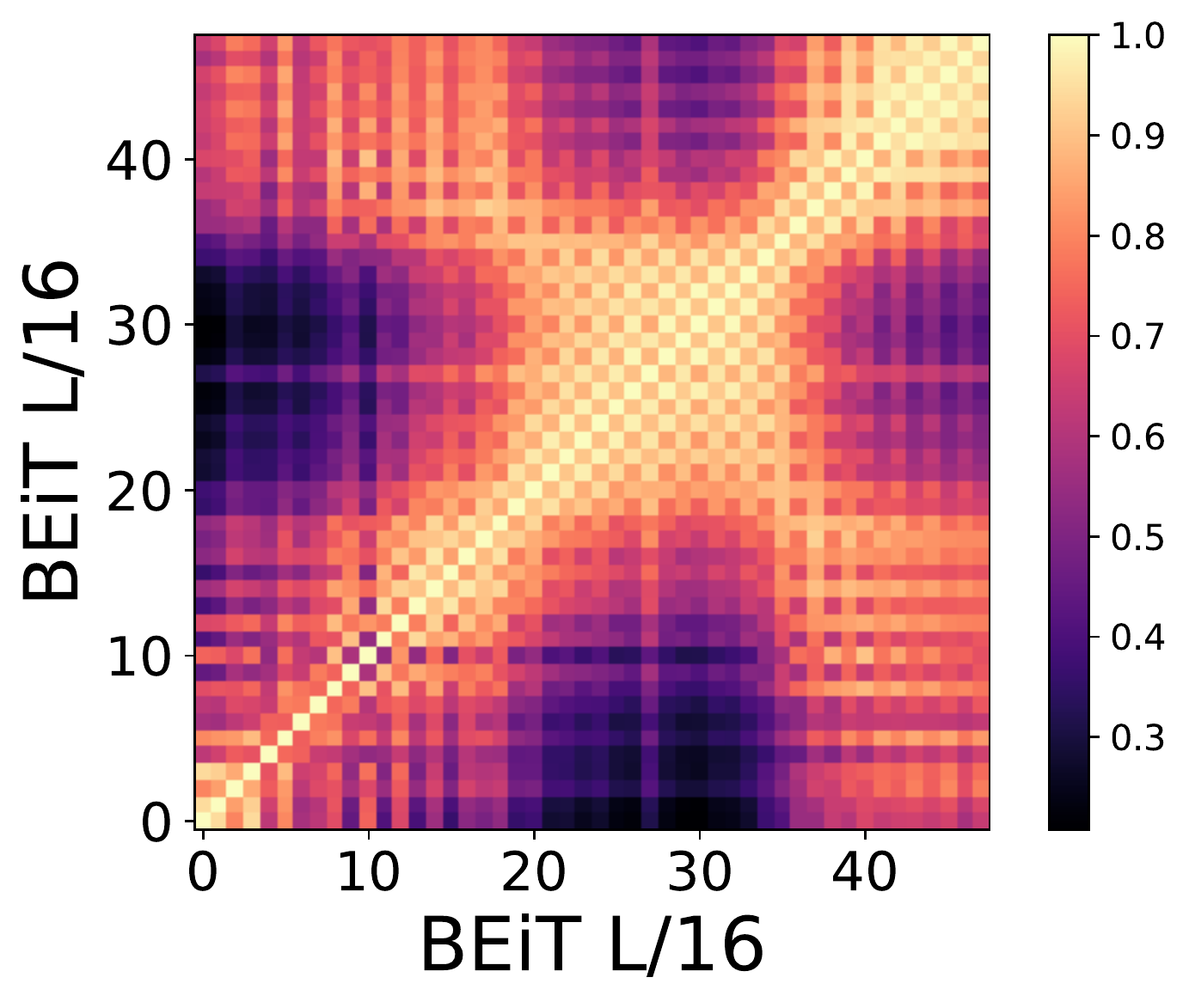}
\end{subfigure}
\caption{CKA across Base and Large BEiT models.}
\label{fig:beit_self}
\end{figure}

\paragraph{X pattern for MAE and BEiT.} As shown in \Cref{fig:mae_self} and \Cref{fig:beit_self}, MAE and BEiT show an X-like pattern in their CKA plot (with the exception of MAE B/16).
This indicates that the late layer features of these models are similar to the early layers but not the middle layers.
We hypothesize that this is due to the reconstruction-oriented nature of the training losses, showing that in their final layers MAE and BEiT are trying to recreate the same local information that is present in the initial layers. 
It should be noted that this X-pattern arises for all sizes of BEiT, but not in MAE B/16.
We attribute this to the fact that MAE has a separate decoder module which is discarded after training, while BEiT does not have a separate decoder. This means that BEiT needs to inherently learn a decoder which leads to emergence of this X pattern at both sizes. For MAE, the fact that the X pattern emerges more clearly for the larger ViT variants suggests that the additional layers of these models start specializing for the task of decoding. This has important implications for the MAE training method, as it suggests that, for larger MAEs, late layers learn to act in a more decoder-like way, which may limit the usefulness of these layers for downstream tasks.

\begin{figure}[t]
    \centering
\begin{subfigure}{0.155\textwidth}
  \includegraphics[width=\textwidth]{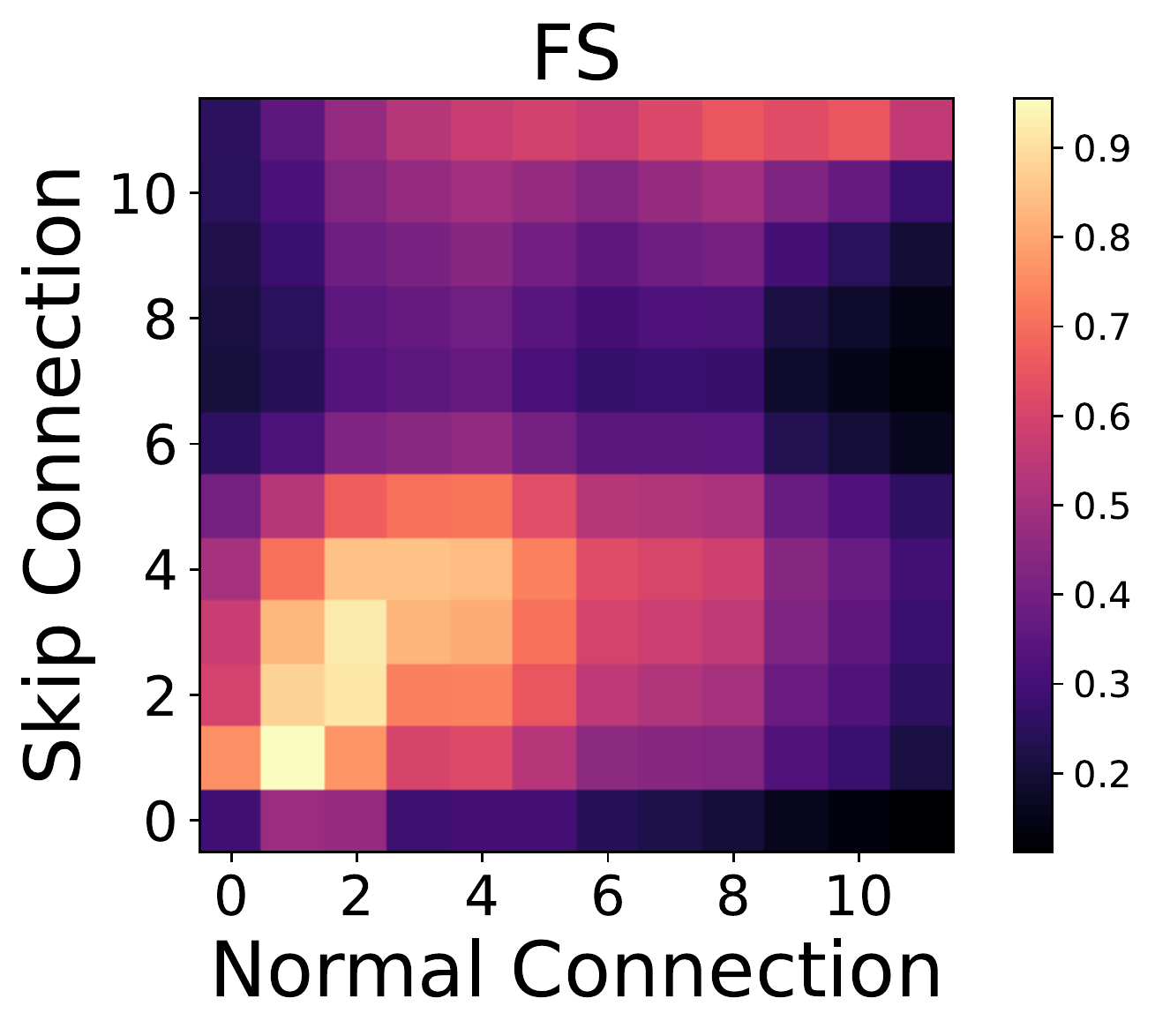}
\end{subfigure}
\begin{subfigure}{0.155\textwidth}
  \includegraphics[width=\textwidth]{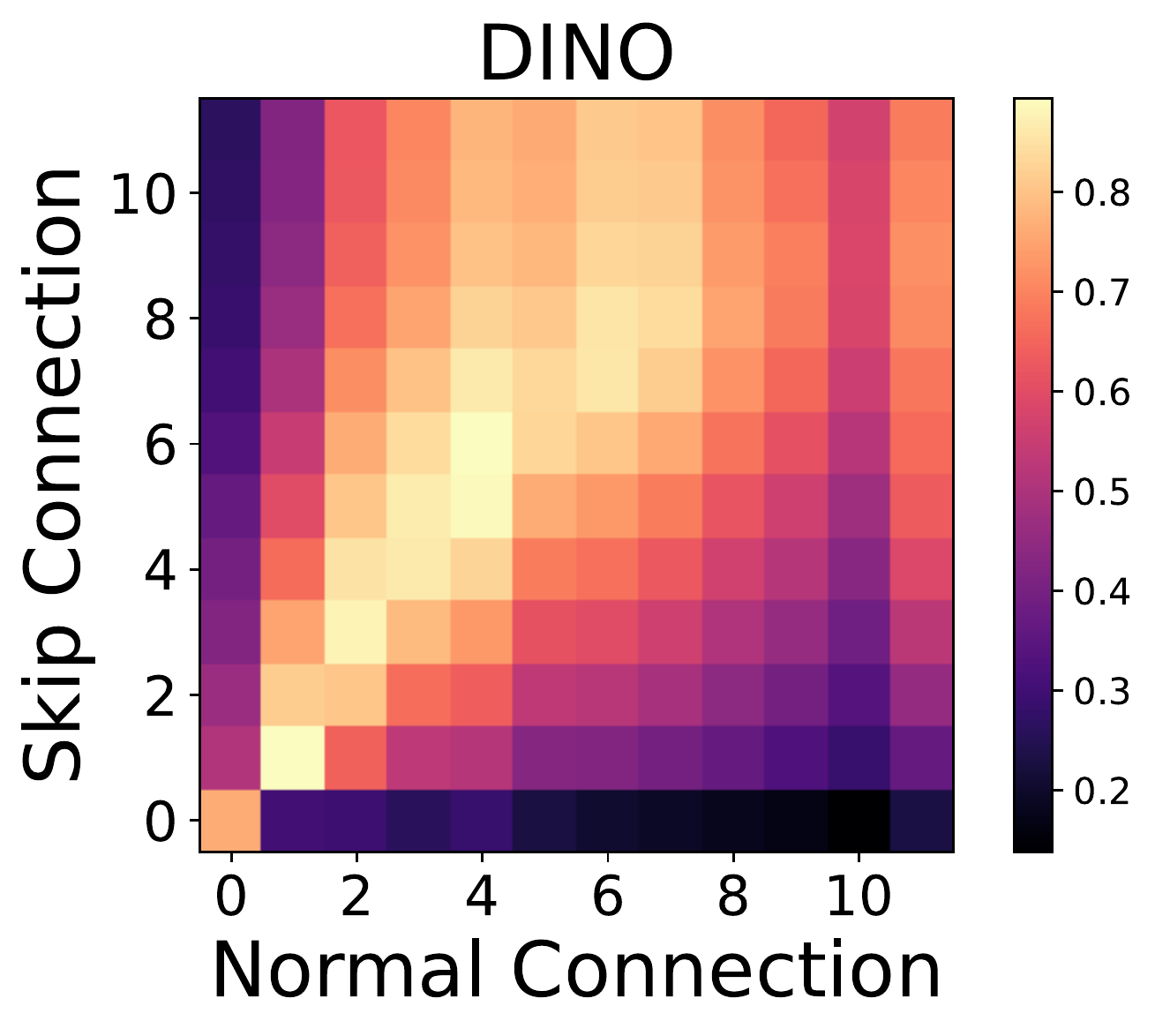}
\end{subfigure}
\begin{subfigure}{0.155\textwidth}
  \includegraphics[width=\textwidth]{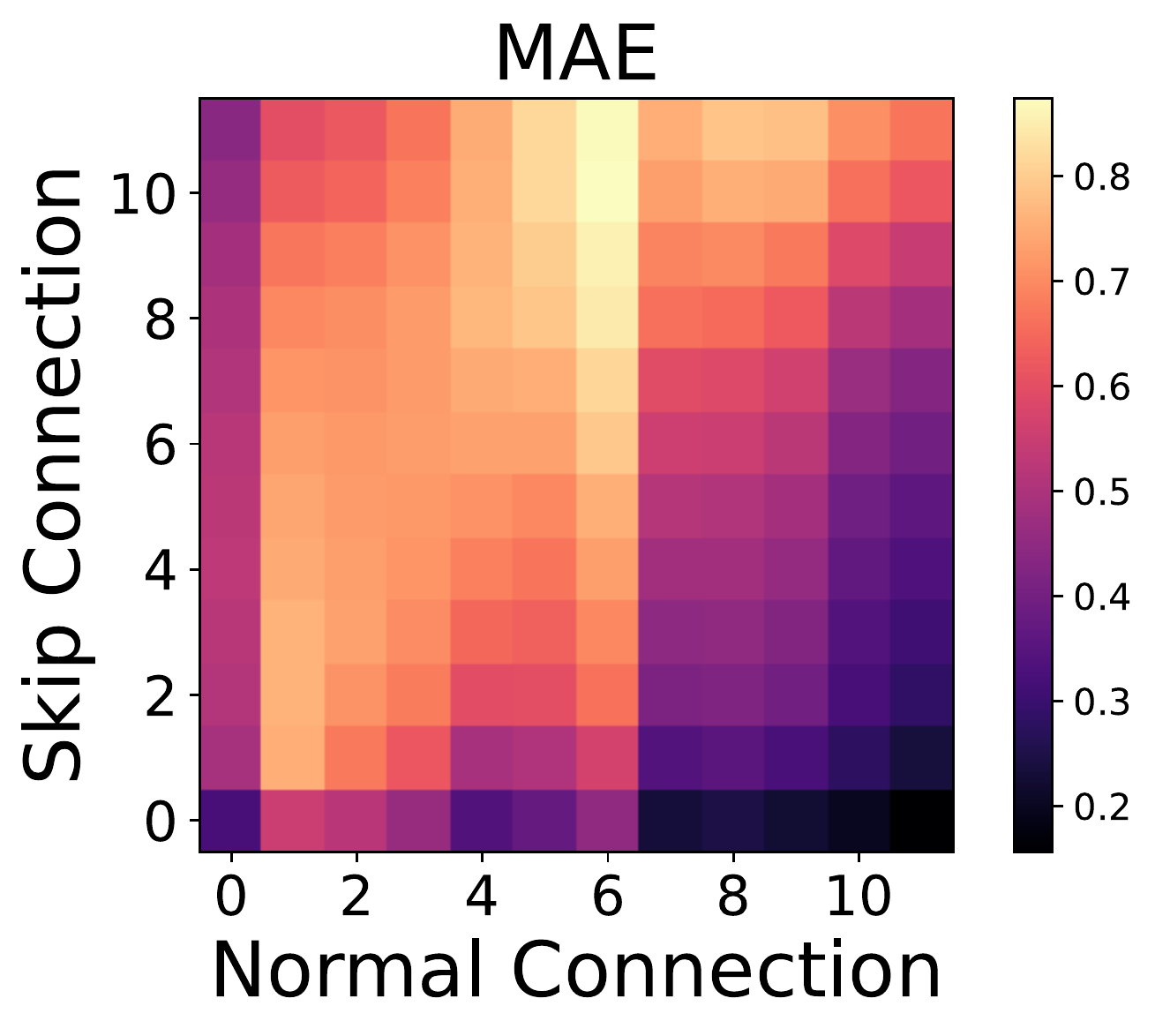}
\end{subfigure}

\medskip
\begin{subfigure}{0.155\textwidth}
  \includegraphics[width=\textwidth]{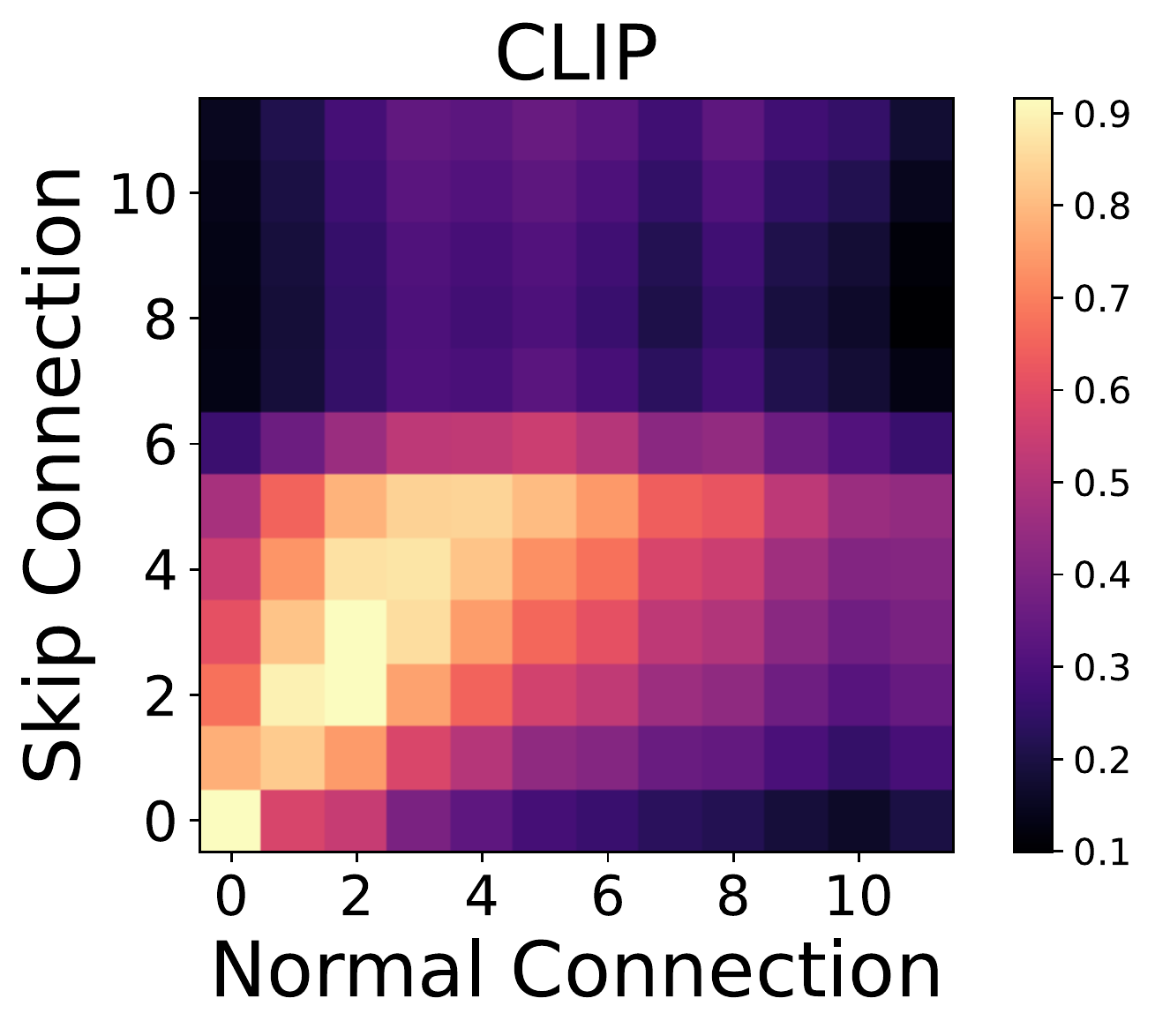}
\end{subfigure}
\begin{subfigure}{0.155\textwidth}
  \includegraphics[width=\textwidth]{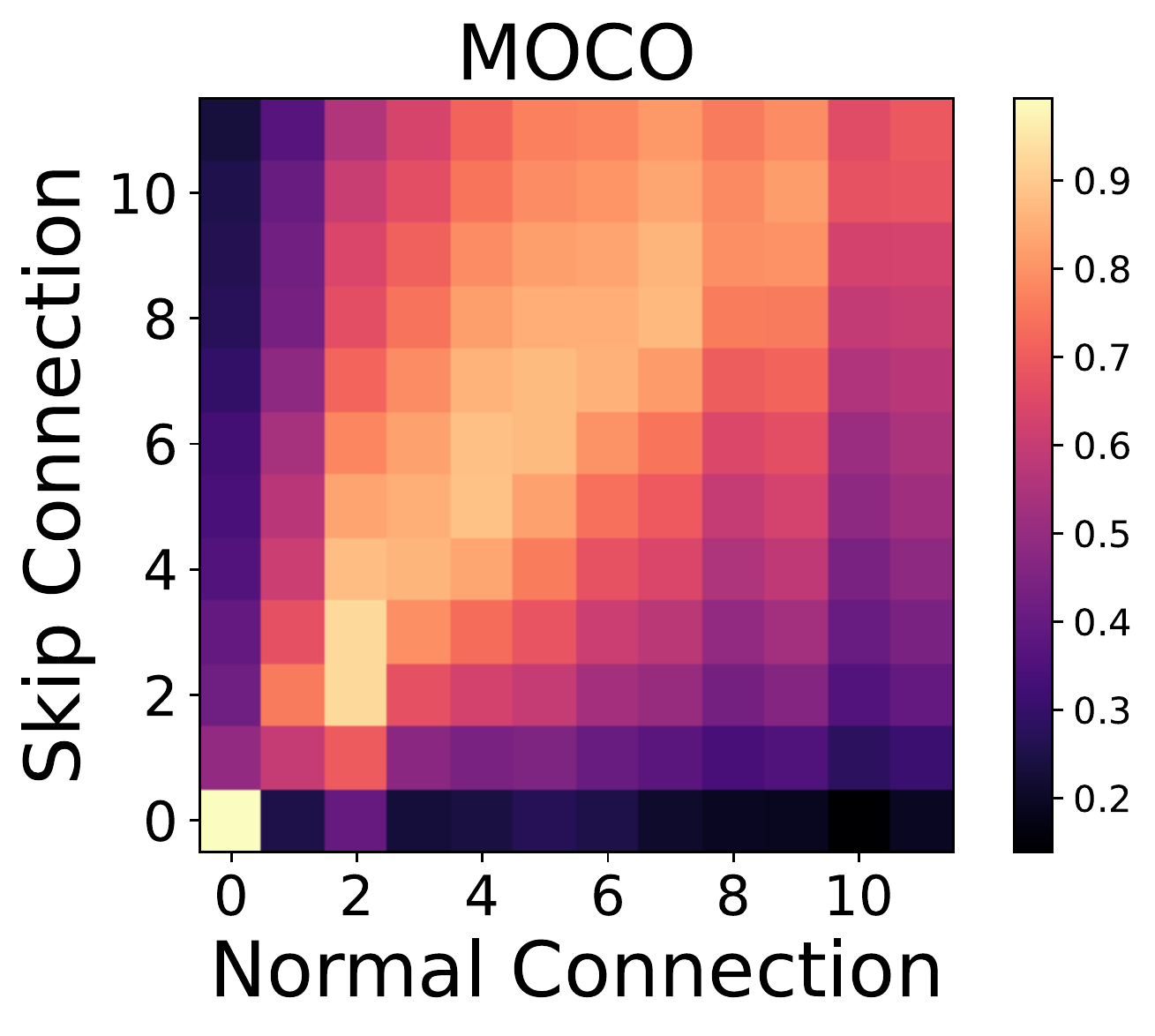}
\end{subfigure}
\begin{subfigure}{0.155\textwidth}
  \includegraphics[width=\textwidth]{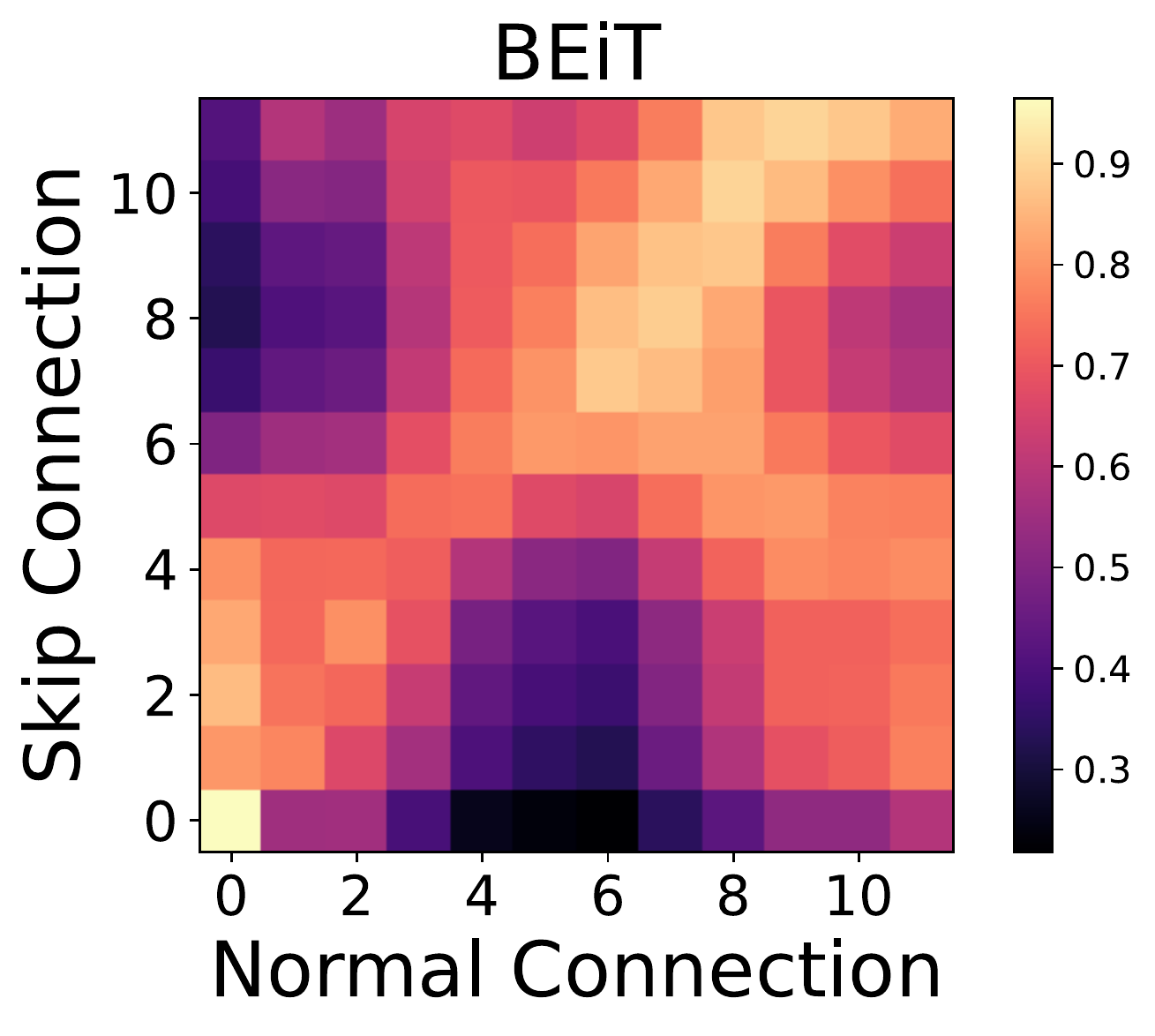}
\end{subfigure}
\caption{\textbf{Residual connection analysis.} We show the CKA similarity between the features coming from the skip-connect (Y-axis) and normal pathway (X-axis) for each MHA layer of each block. Each cell indicates the similarity between the skip connection features and output of normal pathway at that location.}
\label{fig:residual}
\end{figure}

\subsection{Residual Connection Analysis}
\label{sec:residual}

Previously works~\citep{raghu2021vision} have contrasted CNNs and ViTs by comparing the features propagated through the skip connections and normal connections.
We extend this analysis to ViTs trained with varying supervision techniques.
\Cref{fig:residual} shows the CKA between the features coming from the skip connection (Y-axis) and the normal pathway (X-axis) for the MHA layer of each block.
For CLIP and FS we can see a similar trend, initially the skip connection carries similar information as the normal path but after a certain point the information it carries becomes very different (dark regions). This also correlates with the emergence of the Sparse Repeating Patterns observed in \Cref{sec:apdx-att-vis}, providing further evidence that a fundamental shift in information processing behavior occurs in the mid-to-late layers of explicitly supervised ViTs.
For MoCo and DINO, this shift in behavior does not happen and as the depth increases the skip connections and normal pathways still have similar representations. MAE is another special case where, given the reconstruction nature of the loss, at multiple depth locations the normal pathway representation is similar to the skip connection representation. For BEiT, there is again an X-like pattern, likely because it needs to start reconstructing the complete input.
MAE does not show an X-like pattern, despite its similar reconstruction objective. Again we theorize that this occurs because MAE has a separate decoder that is discarded after training.

\subsection{Additional Clustering Analysis}
\label{sec:apdx-feat-clust}

In this section, we expand our feature clustering analysis to the full collection of models. In addition to Cluster Purity, we also report results for Normalized Mutual Information (NMI) and Adjusted Random Index (ARI). We see similar trends for all three clustering metrics.

\begin{figure*}
    \centering
    \begin{subfigure}[b]{\linewidth}
        \centering
        \includegraphics[width=0.76\textwidth]{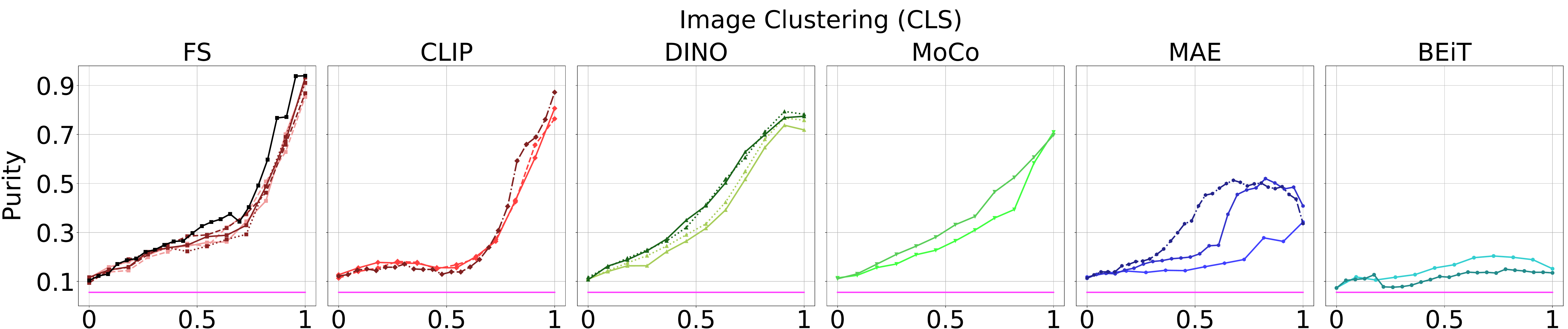}
    \end{subfigure}

    \vspace{0.5em}
    
    \begin{subfigure}[b]{\linewidth}
        \centering
        \includegraphics[width=0.76\textwidth]{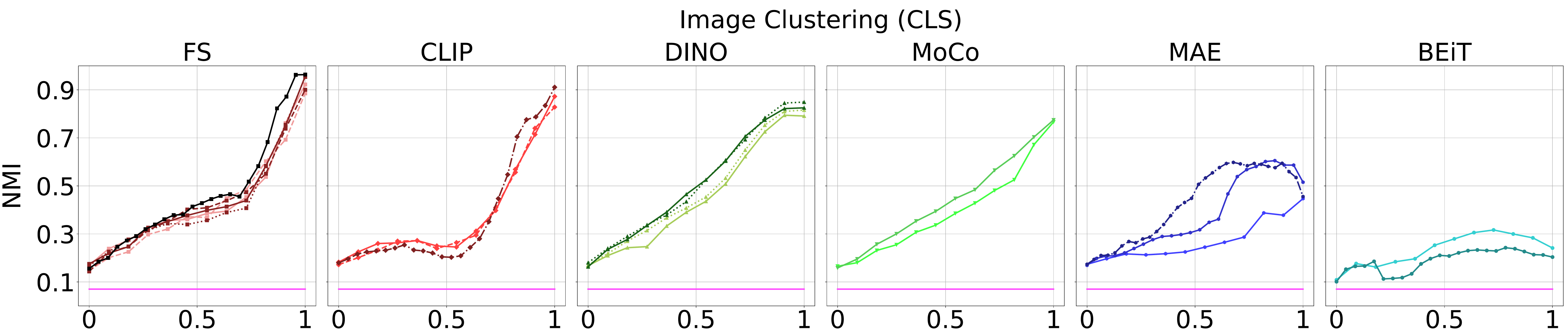}
    \end{subfigure}

    \vspace{0.5em}

    \begin{subfigure}[b]{\linewidth}
        \centering
        \includegraphics[width=0.76\textwidth]{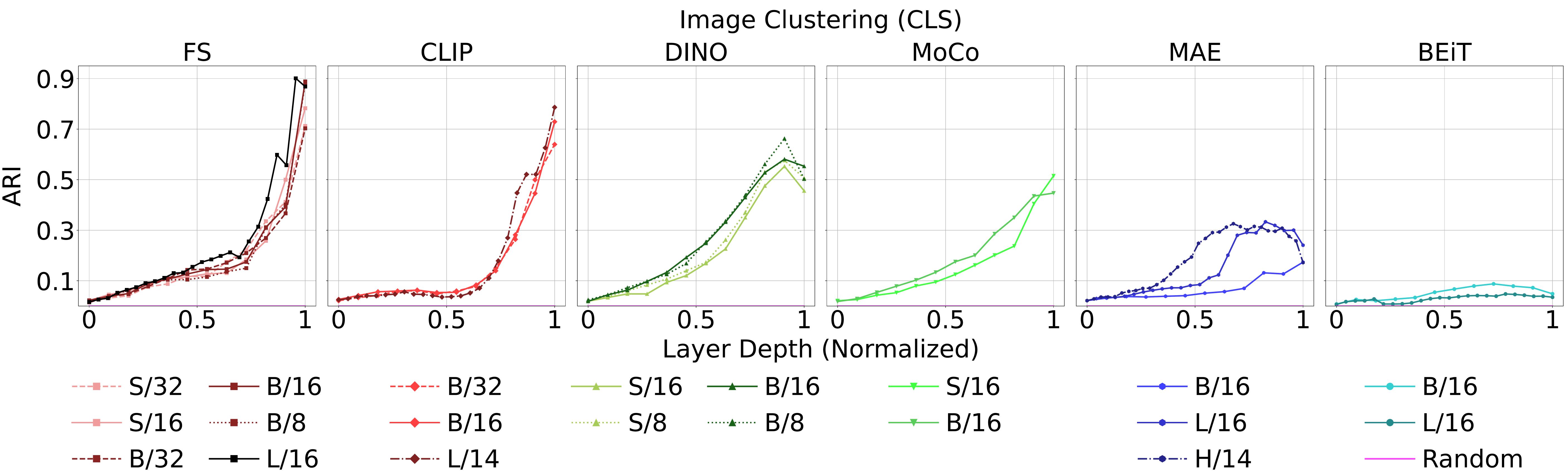}
    \end{subfigure}
    \caption{Expanded CLS feature clustering for image-level labels with ImageNet-50.}
    \label{fig:apdx-cluster-cls}
\end{figure*}

\myparagraph{Image-Level CLS Feature Clustering} Results shown in \Cref{fig:apdx-cluster-cls}. For FS, CLIP, DINO, and MoCo the same general trends hold. Cluster quality rises faster for DINO and MoCo, but FS and CLIP catch up rapidly and overtake at the end. For the deeper model variants (FS L/16 and CLIP L/14) the trends are consistent when plotted against normalized block depth, meaning that semantic information actually emerges half as quickly. For MAE, the larger model variants lead to significantly better cluster purity, which also rises earlier as the models get larger. In contrast, for BEiT cluster purity is generally worse for the larger L/16 model.

\begin{figure*}
    \centering
    \begin{subfigure}[b]{\linewidth}
        \centering
        \includegraphics[width=0.76\textwidth]{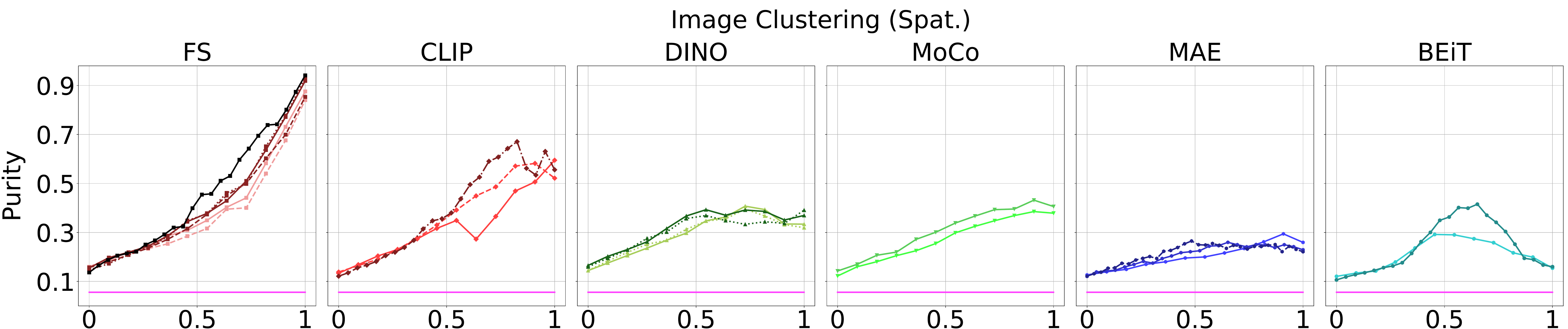}
    \end{subfigure}

    \vspace{0.5em}
    
    \begin{subfigure}[b]{\linewidth}
        \centering
        \includegraphics[width=0.76\textwidth]{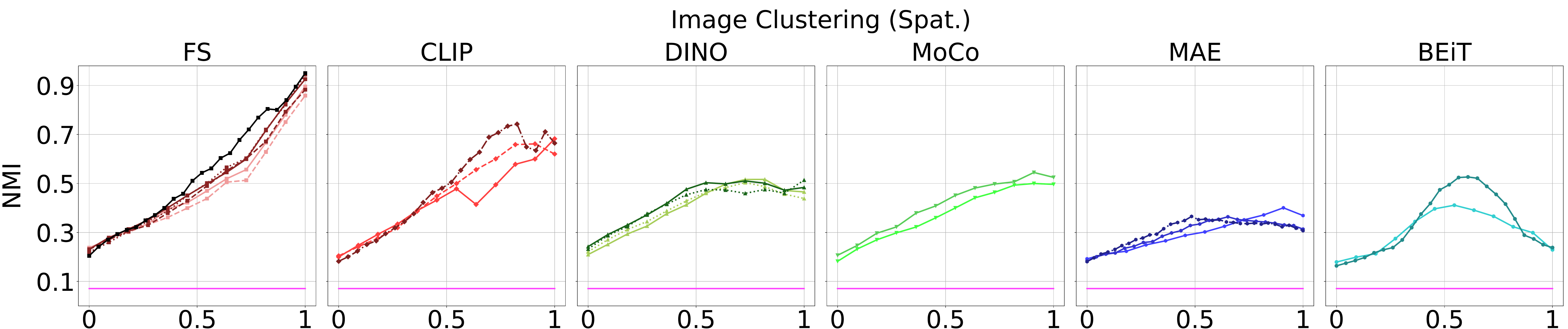}
    \end{subfigure}

    \vspace{0.5em}

    \begin{subfigure}[b]{\linewidth}
        \centering
        \includegraphics[width=0.76\textwidth]{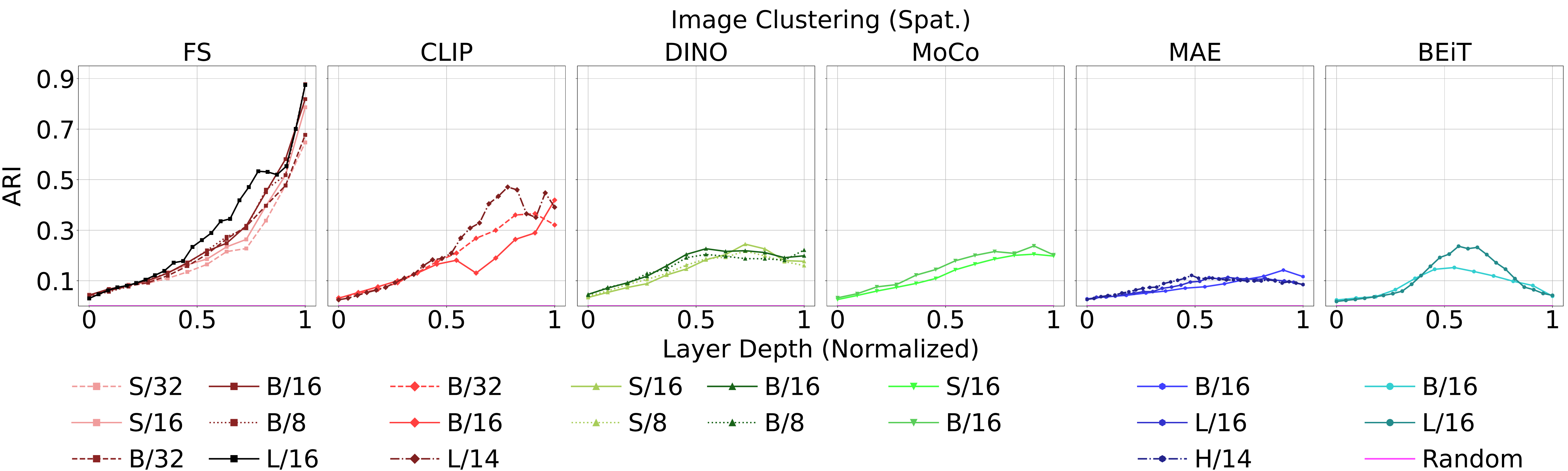}
    \end{subfigure}
    \caption{Averaged spatial feature clustering for image-level labels with ImageNet-50.}
    \label{fig:apdx-cluster-spat}
\end{figure*}

\myparagraph{Image-Level Spatial Feature Clustering} Results shown in \Cref{fig:apdx-cluster-spat}. For FS, CLIP, DINO, and MoCo the same general trends show, though with larger models tending to do slightly better. Interestingly, the spatial features of BEiT L/16 have an increase in cluster purity, which is surprising as its CLS tokens saw a decrease in the previous section.

\begin{figure*}
    \centering
    \begin{subfigure}[b]{\linewidth}
        \centering
        \includegraphics[width=0.76\textwidth]{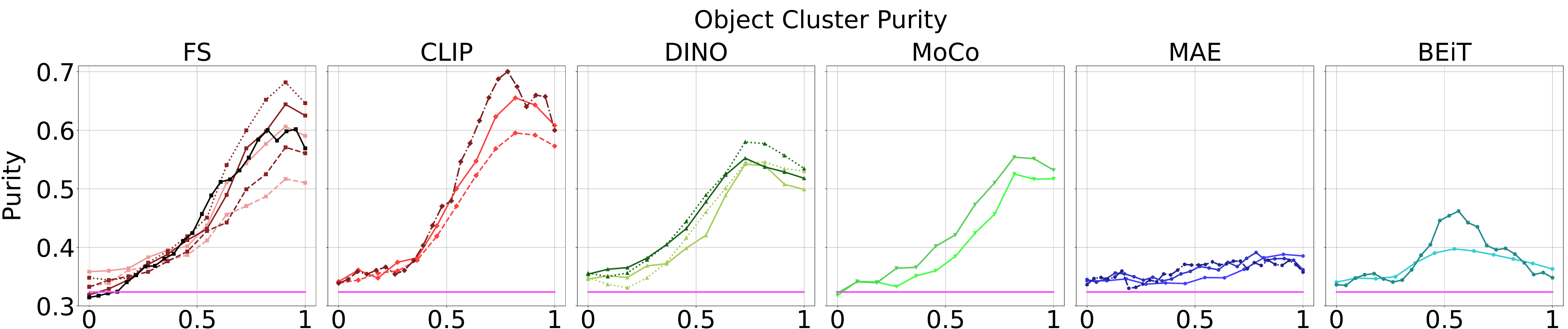}
    \end{subfigure}

    \vspace{0.5em}
    
    \begin{subfigure}[b]{\linewidth}
        \centering
        \includegraphics[width=0.76\textwidth]{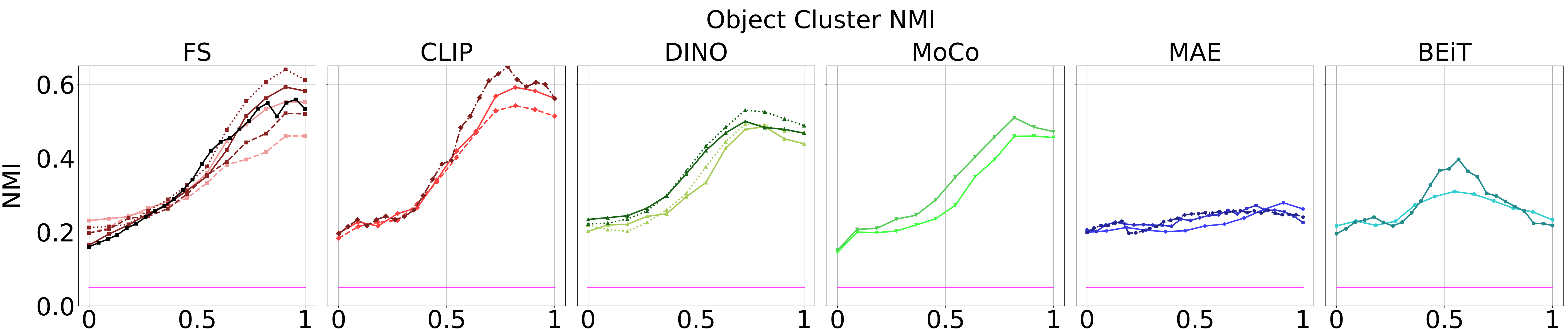}
    \end{subfigure}

    \vspace{0.5em}

    \begin{subfigure}[b]{\linewidth}
        \centering
        \includegraphics[width=0.76\textwidth]{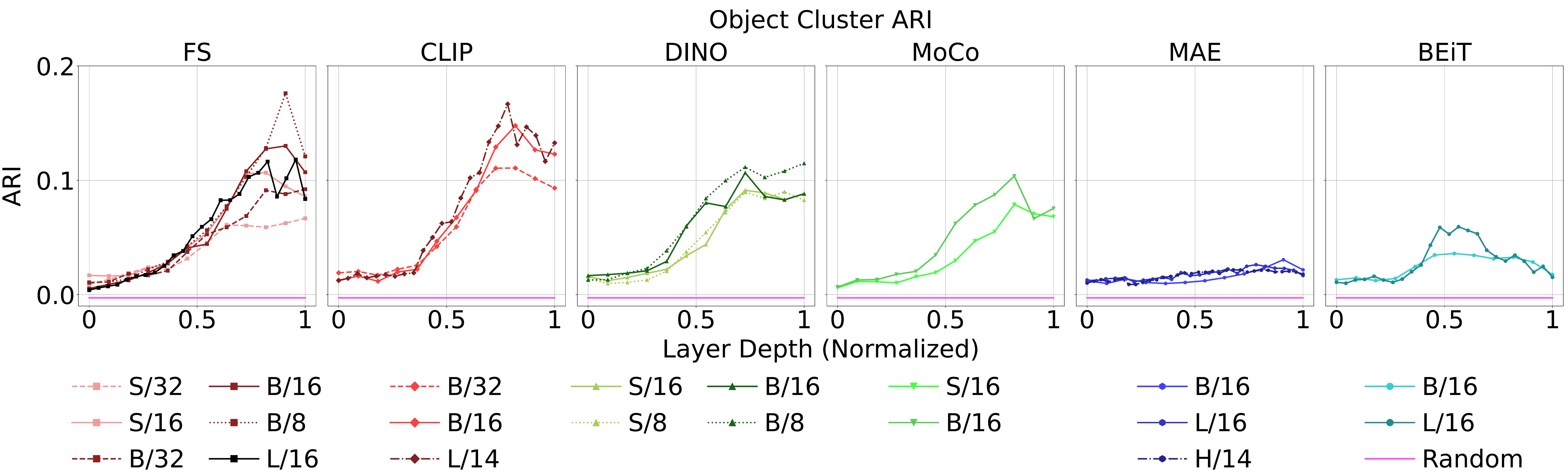}
    \end{subfigure}
    \caption{Expanded spatial feature clustering for object-level labels with COCO.}
    \label{fig:apdx-cluster-objects}
\end{figure*}

\myparagraph{Object-Level Spatial Feature Clustering} Results shown in \Cref{fig:apdx-cluster-objects}. For FS, CLIP, DINO, and MoCo again the general trends hold, with larger models or those with smaller patch size doing slightly better. However, for FS the L/16 variant did worse than B/16. For FS the best scores are achieved by the B/8 variant. Like the previous section, we see a significant boost for BEiT L/16 over B/16.

\begin{figure*}
    \centering
    \begin{subfigure}[b]{\linewidth}
        \centering
        \includegraphics[width=0.76\textwidth]{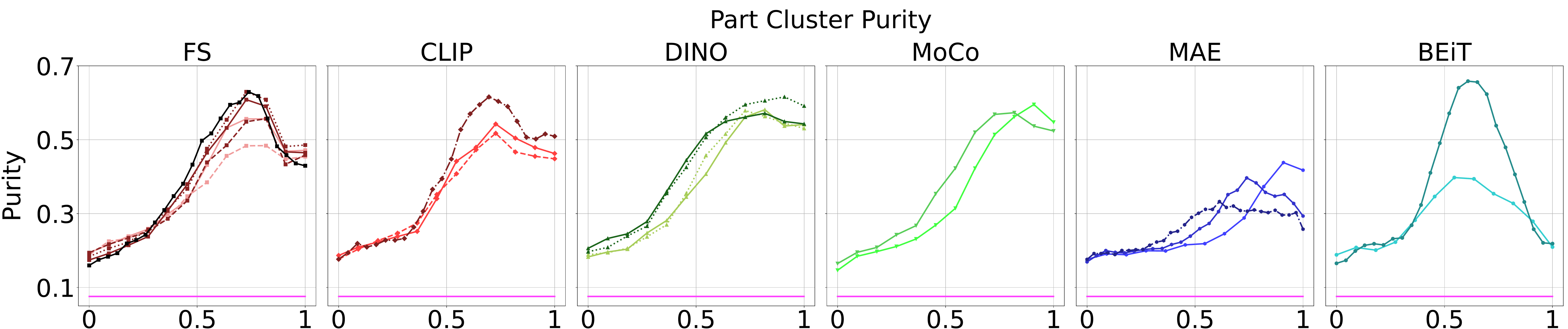}
    \end{subfigure}

    \vspace{0.5em}
    
    \begin{subfigure}[b]{\linewidth}
        \centering
        \includegraphics[width=0.76\textwidth]{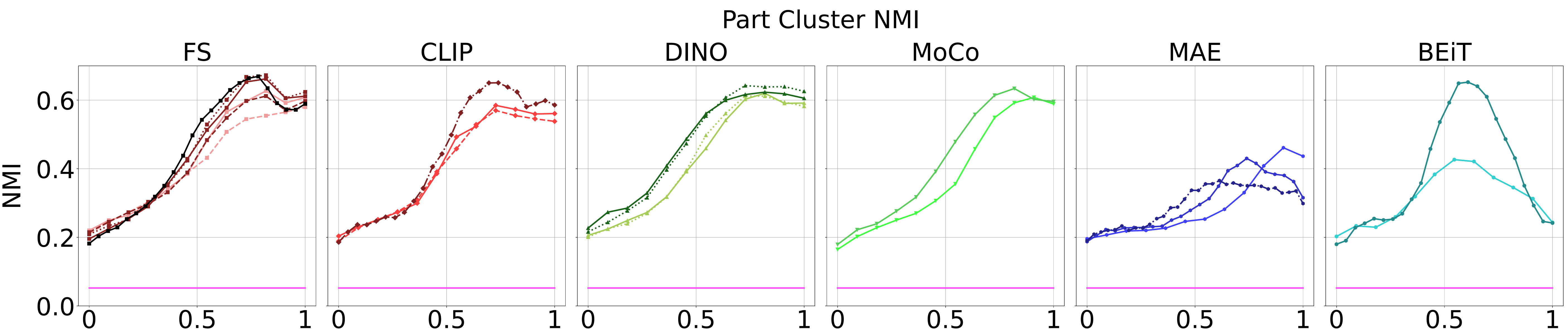}
    \end{subfigure}

    \vspace{0.5em}

    \begin{subfigure}[b]{\linewidth}
        \centering
        \includegraphics[width=0.76\textwidth]{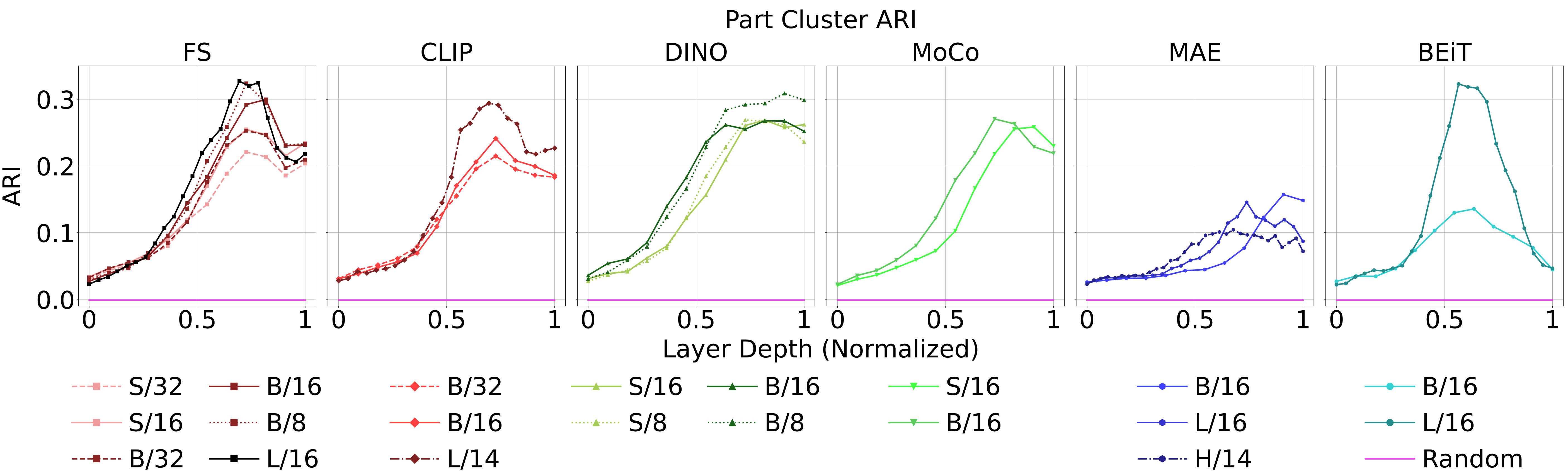}
    \end{subfigure}
    \caption{Expanded spatial feature clustering for part-level labels with PartImageNet.}
    \label{fig:apdx-cluster-parts}
\end{figure*}

\myparagraph{Part-Level Spatial Feature Clustering} Results shown in \Cref{fig:apdx-cluster-parts}. For part-level feature clustering, we previously observed that, for the B/16 models, the self-supervised methods are much more competitive with the explicitly supervised methods. This trend still holds here, with BEiT L/16 performing particularly well, seeing a huge boost over BEiT B/16. For all metrics, BEiT L/16 is on par with the best explicitly and constrastively supervised methods. Larger models and those with smaller patch size generally provide better part-level feature clusters, with the exception of MAE, where the large models actually do worse.


\begin{table*}[t]
 \setlength{\cmidrulewidth}{0.01em}
\renewcommand{\tabcolsep}{6pt}
\renewcommand{\arraystretch}{1.2}
\caption{Best performance for each ViT on each downstream task with the corresponding best layer in parenthesis.}
\centering
\footnotesize
\resizebox{\linewidth}{!}{
\begin{tabular}{@{}lccccccccccccc@{}}
\toprule
 \textbf{Model} & & \multicolumn{10}{c}{\textbf{Task Performance (Best Performing Layer)}} \\
 \cmidrule[\cmidrulewidth](l){3-14}

 Dataset & \multirow{2}{*}{Layers} &     \multicolumn{2}{c}{ImageNet} & \multicolumn{2}{c}{ROxford5k} &\multicolumn{2}{c}{RParis6k} &\multicolumn{3}{c}{DAVIS} &\multicolumn{3}{c}{SPair-71k}\\
 
\cmidrule[\cmidrulewidth](l){3-4}
\cmidrule[\cmidrulewidth](l){5-6} \cmidrule[\cmidrulewidth](l){7-8} \cmidrule[\cmidrulewidth](l){9-11} \cmidrule[\cmidrulewidth](l){12-14}
 
 Metric & & Top-1↑& Top-5↑ & mAP↑ (M) & mAP↑ (H) & mAP↑ (M) & mAP↑ (H) & J Mean↑ & F Mean↑ & J and F Mean↑ & PCK@0.1↑ & PCK@0.05↑ & PCK@0.01↑\\
 \midrule
FS S/32     & 12 & 74.48 (12)  & 90.33 (12)  & 0.33 (12)  & 0.12 (12) & 0.63 (12) & 0.38 (12) & 0.43 (8)  & 0.35 (9)  & 0.39 (9)  & 15.28 (8)  & 4.34 (8)   & 0.13 (8)  \\
FS S/16     & 12 & 80.64 (12) & 93.71 (12) & 0.34 (12) & 0.1 (11)   & 0.66 (12) & 0.40 (12) & 0.57 (8)  & 0.60 (9)  & 0.58 (8)  & 26.37 (8)  & 11.62 (8)  & 0.68 (9)  \\
FS B/32     & 12 & 79.08 (12) & 92.8 (12)   & 0.33 (12) & 0.12 (12) & 0.67 (12) & 0.43 (12) & 0.42 (9)  & 0.34 (9)  & 0.38 (9)  & 15.49 (9)  & 4.19 (8)   & 0.15 (7)  \\
FS B/16     & 12 & 83.79 (12) & 95.01 (12) & 0.45 (12) & 0.19 (12) & 0.72 (12) & 0.51 (12)  & 0.58 (8)  & 0.60 (8)  & 0.59 (8)  & 28.56 (9)  & 12.33 (8)  & 0.63 (7)  \\
FS B/8      & 12 & 85.58 (12) & 95.71 (12) & 0.45 (12) & 0.16 (12) & 0.73 (12)  & 0.51 (12) & 0.66 (7)  & 0.70 (9)  & 0.68 (9)  & 36.09 (9)  & 21.97 (9)  & 1.61 (9)   \\
FS L/16     & 24 & 85.03 (24) & 95.39 (24) & 0.42 (24) & 0.15 (24) & 0.73 (24)  & 0.51 (24) & 0.56 (13) & 0.57 (13) & 0.56 (13) & 30.99 (17) & 13.49 (15) & 0.79 (14) \\
CLIP B/32   & 12 & 70.90 (12) & 89.54 (12) & 0.38 (12) & 0.10 (12) & 0.66 (12) & 0.41 (12) & 0.44 (9)  & 0.37 (9)  & 0.41 (9)  & 18.55 (8)  & 5.37 (8)   & 0.26 (8)  \\
CLIP B/16   & 12 & 75.75 (12) & 92.27 (12) & 0.40 (12) & 0.11 (12) & 0.71 (12) & 0.48 (12) & 0.58 (9)  & 0.62 (9)  & 0.60 (9)  & 30.70 (8)  & 13.61 (8)  & 0.98 (6)   \\
CLIP L/14   & 24 & 80.24 (24) & 94.15 (24)  & 0.45 (24) & 0.17 (24) & 0.70 (24) & 0.49 (24) & 0.57 (14) & 0.62 (17)  & 0.60 (17) & 36.04 (15) & 16.72 (15) & 1.15 (13) \\
DINO S/16   & 12 & 74.61 (12) & 90.08 (12) & 0.38 (12) & 0.14 (12) & 0.61 (12) & 0.33 (12) & 0.60 (11) & 0.63 (11) & 0.61 (11) & 26.72 (9)  & 11.68 (9)  & 0.55 (9)  \\
DINO S/8    & 12 & 74.34 (12)  & 90.15 (12)  & 0.36 (12) & 0.12 (12) & 0.59 (12) & 0.30 (12) & 0.70 (12) & 0.77 (12) & 0.73 (12) & 31.14 (8)  & 18.15 (8)  & 1.48 (8)  \\
DINO B/16   & 12 & 76.06 (12) & 91.40 (12) & 0.37 (12) & 0.11 (12) & 0.62 (12) & 0.35 (12) & 0.59 (12) & 0.61 (12) & 0.60 (12) & 28.28 (9)   & 12.00 (7)  & 0.65 (6)  \\
DINO B/8    & 12 & 77.70 (12) & 92.24 (12) & 0.40 (12) & 0.13 (11) & 0.65 (12) & 0.37 (12) & 0.69 (10) & 0.77 (10) & 0.73 (10)  & 33.17 (8)  & 19.04 (8)  & 1.66 (5)  \\
MoCo S/16   & 12 & 68.71 (12)  & 86.36 (12) & 0.27 (12) & 0.07 (12) & 0.50 (12) & 0.22 (12) & 0.58 (10) & 0.62 (10) & 0.6 (10)   & 24.88 (9)  & 10.92 (9)  & 0.39 (11) \\
MoCo B/16   & 12 & 71.59 (12) & 88.37 (12) & 0.31 (12) & 0.08 (12) & 0.51 (12) & 0.22 (12) & 0.59 (11) & 0.62 (11) & 0.61 (11) & 25.85 (9)  & 10.64 (8)   & 0.43 (10) \\
MAE B/16    & 12 & 45.19 (12) & 65.32 (12) & 0.15 (10) & 0.02 (10)  & 0.28 (10) & 0.08 (10) & 0.54 (11) & 0.54 (12) & 0.54 (12) & 22.65 (11) & 10.59 (11) & 0.44 (11) \\
MAE L/16    & 24 & 60.80 (20) & 78.9 (20)   & 0.19 (21) & 0.03 (21) & 0.35 (21) & 0.11 (21) & 0.55 (23) & 0.56 (23) & 0.55 (23) & 27.65 (19) & 13.02 (22) & 0.60 (21) \\
MAE H/14    & 32 & 63.16 (23) & 79.87 (23) & 0.20 (23) & 0.03 (30)  & 0.39 (23) & 0.13 (23) & 0.58 (31) & 0.61 (30)  & 0.59 (30) & 27.50 (26) & 13.65 (26) & 1.35 (26)  \\
BEiT B/16   & 12 & 26.84 (8)  & 45.12 (8)  & 0.14 (8)  & 0.02 (8)  & 0.20 (8)  & 0.05 (10) & 0.57 (10) & 0.59 (9)  & 0.58 (9)   & 24.11 (8)  & 11.02 (8)   & 0.54 (7)  \\
BEiT L/16   & 24 & 41.24 (18) & 62.79 (18) & 0.16 (18) & 0.02 (18) & 0.25 (17) & 0.06 (17) & 0.58 (17) & 0.64 (15) & 0.61 (15)  & 37.52 (15)  & 18.22 (16) & 1.04 (16) \\
Random & - & 0.10       & 0.49       & 0.02      & 0.01      & 0.04       & 0.03      & 0.03  & 0.08      & 0.06      & 1.32       & 0.34       & 0.02\\

\bottomrule
\end{tabular}
 }
\label{tab:best_performance_all}
\vspace{-1em}
\end{table*}

\section{Downstream Task Analysis}

\subsection{Keypoint Correspondence Additional Details}

As part of our Downstream Task Analysis, we present results for Keypoint Correspondence as an additional local-focused task.
Given an input image with a set of human annotated keypoints, a model must predict the position of corresponding keypoints in a second paired image with the same type of object. Challenges in this task include changes in scale, size, and large intraclass variations. 
Correspondence is a prerequisite step in applications such as pose estimation~\citep{schonberger2016structure}, 3D reconstruction~\citep{schonberger2016structure}, and edit propagation in images and videos~\citep{xu2009efficient}.
We use the SPair-71k~\citep{min2019spair} dataset consisting of 1800 images from 18 categories. Following the evaluation protocol used by \citet{amir2021deep}, we randomly sample 20 image pairs from each category of the test set and compute PCK~\citep{yang2012articulated} (percentage of correct keypoints) for each of the 18 categories. Given the dense ViT spatial token features of a source image, a target image, and source keypoint, we (1) get the corresponding feature vector of the keypoint in the source image, (2) find the nearest neighbor of this feature vector in the target image, and (3) get the 2D location of the nearest neighbor in the target image. The keypoint prediction is considered correct if it is within, a threshold $\alpha \cdot \max(H, W)$ of the groundtruth correspondence, where $\alpha$ is a constant and $(H, W)$ are the height and width of the target image. We report the average PCK@$0.1$, PCK@$0.05$, and PCK@$0.01$.

\subsection{Results for ViT Variants}
\label{sec:apdx-downstream}

\begin{figure*}
    \centering
    \begin{subfigure}[b]{\linewidth}
        \centering
        \includegraphics[width=0.76\textwidth]{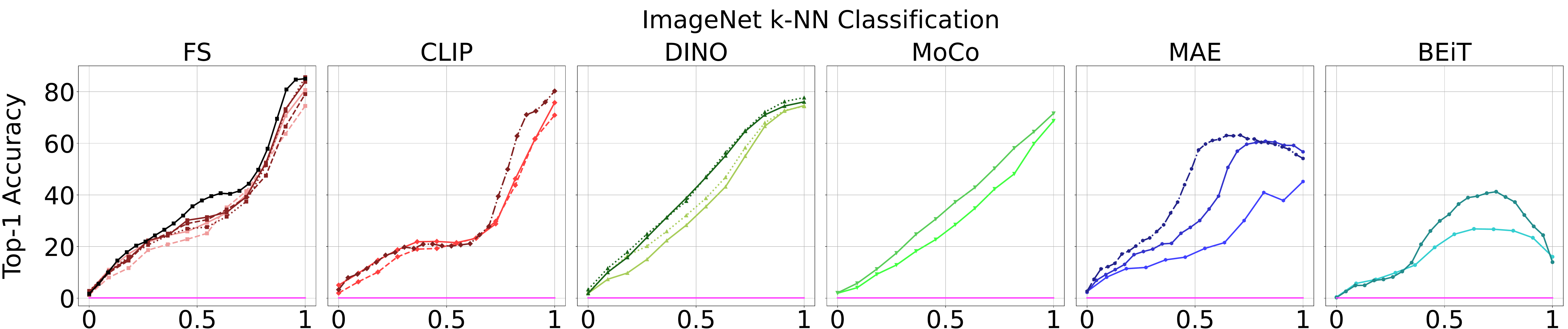}
    \end{subfigure}

    \vspace{0.5em}
    
    \begin{subfigure}[b]{\linewidth}
        \centering
        \includegraphics[width=0.76\textwidth]{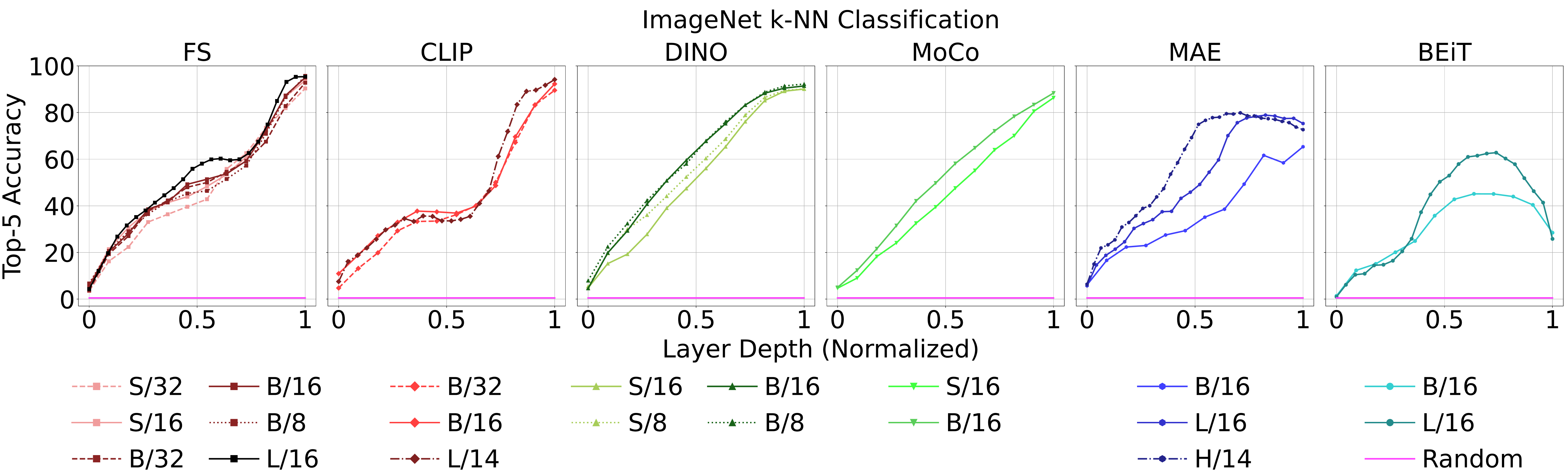}
    \end{subfigure}
    \caption{k-NN ImageNet classification results for all ViT variants.}
    \label{fig:apdx-knn-all}
\end{figure*}

\myparagraph{ImageNet Classification}
As seen in \Cref{fig:apdx-knn-all}, the general trends as reported in the main paper for this task hold for each training method. The FS B/8 has the highest Top-1 and Top-5 performance. For FS, CLIP, DINO and MoCo the trends show that performance improves as we go to later layers. It is also interesting to see that under normalized depth, the larger MAE models peak earlier than the smaller ones and show a higher peak performance. For BEiT, the peak performance occurs at a similar relative depth but is higher for the L/16 model.

\begin{figure*}
    \centering
    \begin{subfigure}[b]{\linewidth}
        \centering
        \includegraphics[width=0.76\textwidth]{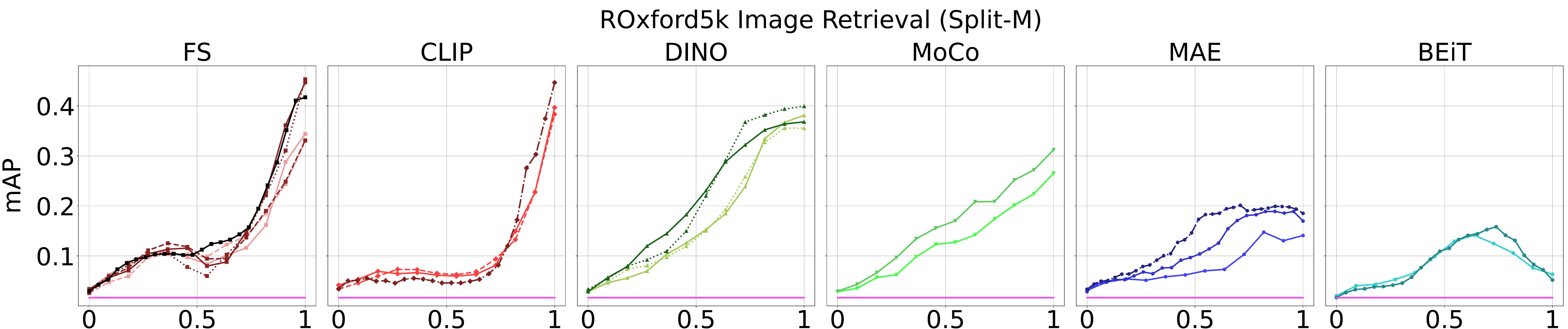}
    \end{subfigure}

    \vspace{0.5em}
    
    \begin{subfigure}[b]{\linewidth}
        \centering
        \includegraphics[width=0.76\textwidth]{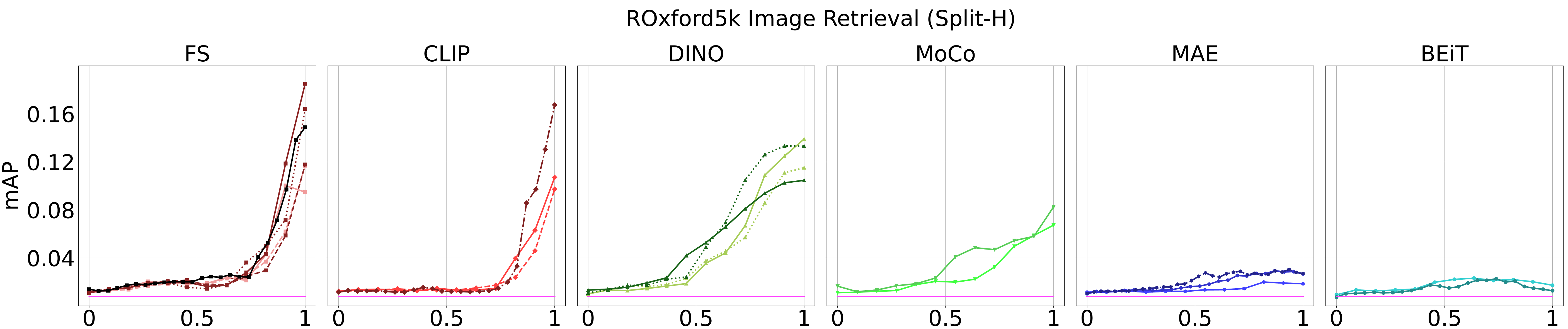}
    \end{subfigure}

    \vspace{0.5em}

    \begin{subfigure}[b]{\linewidth}
        \centering
        \includegraphics[width=0.76\textwidth]{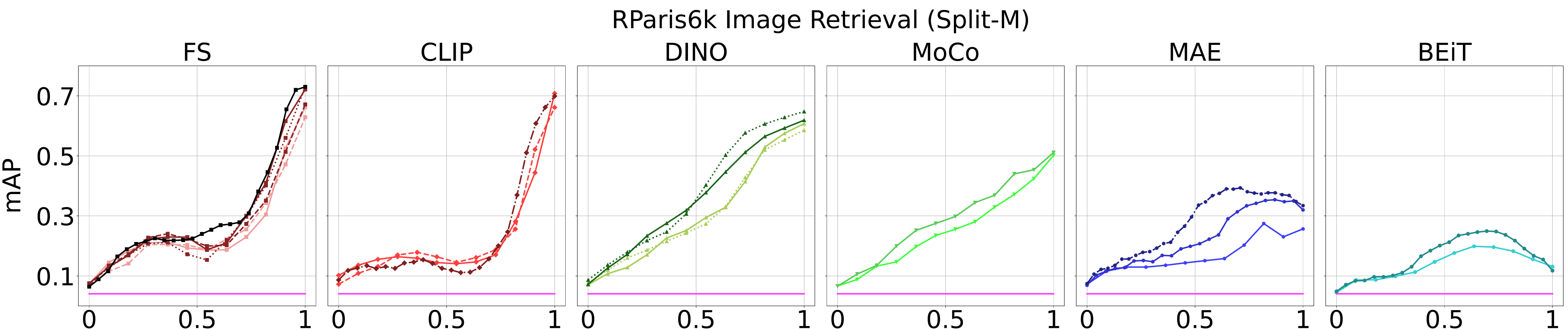}
    \end{subfigure}
    
    \vspace{0.5em}

    \begin{subfigure}[b]{\linewidth}
        \centering
        \includegraphics[width=0.76\textwidth]{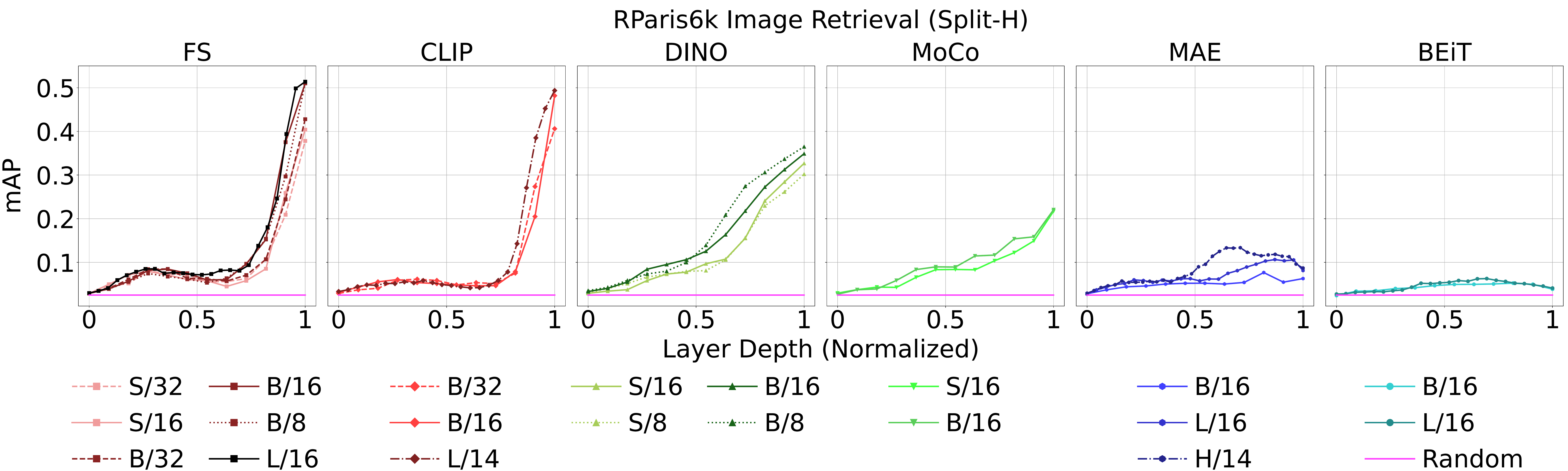}
    \end{subfigure}
    
    \caption{ROxford5k and RParis6k retrieval results for all ViT variants.}
    \label{fig:apdx-retrieval-all}
\end{figure*}

\myparagraph{Image Retrieval}
As shown in \Cref{fig:apdx-retrieval-all}, the Base and Large models for FS and CLIP perform well for this image-level task. This aligns with our observations in the main paper. The FS B/16 performs the best on ROxford5k while the FS B/8 and L/16 are the best performers on RParis6k. For the FS, CLIP, MoCo and DINO models, performance improves as we go to later layers in most cases. MAE and BEiT again peak early in the mid-to-late layers. The general observations and trends for this task are similar to the k-NN task as they are both image-level global tasks. The only difference being that, for this task, the later layers of FS and CLIP show a sudden improvement while the earlier layers are flatter when compared to the trends for k-NN.

\begin{figure*}
    \centering
    \begin{subfigure}[b]{\linewidth}
        \centering
        \includegraphics[width=0.76\textwidth]{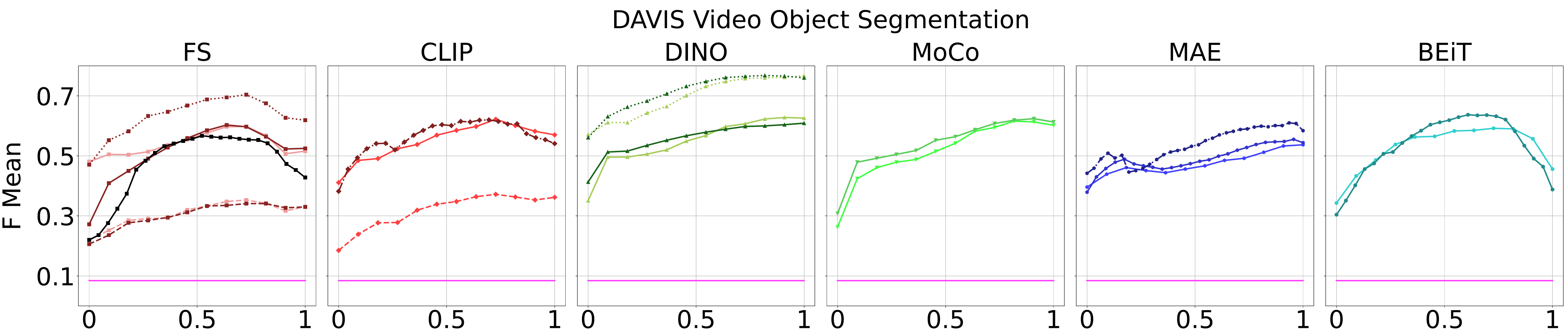}
    \end{subfigure}

    \vspace{0.5em}
    
    \begin{subfigure}[b]{\linewidth}
        \centering
        \includegraphics[width=0.76\textwidth]{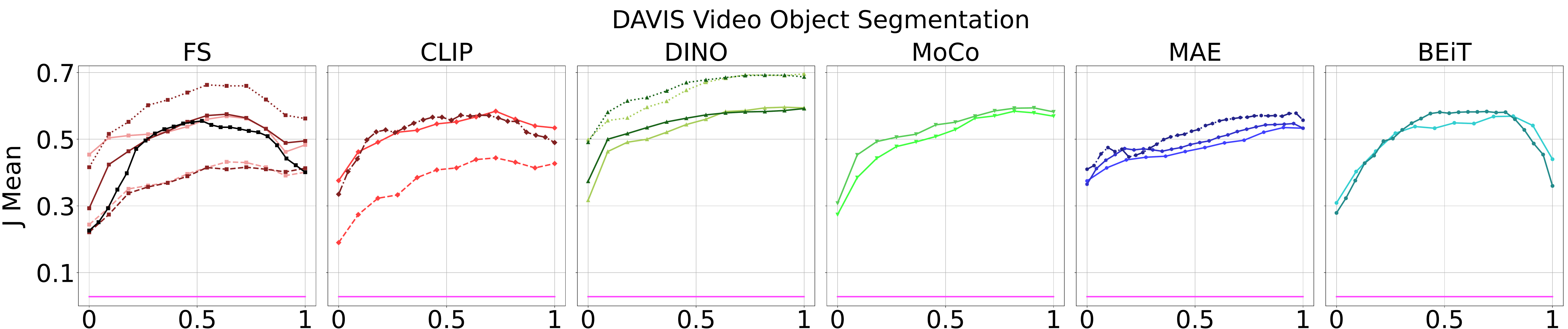}
    \end{subfigure}

    \vspace{0.5em}

    \begin{subfigure}[b]{\linewidth}
        \centering
        \includegraphics[width=0.76\textwidth]{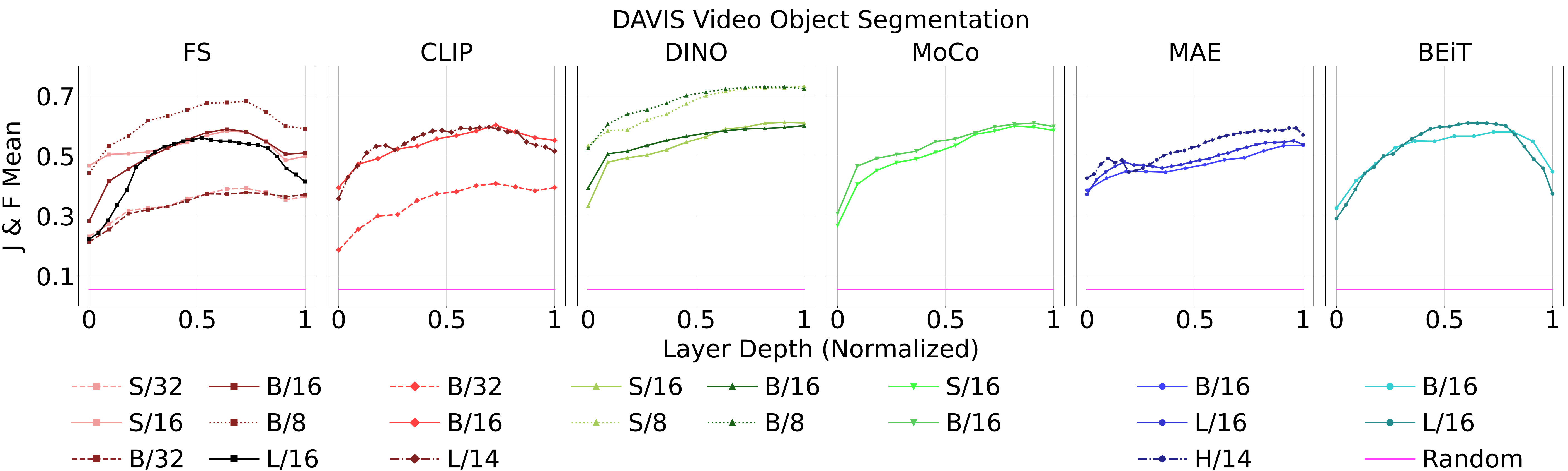}
    \end{subfigure}
    \caption{DAVIS Video Segmentation Propagation comparison for all ViTs}
    \label{fig:apdx-davis-all}
\end{figure*}

\myparagraph{DAVIS Segmentation Propagation}
As shown in \Cref{fig:apdx-davis-all} many models peak at an earlier layer.
The best performance comes from DINO B/8 and S/8, followed closely by FS B/8. Meanwhile, the models with patch size 32 see a significant drop in performance.
Given the dense-prediction-based nature of this task, these methods which are trained with a smaller patch size have finer features which give them a boost in performance. The reconstruction-based models, BEiT and MAE, are also very competitive in this task, performing on par with FS, CLIP, DINO, and MoCo among the models with patch size 16.

\begin{figure*}
    \centering
    \begin{subfigure}[b]{\linewidth}
        \centering
        \includegraphics[width=0.76\textwidth]{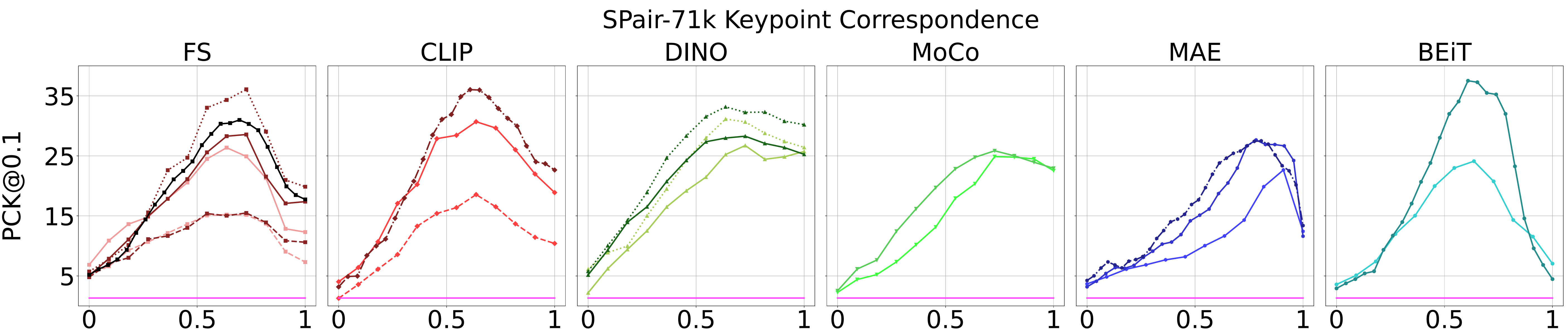}
    \end{subfigure}

    \vspace{0.5em}
    
    \begin{subfigure}[b]{\linewidth}
        \centering
        \includegraphics[width=0.76\textwidth]{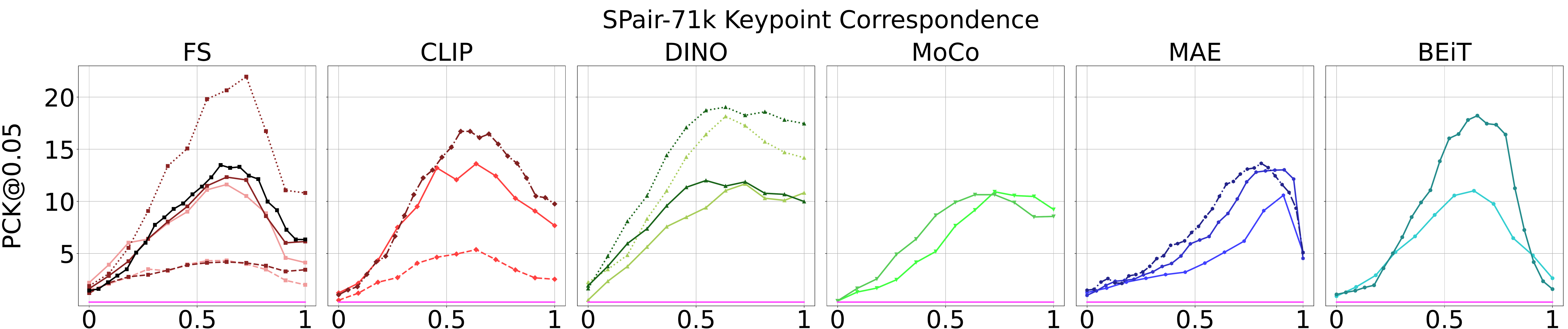}
    \end{subfigure}

    \vspace{0.5em}

    \begin{subfigure}[b]{\linewidth}
        \centering
        \includegraphics[width=0.76\textwidth]{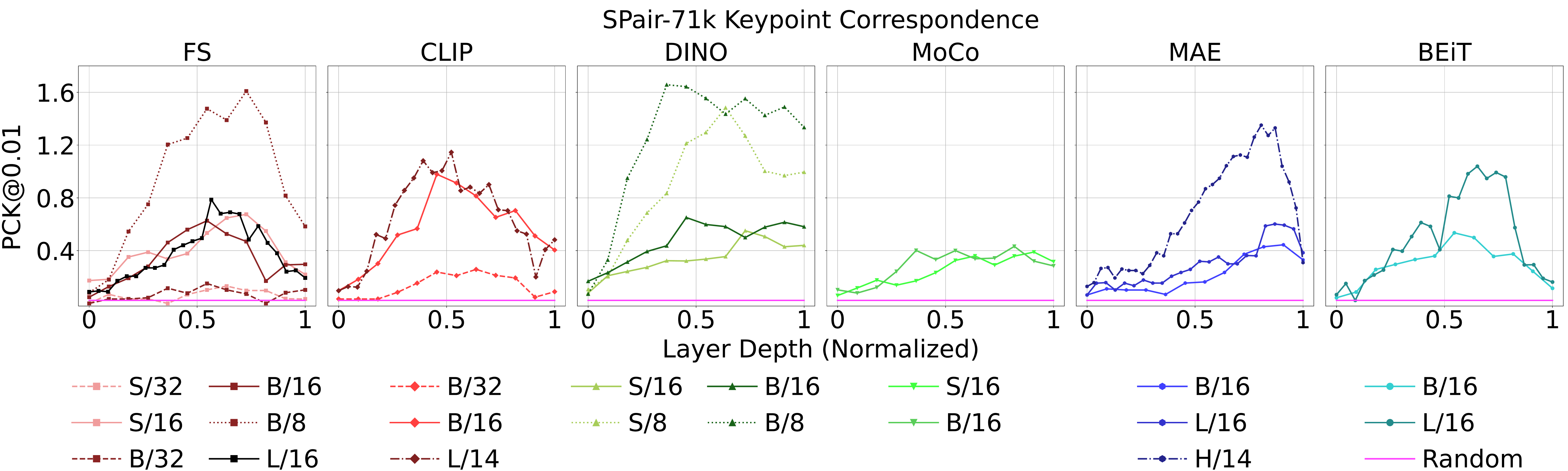}
    \end{subfigure}
    \caption{SPair-71k Keypoint Correspondence comparison for all ViTs}
    \label{fig:apdx-spair-all}
\end{figure*}

\myparagraph{Keypoint Correspondence}
We show comparisons on this task in \Cref{fig:apdx-spair-all}. It should be highlighted that BEiT L/16 performs the best for PCK@0.1, again showing a massive improvement over BEiT B/16. This large boost correlates with the its improved part-level feature purity observed in \Cref{sec:apdx-feat-clust}. 
In general the smaller patch size models like DINO B/8, DINO S/8, CLIP L/14 and FS B/8 are also good performers. Due to their finer feature grids, these methods perform much better at the stricter thresholds (PCK@$0.05$ and PCK@$0.01$). All models on this task peak around mid-to-late layers.

\newpage
\subsection{Summary of Downstream Tasks}
We report the best result for each model along with the layer at which it occurs in \Cref{tab:best_performance_all}. This table captures all downstream tasks and summarizes all metrics for each task. We would like to highlight that \textbf{[1]} different models peak at different layers, based on type of task, local vs. global, and \textbf{[2]} no one model is the best model for all tasks.


\subsection{ImageNet-1k Classification with Linear Probes}
We present additional results for ImageNet classification with Linear Probes in Table \ref{tab:linprob}. For each model, we trained a linear layer on the last layer features for 20 epochs on the ImageNet-1k training set, and report results on the validation set. For BEiT we instead use layer 8, which gave the best k-NN classification results.
This analysis includes variations based on protocols from the compared works. We present results for CLS token features in row 2 and average-pooled spatial token features (proposed by BEiT) in row 3. We also test the addition of a batch normalization layer before the linear layer (proposed by MAE) in rows 4 \& 5.
As FS is trained with ImageNet labels, it performs best in all settings. For approaches with explicit CLS supervision (FS, CLIP, DINO, MoCo), the CLS token features give higher accuracy. For MAE and BEiT, due to the local nature of their supervision, their spatial features give better performance. Batchnorm is generally beneficial for all models and features.

\vspace{-0.2em}
\begin{table}[h!]
 \setlength{\cmidrulewidth}{0.01em}
\renewcommand{\tabcolsep}{6pt}
\renewcommand{\arraystretch}{1.1}
\caption{Accuracy@1 of ViT-B/16 models for Linear Probing on ImageNet-1k val. *required reduced LR for stable training.}
\vspace{-0.7em}
\centering
\footnotesize
\resizebox{\linewidth}{!}{
\begin{tabular}{@{}lcccccc@{}}
\toprule
Feature & FS & CLIP & DINO & MoCo & MAE & BEiT\\
\midrule
CLS & 83.86 & 65.63 & 73.03 & 74.26 & 49.52 & 9.87*\\
Spat. & 82.31 & 52.53 & 37.37 & 62.47 & 52.01 & 29.81\\
CLS+BN & \textbf{84.40} & \textbf{78.63} & \textbf{76.39} & \textbf{74.51} & 59.31 & 41.68\\
Spat.+BN & 82.83 & 74.59 & 68.09 & 68.58 & \textbf{59.86} & \textbf{44.27}\\
\bottomrule
\end{tabular}
  }
\label{tab:linprob}
\vspace{-1.0em}
\end{table}

\end{document}